\documentclass[Afour,sageh,times,doublespace]{sagej}

\usepackage{afterpage}
\usepackage{amsmath}
\usepackage{array}
\usepackage{caption}
\usepackage{moreverb,url}
\usepackage[nolist,nohyperlinks]{acronym}
\usepackage{enumerate}
\usepackage{gensymb}
\usepackage[colorlinks,bookmarksopen,bookmarksnumbered,citecolor=red,urlcolor=red]{hyperref}
\usepackage{bm}
\usepackage{mathtools}
\usepackage{paralist}
\usepackage{amsmath}
\usepackage[T1]{fontenc}
\usepackage{float}
\usepackage[font=small,labelfont=bf,tableposition=top]{caption}
\usepackage{physics}
\usepackage{subcaption}
\usepackage{siunitx}
\usepackage{outlines}
\usepackage{tabularx}
\usepackage{flushend}

\DeclareCaptionLabelFormat{andtable}{#1~#2  \&  \tablename~\thetable}
\newcolumntype{L}[1]{>{\raggedright\let\newline\\\arraybackslash\hspace{0pt}}m{#1}}
\newcolumntype{C}[1]{>{\centering\let\newline\\\arraybackslash\hspace{0pt}}m{#1}}
\newcolumntype{R}[1]{>{\raggedleft\let\newline\\\arraybackslash\hspace{0pt}}m{#1}}

\newcommand\BibTeX{{\rmfamily B\kern-.05em \textsc{i\kern-.025em b}\kern-.08em
T\kern-.1667em\lower.7ex\hbox{E}\kern-.125emX}}

\usepackage{todonotes}

\newcommand{\andrea}[1]{\todo[inline,color=blue!40]{\textbf{Andrea:} #1}}

\newcommand{\diag}{\text{diag}}

\def\volumeyear{2016}
\begin{document}
\runninghead{Tagliabue, Kamel et al.}
\title{Robust Collaborative Object Transportation Using Multiple MAVs}

\author{Andrea Tagliabue\affilnum{*}, Mina Kamel\affilnum{*}, Roland Siegwart and Juan Nieto}

\affiliation{All authors are with the Autonomous Systems Lab, ETH Zurich \\
	\affilnum{*}Authors equally contributed}
\corrauth{Autonomous Systems Lab, 
	ETH Zurich,
	Leonhardstrasse 21 LEE Building,
	8092 Zurich, Switzerland}

\email{mina.kamel@mavt.ethz.ch}

\begin{abstract}
Collaborative object transportation using multiple \acp{MAV} with limited communication  is a challenging problem. In this paper we address the problem of multiple \acp{MAV} mechanically coupled to a bulky object for transportation purposes without explicit communication between agents. The apparent physical properties of each agent are reshaped to achieve robustly stable transportation. Parametric uncertainties and unmodeled dynamics of each agent are quantified and techniques from robust control theory are employed to choose the physical parameters of each agent to guarantee stability. Extensive simulation analysis and experimental results show that the proposed method guarantees stability in worst case scenarios.

\end{abstract}


\maketitle
\section{Introduction}\label{sec:introduction}
This paper is dealing with the problem of collaborative object transportation using multiple \acp{MAV} in limited communication environment relying mainly on each agent sensing capabilities. Flying robots have recently received a great attention in the research community thanks to their simplicity and the advances in miniaturization technologies. A flying robot equipped with proper sensors can be useful tool to gather information about the environment quickly and efficiently~\cite{Bircher2016, nbvp2016, BorderPatroUAV1}. However, there are many limitations due to the power consumption, payload capabilities and size constraints of a single flying robot that makes the employment of a team of aerial robots an appealing approach for many applications. A team of aerial robots equipped with various sensors can collaboratively cover large area and gather information about the environment quickly and reliably, offering a solution to overcome the limitations of a single aerial robot, and a system without a single point of failure. On the other hand, a system with multiple flying robots poses new challenges in localization, control and coordination between agents. More challenges are introduced by the limitation of communication bandwidth.

\begin{figure}[t]
	\centering
	\includegraphics[width=0.5\textwidth]{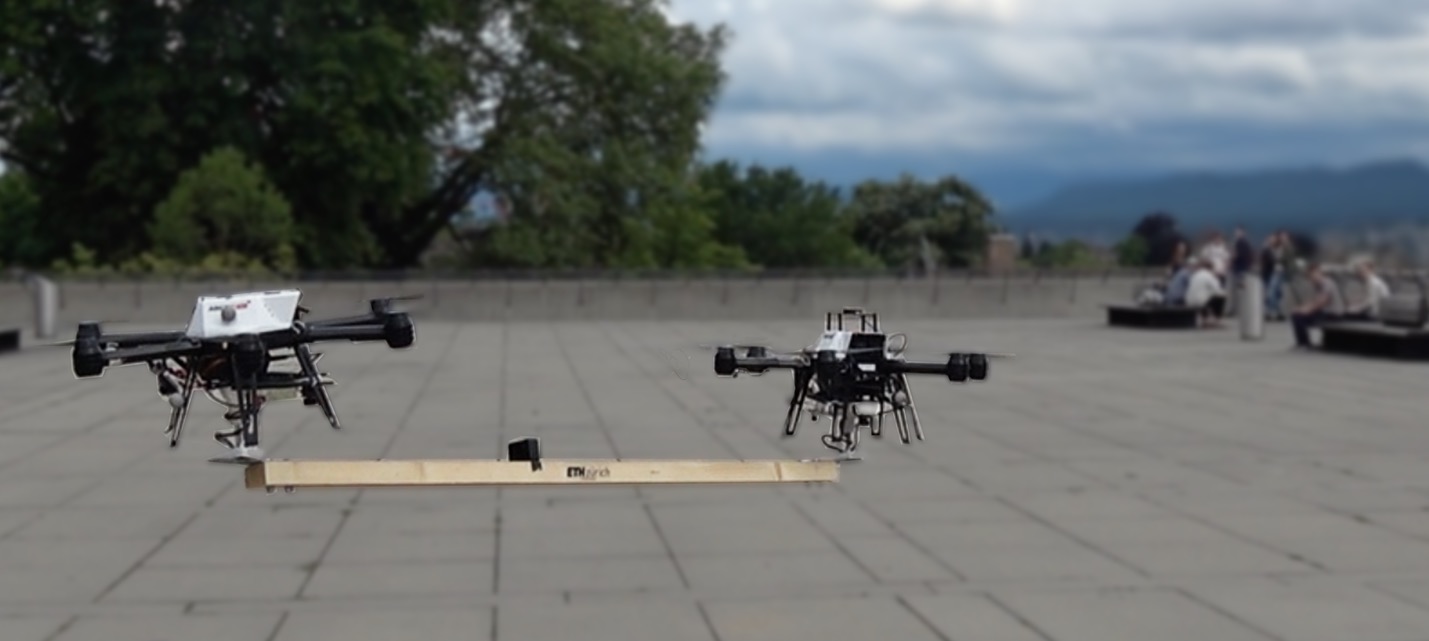}
	\caption{Experimental setup for outdoor collaborative transportation mission. The transported payload is a beam of wood, weighting $1.8$ kg and $1.5$ m long. The two robots are tightly coupled to the payload via a gripper with a spherical joint. During transportation, no communication is needed between the agents and all the algorithms are executed on-board.}
	\label{fig:2MavsOutdoor}
\end{figure}

Currently, most of the applications concerning flying robots are limited to environment sensing and monitoring. A range of new applications will be possible if flying robots acquire the ability to robustly and safely physically interact with the environment and with each other. For instance, industrial inspection relying on physical interaction with the environment \cite{fumagalli2012modeling}, goods delivery \cite{gawel2017aerial} and sensors deployment are challenging applications that require physical interaction.

In this paper we deal with the problem of multiple \acp{MAV} mechanically coupled. This coupling imposes hard kinematic constraints on the system and introduces new challenges from control perspective to guarantee stability. Most of the previous work related to mechanically coupled aerial robots is assuming slung load. Slung load imposes tight constraints on the dynamics of the \ac{MAV} to avoid undesired oscillations and introduces difficulties to navigate in cluttered environment. Another common assumption in previous work is the presence of accurate motion capture system and reliable low latency communication channel. These assumptions are unrealistic for real world system deployment. The approach presented in this paper does not rely on explicit communication between agents while being robust against state estimator imperfections, communication failure and the grasping points on the object. Moreover, the presented approach can generalize to heterogeneous team of flying robots where the capabilities and limitations  of each agent is taken into account. Robust control techniques are employed to analyze the system and to choose the optimal apparent physical parameters to guarantee robust stability of the system.

The proposed approach is evaluated in simulation where $ 5 $ agents are employed transport a bulky object. The approach is also validated in real world experiments in lab environment and outdoor under windy and varying light conditions. The experimental results show the robustness of the proposed approach.

\subsection{Paper Contributions}
The contributions of this paper can be summarized as follows:
\begin{itemize}
	\item A model of the mechanically coupled multiple \acp{MAV} is derived and uncertainty of each model block is quantified.
	\item A method to choose the apparent physical properties of agents is proposed to guarantee robust stability and maximize the performance of the whole system under the aforementioned  quantified uncertainties.
	\item We validate our approach with various experiments using multiple autonomous \acp{MAV} that collaboratively transport bulky objects.
\end{itemize}

\subsection{Assumptions}
Our approach assumes the following:
\begin{itemize}
	\item The \ac{MAV} attitude is decoupled from the object attitude. This can be achieved by mounting the grasping mechanism on spherical joint. In Section \ref{sec:model} we show that a spherical joint satisfies this assumption.
	\item The object to be transported is sufficiently large to accommodate $ N $ agents with sufficient safety distance.
	\item The payload is rigid.
\end{itemize}

The remainder of this paper is organized as follows. In Section \ref{sec:related_work} we present previous work related to collaborative object transportation using flying robots as well as ground robots. In Section \ref{sec:method} we present our approach to deal with the problem of collaborative transportation. Moreover, the controller structure and system architecture are discussed. In Section \ref{sec:model} we present the model of each agent and the whole system model. Additionally we discuss various sources of uncertainties and quantify these uncertainties. In Section \ref{sec:robust_tuning} we discuss the approach to tune the controller to guarantee robust stability against various sources of uncertainties. Finally, in Section \ref{sec:evaluation} extend simulation and experimental studies validate our approach to achieve robust and reliable collaborative object transportation.

\section{Symbols}\label{sec:symbols}
The symbols used in this works are collected in the following tables. For notation, reference frames and system-wide parameters, refer to Table \ref{tab:SystemNotation}. For symbols related to the modeling and low-level control of a single \ac{MAV}, refer to Table \ref{tab:MavModelling}. For symbols related to the admittance controller and force estimator refer, respectively, to Tables \ref{tab:AdmcDefinitionAndModelingSymbols} and \ref{tab:ForceEstimatorDefinitionAndModelingSymbols}. For symbols related to modeling of the payload, refer to Table \ref{tab:PayloadModelling}.
\begin{table*}[ht]
	\centering
	\begin{tabular}{R{1.7cm}L{8.5cm}R{2.5cm}}
		Symbol & Meaning  & Units \\ 
		\hline \hline
		$N$ 		& Number of agents & - \\  
		$I_i$ 		 & Inertial reference frame for the \textit{$i$-th} \ac{MAV} 	 & - \\
		$W$ 	   & Inertial reference frame for the payload 						&  - \\
		$B_i$ 	    & Reference frame of the \textit{$i$-th} \ac{MAV} 	   & - \\
		$P$ 	    & Reference frame of the payload 							& - \\
		$\prescript{}{A}{\boldsymbol{r}}_{BC}$ & Vector from frame $B$ to $C$ expressed in frame $A$ & [\si{\meter}] \\
		$\prescript{}{A}{\mathbf{R}}_{BC}$ & Rotation matrix from frame $C$ to frame $B$ expressed in $A$ & - \\
		$\prescript{}{}{\boldsymbol{e}_{x}}$, 
		$\prescript{}{}{\boldsymbol{e}_{y}}$, 
		$\prescript{}{}{\boldsymbol{e}_{z}}$& Orthogonal, unit norm components of a reference frame & - \\
		$\Delta h$ 	& Increment of altitude commanded by the \ac{FSM} & [\si{\meter}] \\		    
		$\boldsymbol{g}$ & Gravitational acceleration & [\si{\meter \per \second \squared}] \\
		$m_{\text{sys}}$ & Total mass of the system constituted by $N$ agents and payload & [\si{\kilogram}] \\
		$\boldsymbol{J}_{\text{sys}}$ &Total inertia of the system constituted by $N$ agents and payload  & [\si{\newton\meter}] \\	
	\end{tabular}
	\caption{System-wide symbols.}
	\label{tab:SystemNotation}
\end{table*}

\begin{table*}[ht]
	\centering
	\begin{tabular}{R{1.7cm}L{8.5cm}R{2.5cm}}
		Symbol & Meaning  & Units \\
		\hline \hline
		$\boldsymbol{p}$ & Position of a generic \ac{MAV} & [\si{\meter}] \\
		$\boldsymbol{v}$ & Velocity of a generic \ac{MAV} & [\si{\meter \per \second}] \\
		$m$ 		& Mass of a generic \ac{MAV} & [\si{\kilogram}] \\
		$\boldsymbol{n}$ & Rotation speed of the propellers & [\si{\radian\per\second}] \\
		$\boldsymbol{\omega}$ & Rotation rate of a generic \ac{MAV} & [\si{\radian \per \second}] \\
		$\dot{\boldsymbol{\omega}}$ & Angular acceleration of a generic \ac{MAV} & [\si{\radian \per \second \squared}] \\
		$\prescript{}{}{\boldsymbol{J}}$ & Inertia tensor of a generic \ac{MAV}& [\si{\kilogram \meter \squared}] \\
		$\boldsymbol{q}$ & Normalized attitude quaternion & - \\
		$\boldsymbol{\eta}$ & Vector of Euler angles & [\si{\radian}] \\
		$\phi, \theta, \psi$   & Roll, pitch and yaw Euler angles & [\si{\radian}] \\
		$\boldsymbol{M}^{\textit{ext}}$ & External torque & [\si{\newton \meter}] \\
		$\boldsymbol{F}^{\textit{ext}}$ & External force & [\si{\newton}] \\
		$\boldsymbol{F}^{\textit{aero}}$ & Aerodynamic drag force & [\si{\newton}] \\
		$k_{\text{drag}}$ & Propeller drag coefficient  & [\si{\newton \second \cubed \per \radian \squared \per \meter}]\\
		$\boldsymbol{F}^{\textit{prop}}$    & Total thrust produced by propellers & [\si{\newton}]\\
		$F_{\text{prop}}$ & Thrust force produced by propeller along $\prescript{}{B}{\boldsymbol{e}}_{z}$& [\si{\newton}] \\
		$\boldsymbol{M}^{\textit{prop}}$   & Total torque produced by propellers  & [\si{\newton \meter}] \\
		$\boldsymbol{F}^{\textit{cmd}}$ 	  & Total thrust commanded to the propellers & [\si{\newton}] \\
		$F_{\text{cmd}}$ & Commanded thrust force produced by propeller along $\prescript{}{B}{\boldsymbol{e}}_{z}$ & [\si{\newton}] \\
		$\boldsymbol{M}^{\textit{cmd}}$   & Total torque commanded to the propellers & [\si{\newton \meter}] \\
		$\phi_{\text{cmd}}$, $\theta_{\text{cmd}}$, $\psi_{\text{cmd}}$ & Commanded roll, pitch, yaw Euler angles & [\si{\radian}] \\
		$\prescript{}{}{\mathbf{\Lambda}_r}$, $\prescript{}{}{\psi_r}$ & Reference trajectory, heading & [\si{\meter}] or [\si{\radian}] \\
		$\prescript{}{}{\mathbf{\Lambda}_d}$, $\prescript{}{}{\psi_d}$ & Desired trajectory, heading & [\si{\meter}] or [\si{\radian}] \\ 
		$\tau_{\text{att}}$ & Time constant of simplified closed loop model & [\si{\second}]\\
		$\boldsymbol{K}_{P}, \boldsymbol{K}_{D}$ & Positive, diag., gain matrices of PD position controller & - \\
		$K_{P_{\phi}}, K_{D_{\phi}}$ & Gains of the roll attitude controller & - \\
		$\bar{m}$ & Maximum payload of a single agent & \si{\kilogram}  \\
		$\phi_{\text{cmd}}^{max}$, $\theta_{\text{cmd}}^{max}$ & Maximum commanded roll and pitch Euler angles & \si{\radian} \\ \hline
	\end{tabular}
	\caption{\acf{MAV} modeling and low-level control specific symbols.}
	\label{tab:MavModelling}
\end{table*}
\begin{table*}[ht]
	\centering
	\begin{tabular}{R{1.7cm}L{8.5cm}R{2.5cm}}
		Symbol & Meaning  & Units \\
		\hline \hline
		$\boldsymbol{\mathcal{M}}$ & Diag. matrix of virtual mass & [\si{\kilogram}] \\
		$\boldsymbol{\mathcal{C}}$ & Diag. matrix of virtual damping & [\si{\newton \second \per \meter}] \\
		$\boldsymbol{\mathcal{K}}$ & Diag. matrix of virtual spring constant & [\si{\newton \per \meter}] \\
		$\mathcal{J}_{\psi}$ & Virtual inertia around $\boldsymbol{e}_z$  &  [\si{\kilogram \meter \squared}] \\
		$\mathcal{C}_{\psi}$ & Virtual frictional torque around $\boldsymbol{e}_z$& [\si{\newton \meter \second \per \radian}] \\
		$\mathcal{K}_{\psi}$ &  Virtual torsional spring constant around $\boldsymbol{e}_z$ & [\si{\newton\meter\per\radian}] \\
		$\overline{T}, \underbar{$T$}, T_{\text{avg}}$ & Tuning threshold for the admittance controller & [\si{\second}] \\
		$\overline{F}, \underbar{$F$}$ & Tuning threshold for the admittance controller & [\si{\newton}] \\
	\end{tabular}
	\caption{Admittance controller definition and modeling specific symbols.}
	\label{tab:AdmcDefinitionAndModelingSymbols}
\end{table*}
\begin{table*}[ht]
	\centering
	\begin{tabular}{R{1.7cm}L{8.5cm}R{2.5cm}}
		Symbol & Meaning  & Units \\
		\hline \hline
		$\boldsymbol{x}, \boldsymbol{\xi}$ & State vector of the force estimator & -\\
		$\boldsymbol{u}$ & Input vector for prediction step & -\\
		$\boldsymbol{z}$ & Measurement vector & -\\
		$\mathbf{P}$ & State covariance & -\\
		$\mathbf{Q}$ & Process noise covariance & -\\
		$\mathbf{R}$ & Measurement noise covariance & - \\
		$\boldsymbol{F}_{\infty}^{\textit{ext}}$ & Output of the model of the force estimator assuming infinitely fast convergence speed & [\si{\newton}] \\
		$\tau_{\text{est}}$ & Time constant of the force estimator  & [\si{\second}]\\
		$T_s$ & Sampling time of the force estimator & [\si{\second}] \\
		$w_m^i$ & Weight for $i$-th sigma point in mean computation & - \\
		$w_c^i$ & Weight for $i$-th sigma point in covariance computation & - \\
	\end{tabular}
	\caption{Force estimator definition and modeling specific symbols}
	\label{tab:ForceEstimatorDefinitionAndModelingSymbols}
\end{table*}
\begin{table*}[ht]
	\centering
	\begin{tabular}{R{1.7cm}L{8.5cm}R{2.5cm}}
		Symbol & Meaning  & Units \\
		\hline \hline
		$\boldsymbol{p}_{p}$ & Position of the payload & [\si{\meter}] \\
		$\boldsymbol{v}_{p}$ & Velocity of the payload & [\si{\meter \per \second}] \\
		$m_{p}$   & Mass of the payload & [\si{\kilogram}] \\
		$\boldsymbol{F}^{\textit{drag}}$ & Aerodynamic drag force acting on the payload & [\si{\newton}] \\
		$\boldsymbol{M}^{\textit{drag}}$ & Aerodynamic drag torque acting on the payload & [\si{\newton \meter}] \\
		$\boldsymbol{F}^{\textit{agents}}$  & Total force applied by the agents on the payload & [\si{\newton}] \\
		$\boldsymbol{M}^{\textit{agents}}$ & Total torque applied by the robots on the payload &[ \si{\newton \meter}] \\	
	\end{tabular}
	\caption{Payload-specific symbols}
	\label{tab:PayloadModelling}
\end{table*}

\section{Related Work}\label{sec:related_work}
Among the first approaches for collaborative object manipulation is the work presented in \cite{570849} where centralized and decentralized control structures are proposed to accomplish cooperative tasks. Collaborative mobile manipulation of large objects in decentralized fashion is presented in \cite{988979}, where a team of robots is moving in a tightly controlled formation. The approach has been demonstrated on rigid and flexible objects. 

In \cite{multirobotJavier2017} a formation control method based on constrained optimization is presented. The proposed approach deals with static and dynamic obstacles and the authors presented two variants of the algorithm: the first algorithm is a local motion planner where the robots navigate towards the optimal formation while avoiding obstacles, the second algorithm is a global planner based on the sampling of convex regions in free space. The approach has been demonstrated in two applications, a team of aerial robots flying in formation and a team of mobile manipulators that collaboratively carry an object.

Collaborative transportation of slung load with a team of \acp{MAV} is presented in \cite{bernard2011autonomous} and \cite{maza2009multi}.  In this work, the authors propose an orientation controller that compensates for external torques introduced by the payload. However, the \acp{MAV} keep a rigid formation and compliance is introduced by the rope. The authors in \cite{michael2011cooperative} analyzed  the problem of transporting a large payload using a team of \acp{MAV} with cables. The configuration of the team is chosen to guarantee static equilibrium of the payload while respecting constraints on the tension of the cables.

Collaborative manipulation relying only on implicit communication has been presented in \cite{7353473} where two mobile manipulators collaboratively manipulate an object employing a decentralized leader-follower architecture. The interaction forces between manipulators and object are measured by force/torque sensors attached at the end-effector of each manipulator. The leader is aware of the desired object trajectory and implements it via impedance controller law while the follower estimates the desired trajectory through the object motion. 

A collaborative aerial transportation scheme without communication is presented in \cite{gassner2017dynamic} where  a leader-follower architecture is employed. In this work, the proposed approach doesn't rely on the interaction force estimation, instead it employs visual tags fixed on each gripper and on the leader \ac{MAV}. The follower motion is generated by estimating the state of the leader \ac{MAV} and the gripper attached to the payload using the visual tags. The extension of this method to multiple followers is not straight forward.

Our approach differs from all these methods as it is completely decentralized, doesn't require force/torque sensors, doesn't rely on communication between agents and can be easily extended to multiple \acp{MAV}.  

\section{Method}\label{sec:method}
\begin{figure*}
	\centering
	\includegraphics[width=\linewidth]{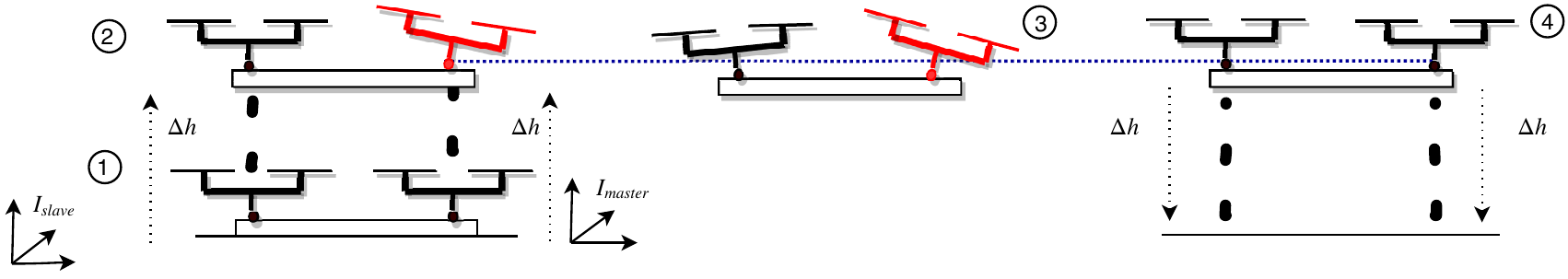}
	\caption{A schematic representation of the collaborative transportation mission.
		(1) Two agents are connected to the payload via a magnetic gripper. A centralized \acf{FSM} sends altitude increment commands $\Delta{h}$ to both the agents and make sure they both fulfill the command. No assumption is make about placement of the inertial reference frame of the agents.
		(2) Once both the \ac{MAV} have reached the desired altitude, the slave(s) agent engages the admittance controller while the master (red) starts to pull the payload, following a trajectory to the destination goal. The admittance controller allows to reshape the inertial properties of the slave agent in order to guarantee maximum compliance to the actions of the leader.
		(3) The slave agent follows the master by sensing the force that it applies on the payload. Sensed interaction force are translated into a reference trajectory via the admittance controller.
		(4) Once the master reaches the destination goal, slave's admittance controller is disengaged and a centralized \ac{FSM} sends altitude decrement commands $\Delta{h}$ to both the agents.}
	\label{fig:MissionDiagram}
\end{figure*}
In this section we present the general strategy for collaborative transportation using multiple \acp{MAV} in communication-denied environment. We provide a description of the control architecture and control algorithms employed to achieve this goal.
\par
The cooperative transportation strategy we propose is based on the biologically inspired master-slaves collaborative paradigm. This approach is typically adopted whenever the behavior of a collectivity of agents, called the slaves, is imposed by a single individual, the master. Many illustrations of this cooperative behavior can be found in nature, for example in colonies of ants active in foraging, as described by \cite{gelblum2015ant}. In this case, when a worker finds an unmanageable item, it begins to move it, triggering the help of additional workers. By sensing the force that the leader ant is applying on the load, the follower workers synchronize their effort, making the transportation to the nest feasible.
\par
 In our context we define the master-slaves method for collaborative transportation as follows. As system setup, we consider a group of $N$, with $N \in \mathbb{N}$ and $N>1$, heterogeneous \acp{MAV} agents connected to a common payload. Each robot is rigidly attached to the structure to be transported via a spherical joint, which guarantees - as we show - attitude decoupling among agents and payload, but kinetically constraints the translational dynamic of each agent to be the same as the motion of the payload. To achieve coordination we select the one agent to be the master (or leader) of the group, while the role of slave (or follower) is assigned to the remaining $N-1$ agents. The task of the master is to lead and steer the group of vehicles connected to the payload towards a desired destination, while keeping a constant transportation altitude. The slaves are not aware of the destination goal, but share the same altitude of the master, assisting their leader in the transportation task. Cooperation and coordination of the slave agents with respect to the movements of the master is achieved via re-shaping the apparent physical properties of each slave, in order that maximum compliance to the actions of the master is guaranteed. Specifically, we modify the inertial properties of each slave \ac{MAV} to make it behave as a passive point-mass, accelerated by any interaction force applied on it and subject to viscous friction directly proportional to its own velocity. This new behavior can be achieved by means of admittance control.
 \par
 The key idea behind admittance control is that a position-controlled mechanical system can achieve arbitrary interaction dynamics by sensing the force coming from the environment and by accordingly generating and following a reference trajectory. In order to do so, every slave agent is equipped with a force estimator, which senses the force applied by the master to the payload. This force information is then used by the admittance controller to regulate the position of the slave agent, so that it guarantees maximum possible compliance to the estimated force.
\par
In a collaborative transportation maneuver, the master \ac{MAV} agents behaves as if it was carrying the payload alone, pulling it towards the desired goal while keeping a constant altitude. For this reason it only executes its on-board reference tracking feedback loop. The reference trajectory which leads it to the destination goal can be provided by a user or by an on-board or off-board planning system.
\par
The proposed transportation strategy has the advantage of not relying on any explicit communication link among the agents. Information is instead shared via the payload itself, which is used as a medium to transfer the interaction force applied by the master to the slaves. This information flow is mono-directional, from the master to the slaves, as the slaves fully act according to the sensed force applied by the master on the transported structure, while the master executes its standard reference tracking algorithms. The nature of this information flow justifies the name for the approach.
\par Further advantages of the proposed method are that the agents
\begin{inparaenum}[(a)]
	\item do not have to agree on a global inertial reference frame, and
	\item do not need to share the destination goal.
\end{inparaenum}
This is made possible by the fact that each slave on-line generates a reference trajectory in the same frame as the external forces are estimated. The destination goal has to be known only by the master and, as a consequence, has to be expressed in the master's reference frame only.
\par In order to be able to transport the object, we assume that the following criteria are met:
\begin{itemize}
    \item the payload is sufficiently large so that all the agents necessary for its transportation can be directly connected on it
	\item the agents can grasp the payload in such a way that its weight force can be equally distributed among the agents.
	\item the payload is rigid.
\end{itemize}

\par In our setup we focus on the transportation maneuver, rather than on the grasping, lifting and deployment of the payload. For this reason we assume that all the agents are already connected to the object.

\subsection{Reaching the transportation altitude}
In order to reach the transportation altitude, which will remain constant during the whole maneuver, we employ a centralized coordinator, based on a \acf{FSM}. The task of this algorithm, which is executed by one of the agents or by an external processor, is to command the same increment of altitude $\Delta{h}$ to all the agents. The \ac{FSM} makes sure that all the agents have fulfilled the command before the next increment of altitude can be sent, and engages the admittance controller of each slave agent once the desired transportation altitude is reached. The same strategy (with $\Delta{h} < 0$) is employed for coordinated landing. A representation of the collaborative transportation mission is shown in Figure \ref{fig:MissionDiagram}.

\subsection{Coordinate system definition}
In this section we define the coordinate frames used for the derivation of the model and the control algorithms of the system.
\par Every \textit{$i$-th} \ac{MAV} agent, with $i \in 1...N$, employed in collaborative transportation makes use of two reference frames. The first is an inertial frame $I_i$ and the second is a body reference frame $B_i$, fixed in the \ac{CoG} of the \textit{$i$-th} agent.
\par Only for modeling and analysis purpose, we describe the position and attitude of the payload via an inertial reference frame $W$ and a reference frame attached in its \ac{CoG} defined as $P$. Similarly, only for modeling and analysis purpose, we introduce the assumption the inertial reference frame $I_i$ of every agent coincides with the payload inertial reference frame $W$, which means that $W = I_i, i=1\dots N$.
\par
All the inertial frames have positive \textit{$z$-axis} pointing upwards (gravity is negative).

\subsection{Agent's low-level control architecture}
\begin{figure*}
	\centering
	\includegraphics[width=1.0 \textwidth]{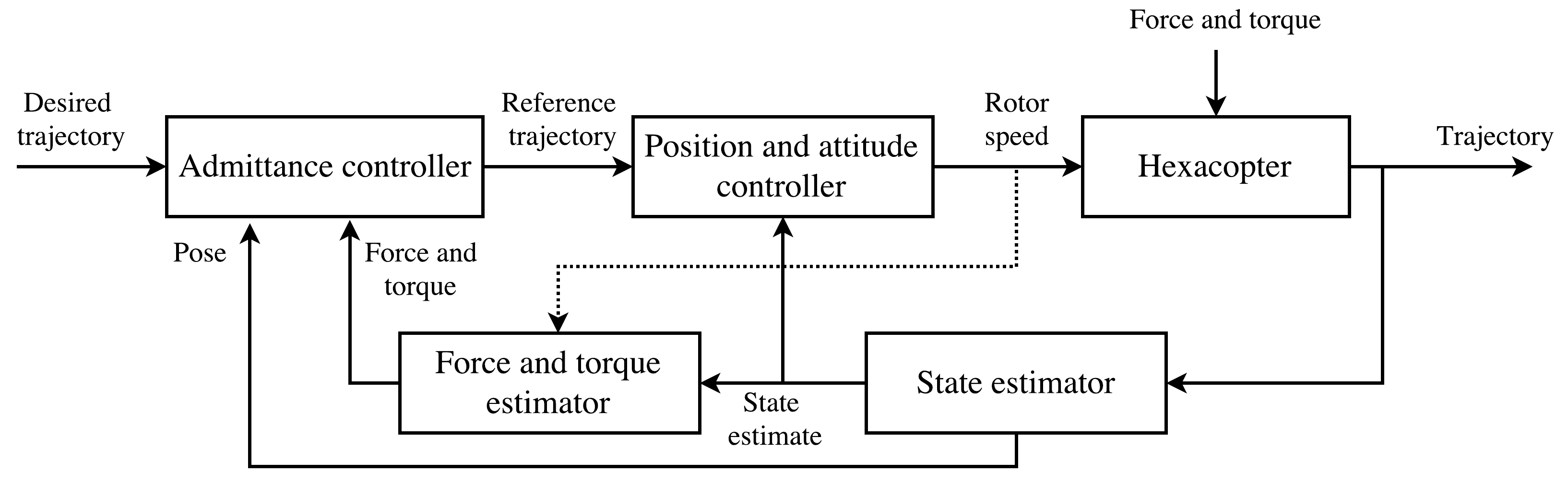}
	\caption{Control scheme for the slave agent. The user-provided desired trajectory is modified by the admittance controller according to the external force and torque acting on the vehicle. The state estimator can be either based on a \ac{VI} navigation system or a \acf{MCS}. The force is not directly measured, but rather estimated using the inertial information coming from the state estimator.}
	\label{fig:SlaveControlScheme}
\end{figure*}
The low-level control architecture of each agent, either master or slave, is composed by two main submodules.
\begin{itemize}
	\item \textbf{State estimator}:
	estimates the robot (a) position, (b) velocity, (c) attitude (expressed as a quaternion), and (d) angular velocity with respect to the agent's inertial coordinate frame $I_i$. Position measurements can either be provided by an external motion capture system or an on-board \ac{VI} navigation system developed by \cite{J.Nikolic2014} at the Autonomous Systems Lab at ETHZ and Skybotix AG. Such measurements are then fused on-board with the information coming from the \ac{IMU} using the multi-sensor estimation framework detailed in \cite{Bloesch2015a}.
	\item \textbf{Position and attitude controller}: track the trajectory generated by the admittance controller by providing a rotational speed command to the rotors. The position controller is detailed in \cite{kamelmpc2016} and \cite{2016arXiv161109240K}. The attitude control loop is cascaded within the position control loop.
\end{itemize}

\subsection{Slave-specific control architecture}
Each of the $N-1$ slave agents, in addition to the detailed low-level control architecture, executes the following on-board control algorithms.
\begin{itemize}
\item \textbf{Force estimator}: estimates the external force acting on the vehicle expressed in the respective inertial coordinate system $I_i$. Force estimates can be obtained from the sole inertial information provided by the on-board \ac{IMU} and the state estimator, or can be additionally combined with the measurement of the rotational speed of the rotors. The availability of this last information depends on the aerial platform used.
\item \textbf{Admittance controller}: generates an on-line reference trajectory (position, velocity,  attitude and angular velocity) for the position and attitude controller, given the estimate of the external force. The trajectory is expressed in the coordinate frame $I_i$.
\end{itemize}
A representation of the control architecture of a slave agent can be found in Figure \ref{fig:SlaveControlScheme}.

\subsection{Admittance Controller Formulation}
\label{subs:AdmittanceControllerFormulation}
The collaborative transportation strategy we propose is based on regulating the dynamic behavior of every slave agent during its interaction with the payload. In general, the dynamic behavior of a robot is defined by how the robot exchanges energy with the environment. This energy exchange can be characterized by of a set of motion and force variables at the point of interaction (also called \textit{interaction port}). By imposing a specific relationship between the motion and force variables we can thus regulate the energy exchanged between robot and environment and reshape its inertial behavior. An extensive explanation of interactive behavior can be found in \cite{hogan2005impedance}, while an overview on impedance and admittance control can be found in \cite{siciliano2016springer}. \cite{augugliaro2013admittance} describe an admittance controller for human-robot interaction.
\par
Admittance control allows to achieve the desired interaction behavior by regulating the motion of the robot according to the sensed or estimated interaction force. The dynamic relationship between motion (position, velocity, acceleration) and force variables is controlled by imposing an apparent inertia, damping and stiffness to the robot via the following equation:
\begin{multline}
	\boldsymbol{\mathcal{M}}
	(
		\ddot{\prescript{}{I}{\boldsymbol{\Lambda}}_{d}} - \ddot{\prescript{}{I}{\boldsymbol{\Lambda}}_{r}}
	)+
	\boldsymbol{\mathcal{C}}
	(
		\prescript{}{I}{\dot{\boldsymbol{\Lambda}}_{d}} - \prescript{}{I}{\dot{\boldsymbol{\Lambda}}_{r}}
	) \\ +
	\boldsymbol{\mathcal{K}}
	(
		\prescript{}{I}{\boldsymbol{\Lambda}_{d}} -
		\prescript{}{I}{\boldsymbol{\Lambda}_{r}}
	) = -\prescript{}{I}{\hat{\boldsymbol{F}}^{\textit{ext}}}
\end{multline}
where $\hat{\boldsymbol{F}}^{\textit{ext}} \in \mathbb{R}^3$ represents the sensed or estimated interaction force, ${\boldsymbol{\Lambda}}_{d} \in \mathbb{R}^3$ the desired motion in absence of interaction, and $\boldsymbol{\Lambda}_{r} \in \mathbb{R}^3$ the reference motion resulting from interaction, output of the admittance controller. The diagonal matrices $\boldsymbol{\mathcal{M}} = \diag(\mathcal{M}_x, \mathcal{M}_y, \mathcal{M}_z)$, $\boldsymbol{\mathcal{C}} = \diag(\mathcal{C}_x, \mathcal{C}_y, \mathcal{C}_z)$ and $\boldsymbol{\mathcal{K}} = \diag(\mathcal{K}_x, \mathcal{K}_y, \mathcal{K}_z)$ define, respectively, the apparent inertia, damping and stiffness of the robot. All the quantities are expressed with respect to the inertial reference frame of each considered agent.
\par Similarly, we can reshape the rotational behavior of the agent around $\prescript{}{B}{\boldsymbol{e}}_{z}$ by imposing the following law of motion:
\begin{multline} \label{eq:admc:yaw}
\mathcal{J}_{\psi}
	(\prescript{}{B}{\ddot{\psi}}_{d} - \prescript{}{B}{\ddot{\psi}}_{r})
	 + \mathcal{C}_{\psi}
	 	(\prescript{}{B}{\dot{\psi}}_{d} - \prescript{}{B}{\dot{\psi}}_{r}) \\
	 + \mathcal{K}_{\psi}
	 (\prescript{}{B}{\psi_{d}} - \prescript{}{B}{\psi_{r}})
= -\prescript{}{}{\boldsymbol{e}}_{z}^{T}\prescript{}{B}{\hat{\boldsymbol{M}}}^{\textit{ext}}
\end{multline}
where $\mathcal{J}_{\psi}$ represents the virtual inertia around $\prescript{}{B}{\boldsymbol{e}}_{z}$, $\mathcal{C}_{\psi}$ corresponds to the virtual damping and $\mathcal{K}_{\psi}$ the elastic constant of the virtual torsion spring. The variables $\psi_{d}$ and $\psi_{r}$ refer respectively to the desired and the reference yaw attitude. All the variables and coefficients in Eq. \ref{eq:admc:yaw} are scalars defined in $\mathbb{R}$.

\subsubsection{Trajectory generation law for maximum compliance} \label{subsubsec:AdmcMaximumCompliance}
An admittance controller generates a new reference trajectory independently for each \textit{$j$-th} axis of the inertial frame as if the \ac{MAV}, with mass $\mathcal{M}_j$, was connected to the desired trajectory via a spring with elastic constant $\mathcal{K}_j$ and via a damper with damping coefficient $\mathcal{J}_j$. The deviation from the desired trajectory is caused by the estimated interaction force, which accelerates the virtual mass. The response to the interaction force and torque can be changed by tuning the virtual spring constant $\mathcal{K}_j$. Increasing the value of $\mathcal{K}_j$ yields to a stiffer spring, which in turn improves the tracking of the desired trajectory. Conversely, setting $\mathcal{K}_j$ to zero guarantees full compliance with the estimated external force. An intuitive representation of an admittance controller and the effect of changing $\mathcal{K}_j$ can be found in Figure \ref{fig:AdmcIntuitiveExplanation}.

\begin{figure}
	\centering
	\includegraphics[width=\linewidth]{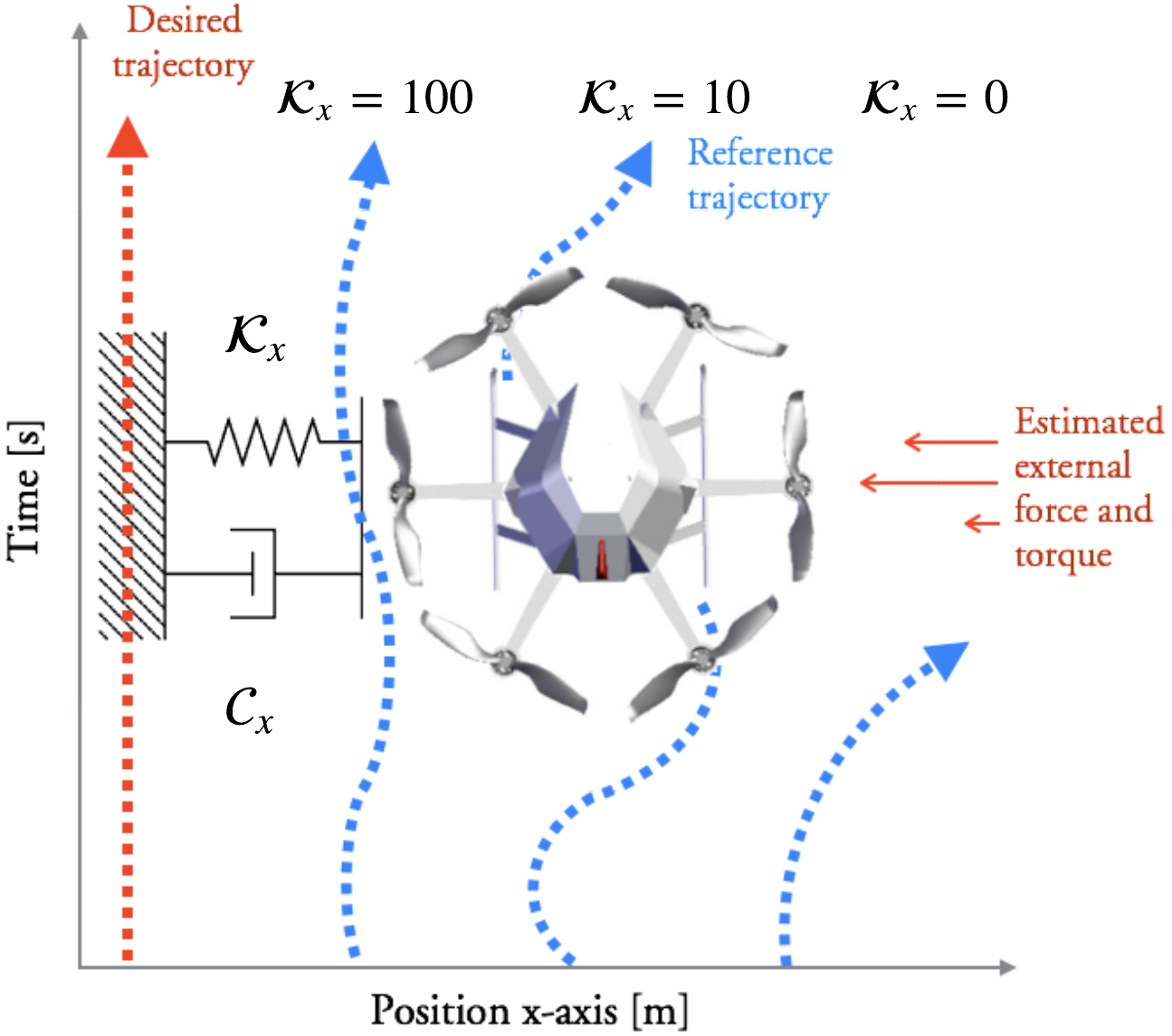}
	\caption{Intuitive explanation of admittance control. Given a desired trajectory, the admittance controller generates a new reference trajectory according to the estimated external forces by simulating a second order \textit{spring-mass-damper} dynamic system. Different values of virtual stiffness $\mathcal{K}_x$ allows to define higher or lower compliance to the sensed interaction force.}
	\label{fig:AdmcIntuitiveExplanation}
\end{figure}




\par In order to guarantee full compliance to the interaction force we set $\mathcal{K}_x$ and $\mathcal{K}_y$ to be zero. The desired position $\boldsymbol{\Lambda}_{d}$ of each slave agent is initialized to the estimated position $\prescript{}{I}{\hat{\boldsymbol{p}}}_{IB}$ of the robot at the instant that the controller is engaged. The desired velocity and acceleration $\dot{\boldsymbol{\Lambda}}_{d}$ and $\ddot{\boldsymbol{\Lambda}}_{d}$ are both set to $\boldsymbol{0}_{3\times1}$, meaning that each slave agent will maintain its position as long as no interaction force is applied. Since we assume that no torque can be applied from the payload to the vehicle and vice versa, due to the connection via the spherical joint, we will assume $\psi_{r}$ to be constant and set to the desired yaw attitude $\psi_{d}$.

\subsubsection{Robustness to noise and undesired constant external force}
\begin{figure*}
	\begin{minipage}{0.45\textwidth}
		\centering
		\raisebox{-0.5\height}{
			\includegraphics[width=\textwidth]{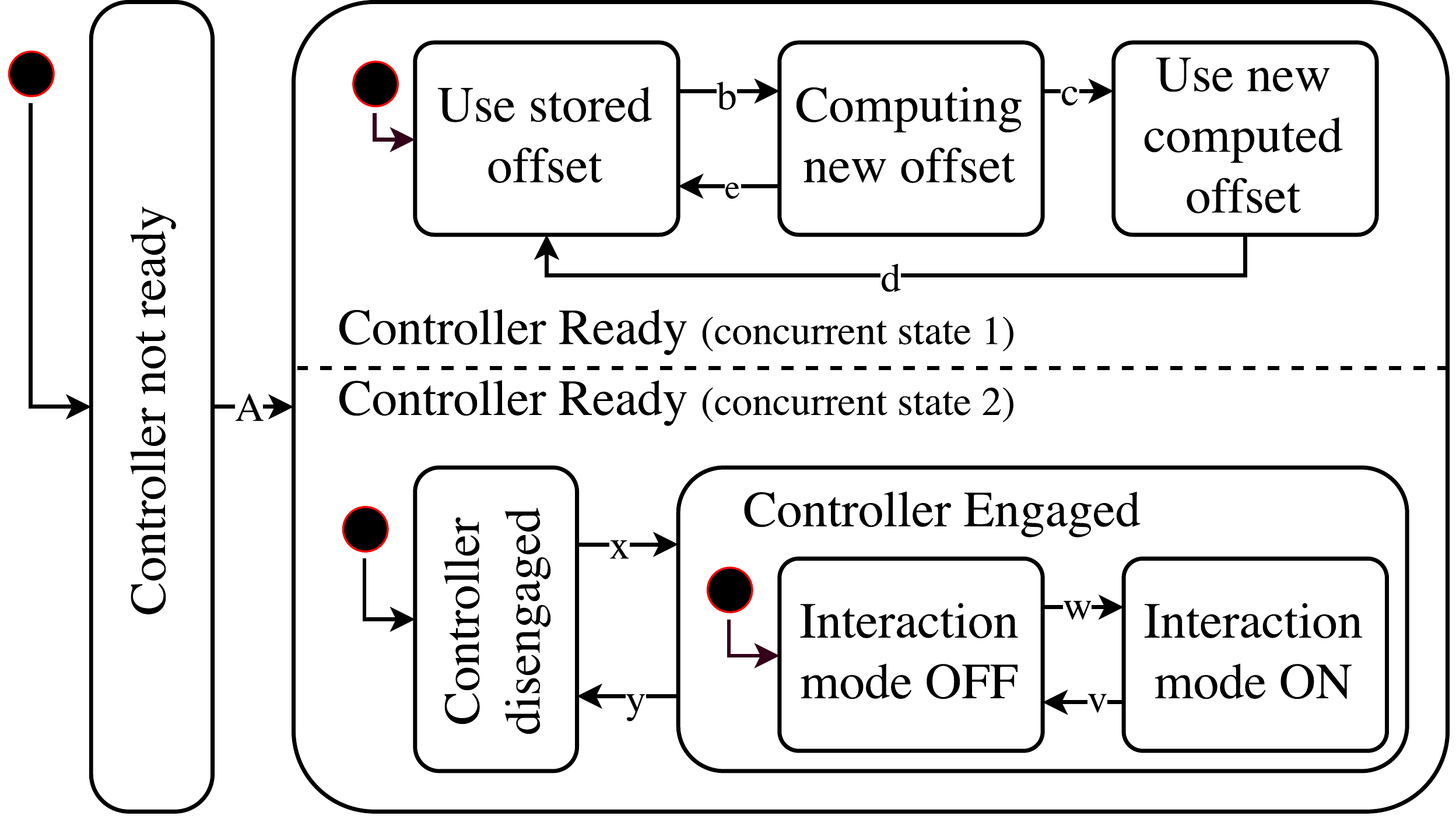}
		}
		\refstepcounter{figure}
		\label{fig:AdmcTansitionFSM}
	\end{minipage}
	\hfill
	\begin{minipage}{0.45\textwidth}
		\centering
		\raisebox{-0.5\height}{
			\scalebox{0.90}{
				\begin{tabular}{r|C{6cm}}
					\hline
					\textbf{Transition} & \textbf{Condition} \\ \hline
					\hline
					A & \texttt{force\_estimate\_received} \&\& \texttt{attitude\_estimate\_received} \\ \hline
					w & ${\hat{F_j}}^{\textit{ext}} > \overline{{F}}$ for $\Delta t > \overline{T}$ \\
					v & $\hat{{F_j}}^{\textit{ext}} < \underbar{${F}$}$ for $\Delta t > \underbar{$T$}$ \\ \hline
					b & \texttt{compute\_offset}\\
					c & $\Delta t >T_{\text{avg}}$ \\
					d, e & \texttt{remove\_offset} \\ \hline
					x & \texttt{engage\_controller}\\
					y & \texttt{disengage\_controller}\\
					\hline
				\end{tabular}
		}}
		\refstepcounter{table}
		\label{tab:AdmcTansitionFSM}
	\end{minipage}
	\setcounter{figure}{\thefigure - 1}
	\setcounter{table}{\thetable - 1}
	\captionlistentry[table]{\acf{FSM} and relative transition table in the admittance controller.}
	\captionsetup{labelformat=andtable}
	\caption{\acf{FSM} and relative transition table in the admittance controller. The value $\overline{{F}}$ and \underbar{${F}$} represent the engagement and disengagement force threshold to reject noisy force estimations. $\overline{T}$ defines the time that the estimated force has to exceed the given threshold before the controller can start to generate a new trajectory. Similarly, $\underbar{$T$}$ defines the time that the estimated force has to be below \underbar{${F}$} before the controller stops updating the reference trajectory. The scalar ${\hat{F_i}}^{\textit{ext}}$ represents the component of the external force on the \textit{$i$-th axis}. The \texttt{commands} are interfaces available to the user or to higher-level control interfaces. An example of the engagement and disengagement logic of the admittance controller, given a sinusoidal input signal, can be found in Figure \ref{fig:AdmcLogic}.}
\end{figure*}

If no force acts on the vehicle and the desired trajectory is constant, noise and other factors not included in the model (such as propeller efficiency) may result in a non-zero force estimate, which in turn causes a drift in the reference pose $\boldsymbol{\Lambda}_r$. Similarly, external factors such as constant wind or model-mismatches in the force and torque estimator may cause a constant offset in the estimated external disturbances.
\par
\begin{figure}
	\centering
	\includegraphics[width=0.9\linewidth]{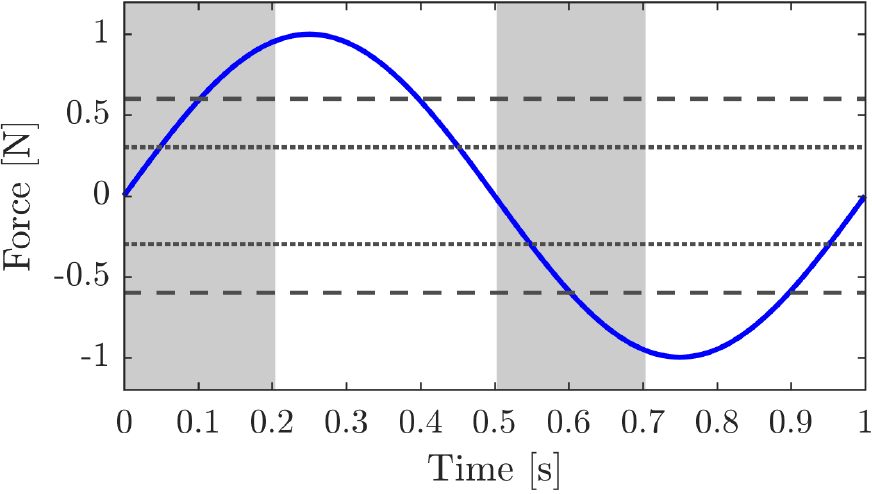}
	\caption{Example of the engagement and disengagement logic of the admittance controller.The gray area represent the disengaged admittance controller. The dashed line corresponds the upper threshold $\overline{{F}}$, while the dotted line represents the lower threshold $\underbar{${F}$}$. $\overline{{F}}$ and $\underbar{${F}$}$ have been set to $0.6$ and $0.3$ N respectively. The time threshold are set to $\overline{T} = 0.1 $ and $\underbar{T}=0.05$ s. }
	\label{fig:AdmcLogic}
\end{figure}

 In order to avoid undesired drifts in the reference trajectory, the admittance controller is integrated into a \ac{FSM}, which monitors the magnitude of the force on each axis and decides whether to reject or take into account the estimated external force to compute a new trajectory reference, similar to the work of \cite{augugliaro2013admittance}. Model mismatches and offset in the force/torque estimation are taken into account by averaging the estimated external force for a given time $T_{\text{avg}}$ while the helicopter is hovering and no external forces are applied. The computed offset is subtracted from the estimate produced by the force and torque estimator. The state chart of the integrated \ac{FSM} and relative transition table are illustrated in Figure \ref{fig:AdmcTansitionFSM} and Table \ref{tab:AdmcTansitionFSM}. An example of the engagement/disengagement logic is provided in Figure \ref{fig:AdmcLogic}.

\section{Force Estimation Techniques}
In this section we present two different strategies adopted to estimate the external forces and torques acting on a \ac{MAV}. The first strategy is based on a reduced model of the \ac{MAV} where the closed loop behavior of the attitude dynamics is approximated by a second order system. System identification techniques are used to find the parameters of the attitude dynamics system. The second strategy is based on the full system dynamics, where the \ac{MAV} rotors speed is considered as input. The use of reduced model strategy is appealing since few commercial \acp{MAV} have motor speeds feedback, making the reduced model strategy more general and suitable for many commercial platforms. The performance of the two strategies is compared in Section \ref{sec:evaluation}.

\paragraph{Premises}
We consider $I$ as the inertial reference frame of the vehicle, with positive \textit{$z$-axis} pointing upward, and $B$ as the body frame, centered in the \ac{CoG} of the \ac{MAV}. Moreover, we define $\prescript{}{I}{\boldsymbol{p}}$ and  $\prescript{}{I}{\boldsymbol{v}}$ respectively as the position and velocity of the \ac{MAV} expressed in the inertial frame $I$. We lighten the  equations by introducing $U_{1}$, $U_{2}$ and $U_{3}$ as the total torque produced by the propellers around, respectively, $x_B$, $y_B$ and $z_B$, and $U_{4}$ as the total thrust produced by the propellers along $z_B$. In the case of an hexa-copter, we compute $U_{i}$, $i = 1, \dots, 4$ from the rotor speed $n_{i}$, $i=1, \dots, 6$, using the allocation matrix, as defined in \cite{Achtelik2013}.

\subsection{Reduced Model Wrench Estimation}
In this subsection we present the reduced \ac{MAV} model used in the formulation of a filter to estimate external wrenches acting on the \ac{MAV}. The reduced model parametrizes the \ac{MAV} attitude using Euler angles.
\subsubsection{MAV Reduced Model}
\paragraph{Translational dynamics}
 The attitude of the vehicle is represented by the roll, pitch and yaw Euler angles $\boldsymbol{\eta} = (\phi, \theta, \psi)^{T}$ while  $\boldsymbol{\bm{R}_{IB}(\eta)}$ represents the rotation matrix from the fixed body frame to the inertial frame. Taking into account the following forces: (a) gravity, (b) thrust, (c) linear drag forces, (d) external forces due to interaction with the vehicle, the translational dynamics can be written as follows:
\begin{subequations}
\label{eq:transldyn_reduced_model}
	\begin{align}
	\prescript{}{I}{\dot{\boldsymbol{p}}} &= \prescript{}{I}{\boldsymbol{v}} \\
	\prescript{}{I}{\dot{\boldsymbol{v}}}
	&= \frac{1}{m}\boldsymbol{\bm{R}_{IB}(\eta)}\Bigg(
	\begin{bmatrix}
	0\\
	0\\
	U_{4}\\
	\end{bmatrix}
	-
	\boldsymbol{K}_{\text{drag}}\prescript{}{B}{\boldsymbol{v}} \Bigg)
	-
	\begin{bmatrix}
	0\\0\\g
	\end{bmatrix}
	+
	\frac{1}{m} \boldsymbol{\prescript{}{I}{F}}^{ext}
	\end{align}
\end{subequations}

\noindent where $ m $ is the vehicle mass, $\boldsymbol{K}_{\text{drag}}$ is positive diagonal matrix that represents the drag coefficients of the vehicle, $ g $ is the gravitational acceleration and $ \boldsymbol{\prescript{}{I}{F}}^{ext}  $ is the vector of external forces acting on the vehicle expressed in inertial frame.

\paragraph{Rotational dynamics}
The closed loop behavior of the rotational dynamics can be approximated using a second order system. Assuming small attitude angle, we can approximate the rotational dynamics as follows:
\begin{equation}\label{eq:rotational_dynamics_reduced}
\ddot{\phi} = \omega_{n, \phi}^{2}(k_{\phi} \phi_{\text{cmd}} - \phi )
-
2\xi_{\phi}\omega_{n, \phi}\dot{\phi}
+
\frac{\prescript{}{B}M_{x}^{ext}}{J_{xx}}
\end{equation}
\noindent where $ \omega_{n, \phi}, k_{\phi} $ and $ \xi_{\phi} $  are positive constants. $ J_{xx} $ is the inertia around $ x_B $ and  $ \prescript{}{B}M_{x}^{ext} $ represents the external torque around $ x_B $. Equation~\eqref{eq:rotational_dynamics_reduced} represents the rotational dynamics around $ x_B $ axis, similar equations are used for rotations around $ y_B $ and $ z_B $ axes.

\subsubsection{Filter Formulation}
An augmented state \ac{EKF} is used to estimate the external wrenches using the aforementioned reduced model. The state of the \ac{EKF} is defined by:
\begin{equation}
\boldsymbol{x} =
\begin{bmatrix}
\prescript{}{I}{\boldsymbol{p}}^T& \prescript{}{I}{\boldsymbol{v}}^T& \boldsymbol{\eta}^T& \prescript{}{B}{\boldsymbol{\omega}}^T& {\prescript{}{I}{\boldsymbol{F}}^{ext}}^T& {\prescript{}{B}{\boldsymbol{M}}^{ext}}^T \\
\end{bmatrix}^T.
\end{equation}

The dynamics in Equations~\eqref{eq:transldyn_reduced_model} and~\eqref{eq:rotational_dynamics_reduced} are used as process model in the \ac{EKF} formulation. The external forces and torques are assumed to be driven only by zero mean white noise since their dynamics is unknown.

The pose of the \ac{MAV} is directly measured by the external motion capture system or by the on-board localization algorithm. Therefore, the measurement vector is given by:
\begin{equation}
\boldsymbol{z} =
\begin{bmatrix}
\prescript{}{I}{\boldsymbol{p}}^T& \boldsymbol{\eta}^T\\
\end{bmatrix}^T.
\end{equation}
\noindent The measurement is assumed to be corrupted by additive white noise.

\noindent The input vector of the filter is given by the total thrust and attitude commands:
 \begin{equation}
 \boldsymbol{u} =
 \begin{bmatrix}
 \phi_{\text{cmd}} & \theta_{\text{cmd}} & \psi_{\text{cmd}} & U_{4}\\
 \end{bmatrix}^T.
 \end{equation}

\subsection{Full Model Wrench Estimation}
In this section, we present the external wrench (force and torque) estimator based on the  \ac{UKF}, with the attitude represented as a unit quaternion. First, we introduce a nonlinear discrete time model of the rotational and translational dynamics of a multi-copter subject to external force and torque, to be used in the derivation of the process model of the filter. Second, we derive a linear measurement model to be used in the update step. Third, we describe the prediction and update step for the whole state of the filter with the exception of the attitude quaternion. In the end, we describe the singularity free prediction and update steps which take into account the attitude quaternion.

\subsubsection{Multi-rotor model with external force and torque}
We present the rotational and translational dynamic for a generic multi-rotor, subject to external force and torque. A detailed description of the derivation of the model for a quad-copter is provided by \cite{bouabdallah2007design}.


\subsection{Reduced Model Wrench Estimation}
In this subsection we present the reduced \ac{MAV} model used in the formulation of a filter to estimate external wrenches acting on the \ac{MAV}. The reduced model parametrizes the \ac{MAV} attitude using Euler angles.
\subsubsection{MAV Reduced Model}
\paragraph{Translational dynamics}
The attitude of the vehicle is represented by the roll, pitch and yaw Euler angles $\boldsymbol{\eta} = (\phi, \theta, \psi)^{T}$ while  $\boldsymbol{\bm{R}_{IB}(\eta)}$ represents the rotation matrix from the fixed body frame to the inertial frame. Taking into account the following forces: (a) gravity, (b) thrust, (c) linear drag forces, (d) external forces due to interaction with the vehicle, the translational dynamics can be written as follows:
\begin{subequations}
	\label{eq:transldyn_reduced_model}
	\begin{align}
	\prescript{}{I}{\dot{\boldsymbol{p}}} &= \prescript{}{I}{\boldsymbol{v}} \\
	\prescript{}{I}{\dot{\boldsymbol{v}}}
	&= \frac{1}{m}\boldsymbol{\bm{R}_{IB}(\eta)}\Bigg(
	\begin{bmatrix}
	0\\
	0\\
	F_{\text{cmd}}\\
	\end{bmatrix}
	-
	\boldsymbol{K}_{\text{drag}}\prescript{}{B}{\boldsymbol{v}} \Bigg)
	+
	\frac{1}{m} \boldsymbol{\prescript{}{I}{F}}^{\textit{ext}}
	- \prescript{}{I}{\boldsymbol{g}}
	\end{align}
\end{subequations}

\noindent where $ m $ is the vehicle mass, $\boldsymbol{K}_{\text{drag}}$ is positive diagonal matrix that represents the drag coefficients of the vehicle, $ \boldsymbol{g} $ is the gravitational acceleration and $ \boldsymbol{\prescript{}{I}{F}}^{\textit{ext}}  $ is the vector of external forces acting on the vehicle expressed in inertial frame.

\paragraph{Rotational dynamics}
The closed loop behavior of the rotational dynamics can be approximated using a second order system. Assuming small attitude angle, we can approximate the rotational dynamics as follows:
\begin{equation}\label{eq:rotational_dynamics_reduced}
\ddot{\phi} = \omega_{n, \phi}^{2}(k_{\phi} \phi_{\text{cmd}} - \phi )
-
2\xi_{\phi}\omega_{n, \phi}\dot{\phi}
+
\frac{\prescript{}{B}M_{x}^{\textit{ext}}}{J_{xx}}
\end{equation}
\noindent where $ \omega_{n, \phi}, k_{\phi} $ and $ \xi_{\phi} $  are positive constants. $ J_{xx} $ is the inertia around $ x_B $ and  $ \prescript{}{B}M_{x}^{\textit{ext}} $ represents the external torque around $ x_B $. Equation~\eqref{eq:rotational_dynamics_reduced} represents the rotational dynamics around $ x_B $ axis, similar equations are used for rotations around $ y_B $ and $ z_B $ axes.

\subsubsection{Filter Formulation}
An augmented state \ac{EKF} is used to estimate the external wrenches using the aforementioned reduced model. The state of the \ac{EKF} is defined by:
\begin{equation}
\boldsymbol{x} =
\begin{bmatrix}
\prescript{}{I}{\boldsymbol{p}}^T& \prescript{}{I}{\boldsymbol{v}}^T& \boldsymbol{\eta}^T& \prescript{}{B}{\boldsymbol{\omega}}^T& {\prescript{}{I}{\boldsymbol{F}}^{\textit{ext}}}^T& {\prescript{}{B}{\boldsymbol{M}}^{\textit{ext}}}^T \\
\end{bmatrix}^T.
\end{equation}

The dynamics in Equations~\eqref{eq:transldyn_reduced_model} and~\eqref{eq:rotational_dynamics_reduced} are used as process model in the \ac{EKF} formulation. The external forces and torques are assumed to be driven only by zero mean white noise since their dynamics is unknown.

The pose of the \ac{MAV} is directly measured by the external motion capture system or by the on-board localization algorithm. Therefore, the measurement vector is given by:
\begin{equation}
\boldsymbol{z} =
\begin{bmatrix}
\prescript{}{I}{\boldsymbol{p}}^T& \boldsymbol{\eta}^T\\
\end{bmatrix}^T.
\end{equation}
\noindent The measurement is assumed to be corrupted by additive white noise.

\noindent The input vector of the filter is given by the total thrust and attitude commands:
\begin{equation}
\boldsymbol{u} =
\begin{bmatrix}
\phi_{\text{cmd}} & \theta_{\text{cmd}} & \psi_{\text{cmd}} & F_{\text{cmd}}\\
\end{bmatrix}^T.
\end{equation}

\subsection{Unscented Kalman Filter with Attitude Quaternion for Wrench Estimation}
In this section, we present the external wrench estimator based on the \ac{UKF}, with the attitude represented as a unit quaternion. The \ac{UKF} was first introduced by \cite{julier2004unscented}. An approach which takes into account quaternion for unscented filtering is described by \cite{crassidis2003unscented}. An \ac{UKF} for wrench estimation on \acp{MAV} with attitude represented as quaternion is described in the work of \cite{mckinnon2016unscented}. Our Work differs from \cite{mckinnon2016unscented} since we take into account the findings about the attitude reset step described by \cite{mueller2016covariance}.

\subsubsection{MAV Full Model}
We present the rotational and translational dynamic for a generic multi-rotor, subject to external force and torque. The attitude is parametrized using the unitary quaternion. A detailed description of the derivation of the model for a quad-copter is provided by \cite{bouabdallah2007design}.

\paragraph{Translational dynamics}
The attitude of the robot is represented by the unitary quaternion $\boldsymbol{q}$, while $\boldsymbol{R_{IB}(q)}$ represent the rotation matrix from the body frame $B$ to the inertial reference $I$. We express the linear acceleration $\prescript{}{I}{\dot{\boldsymbol{v}}}$ with respect to $I$ reference frame by taking into account the following forces: (a) gravity, (b) thrust, (c) aerodynamic effects, (d) external forces due to interaction with the vehicle. We obtain:
\begin{subequations}
	\label{transldyn}
	\begin{align}
\prescript{}{I}{\dot{\boldsymbol{p}}} &= \prescript{}{I}{\boldsymbol{v}} \\
\prescript{}{I}{\dot{\boldsymbol{v}}}
 & = \frac{1}{m}
	\boldsymbol{R_{IB}(q)}
	\Bigg(
			\begin{bmatrix}
			0\\
			0\\
			F_{\text{cmd}}\\
			\end{bmatrix}
			-\prescript{}{B}{\boldsymbol{F}}^{\textit{aero}}
	\Bigg)
+ \frac{1}{m} \prescript{}{I}{\boldsymbol{F}}^{\textit{ext}}
-
\prescript{}{I}{\boldsymbol{g}}
	\end{align}
\end{subequations}
where $\prescript{}{I}{\boldsymbol{F}}^{\textit{ext}}$ are the external forces that act on the vehicle expressed in $I$ reference frame and $m$ is the mass of the vehicle. For an hexa-copter, $\prescript{}{B}{\boldsymbol{F}}^{\textit{aero}}$ is defined as:
\begin{equation}
\prescript{}{B}{\boldsymbol{F}}^{\textit{aero}} = k_{\text{drag}} \sum_{i=1}^{6}n_{i}^2
\begin{bmatrix}
\boldsymbol{e}_x^T \\
\boldsymbol{e}_y^T \\
\boldsymbol{0}_{3 \time 1}^T
\end{bmatrix}
\prescript{}{B}{\boldsymbol{v}}_{IB}
\end{equation}
The scalar $k_{\text{drag}}$ lumps the aerodynamic drag force of the body and the aerodynamic forces due to the blade flapping. The \ac{MAV} model parameters can be estimated using the method proposed by \cite{burri2016maximum}.

\paragraph{Rotational dynamics}
We express the angular acceleration $\prescript{}{B}{\dot{\boldsymbol{\omega}}}$ in $B$ frame by taking into account the following torques:
\begin{inparaenum}[(a)]
     \item total torque produced by the actuators,
	\item external torque around $\prescript{}{B}{\boldsymbol{e}}_{z}$,
	\item inertial effects
\end{inparaenum}
We obtain:
\begin{equation}
\prescript{}{B}{\boldsymbol{\dot{\omega}}} = \prescript{}{B}{\boldsymbol{J}}^{-1}
\big(
 \prescript{}{B}{\boldsymbol{M}}^{\textit{prop}}
- \prescript{}{B}{\boldsymbol{\omega}}\times \prescript{}{B}{\boldsymbol{J}}\prescript{}{B}{\boldsymbol{\omega}}
+\prescript{}{B}{\boldsymbol{M}}^{\textit{ext}}
\big)
\label{rotdyn}
\end{equation}
where $\prescript{}{B}{\boldsymbol{M}^{\textit{ext}}}$ is the external torque due to interaction with the robot, expressed in the body frame $B$ of the vehicle. The matrix $\prescript{}{B}{\boldsymbol{J}}$ is the diagonal inertia tensor of the \ac{MAV} with respect to  $B$  frame.

\subsubsection{Filter definition}
\paragraph{Process model}
\label{ukf:subs:process_model}
 We augment the state vector of the system given by  Eq. \ref{transldyn} and Eq. \ref{rotdyn} so that we take into account the external wrench. We assume that the change of external force and torque is purely driven by zero mean additive process noise, whose covariance is a tuning parameter of the filter. Since we are not interested in the torque components around $\prescript{}{B}{\boldsymbol{e}}_x$ and $\prescript{}{B}{\boldsymbol{e}}_y$, we only consider $M^{\textit{ext}}_x$ to reduce the computational complexity. We obtain:
\begin{equation} \label{ukf:pred:sysdyn}
\boldsymbol{x} =
\begin{bmatrix}
\prescript{}{I}{\boldsymbol{p}}^T&\prescript{}{I}{\boldsymbol{v}}^T&\boldsymbol{q}^T&\prescript{}{B}{\boldsymbol{\omega}}^T&{\prescript{}{I}{\boldsymbol{F}}^{\textit{ext}}}^T&\prescript{}{B}{M}_{z}^{\textit{ext}}
\end{bmatrix}^T
\end{equation}
The input vector $\boldsymbol{u}$ of the filter is given by
\begin{equation}
\boldsymbol{u} = \boldsymbol{n}.
\end{equation}
In order to define the state covariance and the process noise covariance matrices, we introduce a new representation $\boldsymbol{\xi}$ of the state ${\boldsymbol{x}}$, where we substitute the attitude quaternion $\boldsymbol{q}$ with a $3 \times 1$ attitude error vector $\boldsymbol{\epsilon}$. This approach, which is called \ac{USQUE} and was proposed by \cite{crassidis2003unscented}, allows us to avoid to incur in singular representations of the state covariance. The mapping from the unitary attitude quaternion to the attitude error vector and vice-versa is done via the \acp{MRP}, which are illustrated in appendix A. Further information can be also found in \cite{shuster1993survey}. The obtained state vector is defined as:
\begin{equation}
\boldsymbol{\xi} =
\begin{bmatrix}
\prescript{}{I}{\boldsymbol{p}}^T&\prescript{}{I}{\boldsymbol{v}}^T&\boldsymbol{\epsilon}^T& \prescript{}{B}{\boldsymbol{\omega}}^T&{\prescript{}{I}{\boldsymbol{F}}^{\textit{ext}}}^T&\prescript{}{B}{M}_{z}^{\textit{ext}}  \\
\end{bmatrix}^T
\end{equation}
\par We can now introduce $\mathbf{P}$, the $16 \times 16$ covariance matrix associated to the state $\boldsymbol{\xi}$, and $\mathbf{Q}$, the $16 \times 16$ time invariant process noise diagonal covariance matrix, under the assumption that our system is subject to zero mean, additive Gaussian process noise.
\par Given the sampling time $T_s$, we discretize all the states using forward Euler method, with the exception of the normalized attitude quaternion $\boldsymbol{q}_k$:
\begin{equation} \label{eq:NonLinearFullSystemDynamics}
\boldsymbol{x}_{k+1} = {f}_k({\boldsymbol{x}}_k, {\boldsymbol{u}}_k)
\end{equation}
where the external force and torque are updated according to:
\begin{equation}
\begin{aligned}
\prescript{}{I}{\boldsymbol{F}}_{k+1}^{\textit{ext}} &= \prescript{}{I}{\boldsymbol{F}}_k^{\textit{ext}} \\
\prescript{}{B}{M}_{z,k+1}^{\textit{ext}} &= \prescript{}{B}{M}_{z,k}^{\textit{ext}}. \\
\end{aligned}
\end{equation}
The attitude quaternion is integrated using the approach proposed in \cite{crassidis2003unscented}.
\begin{subequations}
	\begin{align}\label{d_att}
	\boldsymbol{q}_{k+1} = & \boldsymbol{\Omega}(\prescript{}{B}{\boldsymbol{\omega}}_k)\boldsymbol{q}_k \\
	& \boldsymbol{\Omega}(\prescript{}{B}{\boldsymbol{\omega}}_k) =
		\begin{bmatrix}
		\Upsilon_{k} \mathbf{I}_{3 \times 3}  & \mathbf{\Psi}_k \\
		-\mathbf{\Psi}_k^T &  \Upsilon
		\end{bmatrix} \\
	& \mathbf{\Psi}_k = \sin{\left(\frac{1}{2} \norm{\boldsymbol{\omega}_k}T_s\right)}\frac{\prescript{}{B}{\boldsymbol{\omega}}_k}{\norm{\prescript{}{B}{\boldsymbol{\omega}}_k}} \\
	& \Upsilon_{k} =  \cos(\frac{1}{2}\norm{\prescript{}{B}{\boldsymbol{\omega}}_k} T_s).
	\end{align}
\end{subequations}

\paragraph{Measurement model}
We define the measurement vector $\boldsymbol{z}$ in Eq. \ref{ukf:meas_vect}, the linear measurement function in Eq. \ref{ukf:meas_funct} and the  associated $12 \times 12$ diagonal measurement noise covariance matrix $\mathbf{R}$, under the assumption that the measurements are subject to zero-mean additive Gaussian noise.
\begin{equation}
	\boldsymbol{z} =
	\begin{bmatrix}
	\label{ukf:meas_vect}
	{\prescript{}{I}{\boldsymbol{p}}^m}^T& {\prescript{}{I}{\boldsymbol{v}}^m}^T& {\boldsymbol{\epsilon}^m}^T& 	{\prescript{}{B}{\boldsymbol{\omega}}^m}^T
	\end{bmatrix}^T
\end{equation}
\begin{subequations}\label{ukf:meas_funct}
	\begin{align}
		\boldsymbol{z}_k =&  \boldsymbol{H}{\boldsymbol{\xi}}_k \\
		& \boldsymbol{H} =
		\begin{bmatrix}
			\mathbf{I}_{12\times12} & \mathbf{0}_{12 \times 4}
		\end{bmatrix}
	\end{align}
\end{subequations}
As a remark, we observe that the measured attitude quaternion $\boldsymbol{q}_k^m$ is taken into account in the measurement vector via the measured attitude error vector $\boldsymbol{\epsilon}_k^m$. The mapping between the two is done using the \acp{MRP}.

\paragraph{Notation declaration:}
From now on, we will use the following notation:
\begin{itemize}
	\item $\hat{\boldsymbol{x}}_{k-1}^{+}$ denotes the estimated state before the prediction step;
	\item $\hat{\boldsymbol{x}}_{k}^{-}$ denotes the estimated state after the prediction step and before the update step;
	\item $\hat{{\boldsymbol{x}}}_{k}^{+}$ denotes the estimated state after the update step.
\end{itemize}
The same notation applies for the state covariance matrix  $\mathbf{P}_k$.


\subsubsection{Position, velocity, angular velocity, external wrench estimation}
\label{ukf_simpl_updt}
\paragraph{Initialization}
\begin{enumerate}
\item We initialize the algorithm with:
\begin{equation}
\begin{aligned}
\hat{\boldsymbol{\xi}}_0^+ &= E[\boldsymbol{\xi}_0] \\
\mathbf{P}_0^+ &= E[(\boldsymbol{\xi}_0 - \hat{\boldsymbol{\xi}}_0^+)(\boldsymbol{\xi}_0-\hat{\boldsymbol{\xi}}_0^+)^T]
\end{aligned}
\end{equation}
\end{enumerate}
\paragraph{Prediction step}
The predicted value for all the states with the exception of the attitude is computed using the standard \ac{UKF} prediction step, as explained in \cite{Julier1996} and \cite{DSimon2006}.
\begin{enumerate}
	\setcounter{enumi}{1}
\item Generate $2n+1$ sigma points $\hat{\boldsymbol{\xi}}_{k-1}^{(i)}$, centered around the mean $\hat{\boldsymbol{\xi}}_{k-1}^+$ and with spread $\boldsymbol{\xi}^(i)$ computed from the covariance $\mathbf{P}_{k-1}^+$. The number of states of the filter $n$ corresponds, in our case, to $16$.
\begin{equation} \label{ukf:pred:sigma_points}
\begin{aligned}
\hat{\boldsymbol{\xi}}_{k-1}^{(i)} &= \hat{\boldsymbol{\xi}}_{k-1}^+  + \boldsymbol{\xi}}^{(i) &i = 0, \dots, 2n \\
&\boldsymbol{\xi}^{(0)}= \boldsymbol{0}_{16} \\
&\boldsymbol{\xi}^{(i)} = \left (\sqrt{(\lambda + n)\mathbf{P}_{k-1}^+}\right ) _i^T &i = 1,\dots, n \\
&\boldsymbol{\xi}^{(i+n)}= - \left (\sqrt{(\lambda + n)\mathbf{P}_{k-1}^+}\right)_i^T &i = 1, \dots, n
\end{aligned}
\end{equation}
The spread of the sigma points around the mean can be controlled via the scalar tuning parameter $\lambda$. The matrix square root is computed using the Cholesky decomposition. The notation $(\boldsymbol{A})_i$ denotes the \textit{$i$-th} row of the matrix $\boldsymbol{A}$.

\item Propagate the sigma points through the non-linear system dynamic model in Eq. \ref{eq:NonLinearFullSystemDynamics}, by also taking into account the current measurement of the speed of the rotors in $\boldsymbol{u}_k$.
\begin{equation} \label{ukf:pred:propagation}
\begin{aligned}
\hat{\boldsymbol{x}}_{k}^{(i)} = f_k(\hat{\boldsymbol{x}}_{k-1}^{(i)}, \boldsymbol{u																																																																																																																																																																																																																																																																																																																																																																																																																																																																																																																																																																																																																																																																																																																																																																																																																																																																																																																																																																																			}_k) &&  i=0,\dots, 2n
\end{aligned}
\end{equation}
The mapping from the sigma point $\hat{\boldsymbol{\xi}}_{k-1}^{(i)}$ containing the attitude error vector to the sigma point $\hat{\boldsymbol{x}}_{k-1}^{(i)}$ containing the attitude expressed as a unit quaternion is done through the \acp{MRP} and is detailed in Subsection \ref{subsec:QuaternionBasedAttitudeEstimation}.

\item Compute mean and covariance of the propagated sigma points to obtain the mean (Eq.\ref{ukf_pred_mean}) and covariance (Eq. \ref{ukf_pred_cov}) of the predicted (a priori) state.
\begin{equation} \label{ukf_pred_mean}
\hat{\boldsymbol{\xi}}_{k}^-= \sum_{i = 0}^{2n} {w_m^i} \hat{\boldsymbol{\xi}}_{k}^{(i)}
\end{equation}
\begin{equation} \label{ukf_pred_cov}
\mathbf{P}_{k, \text{prereset}}^- = \sum_{i = 0}^{2n}[{w_c^i}(\hat{\boldsymbol{\xi}}_{k}^{(i)} -\hat{\boldsymbol{\xi}}_{k}^-)(\hat{\boldsymbol{\xi}}_{k}^{(i)} -\hat{\boldsymbol{\xi}}_{k}^-)^T] + \mathbf{Q}_k
\end{equation}
 Where we have introduced the scalars $w_m^i$ and $w_c^i$ as the weight factors to be assigned to every sigma point. A common tuning choice is to set weight $0$ for the $0$-th sigma point and equal weight $1/2n$ for every other. The conversion from the representation $\hat{\boldsymbol{x}}_{k}^{(i)}$ to $\hat{\boldsymbol{\xi}}_{k}^{(i)}$ is explained Subsection \ref{subsec:QuaternionBasedAttitudeEstimation}.
\item Introduce the covariance correction step due to the attitude reset (in Eq. \ref{eq:ukf:predicted_attitude_quaternion}), as explained in \cite{mueller2016covariance}.
Change the predicted covariance as:
\begin{equation}
\mathbf{P}_{k}^- =
\boldsymbol{T}_{\text{reset}}(\hat{\boldsymbol{\epsilon}}_{k}^-)
\mathbf{P}_{k, \text{prereset}}^-
\boldsymbol{T}_{\text{reset}}(\hat{\boldsymbol{\epsilon}}_{k}^-)^T.
\end{equation}
Where the covariance reset matrix is computed from:
\begin{equation}
\boldsymbol{T}_{\text{reset}}(\hat{\boldsymbol{\epsilon}}_{k}^-) =
 \diag(\boldsymbol{I}_{6 \times 6}, \boldsymbol{R}(\frac{1}{2}\hat{\boldsymbol{\epsilon}}_{k}^-), \boldsymbol{I}_{7 \times 7})
\end{equation}
and  $\boldsymbol{R}(\frac{1}{2}\hat{\boldsymbol{\epsilon}}_{k}^-)$ represents the rotation matrix corresponding to half of the rotation in $\hat{\boldsymbol{\epsilon}}_{k}^-$.
\end{enumerate}

\paragraph{Linear Kalman filter based update step}
Because the measurement model corresponds to a linear function, we can use the standard KF update step for all the states in the vector state $\boldsymbol{\xi}_k$.
\begin{enumerate}
\setcounter{enumi}{5}
\item First, compute the Kalman gain matrix as
\begin{equation}
\boldsymbol{K}_k = \mathbf{P}_k^- \boldsymbol{H}^T (\boldsymbol{H} \mathbf{P}_k^- \boldsymbol{H}^T + \mathbf{})^{-1}
\end{equation}

\item Then, the updated state is obtained from
\begin{equation} \label{eq:UpdateStateMean}
\hat{\boldsymbol{\xi}}_k^+= \hat{\boldsymbol{\xi}}_k^- + \boldsymbol{K}_k(\boldsymbol{z}_k - \boldsymbol{H}\hat{\boldsymbol{\xi}}_k^- )
\end{equation}

\item Finally, the updated state covariance is obtained from
\small
\begin{equation} \label{eq:UpdateStateCovariance}
\mathbf{P}_{k, \text{prereset}}^+ = (\mathbf{I}_{16\times16} - \boldsymbol{K}_k \boldsymbol{H}) \mathbf{P}_k^- (\mathbf{I}_{16\times16} - \boldsymbol{K}_k \boldsymbol{H})^T + \boldsymbol{K}_k \mathbf{} \boldsymbol{K}_k^T
\end{equation}
\normalsize
and, including the covariance correction step, we obtain:
\begin{equation}
\mathbf{P}_{k}^+ =
\boldsymbol{T}_{\text{reset}}(\hat{\boldsymbol{\epsilon}}_{k}^+)
\mathbf{P}_{k, \text{prereset}}^+
\boldsymbol{T}_{\text{reset}}(\hat{\boldsymbol{\epsilon}}_{k}^+)^T.
\end{equation}
\end{enumerate}

\subsubsection{Quaternion based attitude estimation} \label{subsec:QuaternionBasedAttitudeEstimation}
In this section we introduce the attitude quaternion estimation approach based on the \ac{USQUE} method, proposed by \cite{crassidis2003unscented}. The key strength of this technique is that singular representations of the attitude are avoided by estimating the error between the measured attitude and its predicted value. This error is always assumed to be smaller than $\pi$ rad.
\paragraph{Initialization}
\begin{enumerate}
	\item Consider the attitude error vector $\boldsymbol{\epsilon}_k$ in the state vector $\hat{\boldsymbol{\xi}}_k$. At every iteration, initialize $\hat{\boldsymbol{\epsilon}}_{k-1}^+ = \boldsymbol{0}_3$.
\end{enumerate}

\paragraph{Prediction step}
\begin{enumerate}
\setcounter{enumi}{1}
\item First we generate a set of quaternion sigma points $\hat{\boldsymbol{q}}_{k-1}^{(i)}$, with $i=0, \dots, 2n$.
\par
Select from every sigma point $\hat{\boldsymbol{\xi}}_{k-1}^{(i)}$ in Eq. \ref{ukf:pred:sigma_points} the attitude error part $\hat{\boldsymbol{\epsilon}}_{k-1}^{(i)}$.
Convert the attitude error vector sigma point $\hat{\boldsymbol{\epsilon}}_{k-1}^{(i)}$ in its quaternion representation $\delta \hat{\boldsymbol{q}}_{k-1}^{(i)}$ by using the \acp{MRP}. Obtain the attitude quaternion sigma points $\hat{\boldsymbol{q}}_{k-1}^{(i)}$ by rotating the estimate of the current attitude quaternion $\hat{\boldsymbol{q}}_{k-1}^+$ by the computed delta attitude quaternion sigma points $\delta{\hat{\boldsymbol{q}}}_{k-1}^{(i)}$ as in \ref{eq:attitude_quat:quat_sigma_points}.
\begin{equation} \label{eq:attitude_quat:quat_sigma_points}
\begin{aligned}
	&\delta \hat{\boldsymbol{q}}_{k-1}^{(i)} \leftarrow \textbf{MRP}(\hat{\boldsymbol{\epsilon}}_{k-1}^{(i)})\\
	\hat{\boldsymbol{q}}_{k-1}^{(i)}=&\delta{\hat{\boldsymbol{q}}}_{k-1}^{(i)} \otimes \hat{\boldsymbol{q}}_{k-1}^+\\
	& i = 0, \dots, 2n
\end{aligned}
\end{equation}
Convert the sigma points $\hat{\boldsymbol{\xi}}_{k-1}^{(i)}$ computed in Eq. \ref{ukf:pred:sigma_points} to $\hat{\boldsymbol{x}}_{k-1}^{(i)}$ by replacing  $\hat{\boldsymbol{\epsilon}}_{k-1}^{(i)}$ with the computed $\hat{\boldsymbol{q}}_{k-1}^{(i)}$.
\item Propagate the sigma points through the non-linear system dynamic model, as detailed in Eq. \ref{ukf:pred:propagation}. Retrieve the set of propagated quaternion sigma points $\hat{\boldsymbol{q}}_{k}^{(i)}$ from $\hat{\boldsymbol{x}}_{k}^{(i)}$.
\item Compute the mean and covariance of the propagated quaternion sigma points. For the mean, simply select $\tilde{\boldsymbol{q}}_{k}^{(0)}$ as the average of $\hat{\boldsymbol{q}}_{k}^{(i)}$, $i=0,\dots,2n$ to obtain the predicted attitude quaternion $\hat{\boldsymbol{q}}_{k}^-$.
\begin{equation}
\label{eq:ukf:predicted_attitude_quaternion}
\hat{\boldsymbol{q}}_{k}^- = \hat{\boldsymbol{q}}_{k}^{(0)}
\end{equation}
As an alternative, it is possible to fully compute the average on the quaternion manifold using the method proposed in \cite{Markley2007}. For the covariance, first compute the propagated delta quaternions sigma points $\delta\hat{\boldsymbol{q}}_k^{(i)}$ and obtain a set of delta sigma points $\hat{\boldsymbol{\epsilon}}_k^{(i)}$ by using the \acp{MRP}.
\begin{equation}
	\begin{aligned}
	&\delta \hat{\boldsymbol{q}}_k^{(i)} = \hat{\boldsymbol{q}}_{k}^- \otimes (\hat{\boldsymbol{q}}_k^{(i)})^{-1} \\
	\hat{\boldsymbol{\epsilon}}_k^{(i)} \leftarrow \textbf{MRP}(&\delta \hat{\boldsymbol{q}}_k^{(i)})\\
	&i=0,\dots,2n
	\end{aligned}
\end{equation}
Convert the propagated sigma points $\hat{\boldsymbol{x}}_{k}^{(i)}$ computed in Eq. \ref{ukf:pred:propagation} to $\hat{\boldsymbol{\xi}}_{k}^{(i)}$ by replacing $\hat{\boldsymbol{q}}_k^{(i)}$ with $\hat{\boldsymbol{\epsilon}}_k^{(i)}$.
Compute the predicted error attitude mean $\hat{\boldsymbol{\epsilon}}_k^-$ as in Eq. \ref{ukf_pred_mean} and covariance as in Eq. \ref{ukf_pred_cov}.
\end{enumerate}

\paragraph{Update Step}
\begin{enumerate}
\setcounter{enumi}{4}
\item Given an attitude measurement $\boldsymbol{q}_k^m$, compute the rotation error between $\hat{\boldsymbol{q}}_{k-1}^+$ and $\boldsymbol{q}_k^m$  as in Eq. \ref{ukf:att:meas}. Transform the quaternion $\delta \boldsymbol{q}_k^m$ in the error vector $\boldsymbol{\epsilon}_k^m$
by using the \acp{MRP}.
\begin{equation} \label{ukf:att:meas}
\begin{aligned}
&\delta \boldsymbol{q}_k^m = \hat{\boldsymbol{q}}_{k-1}^+ \otimes {(\boldsymbol{q}_k^m)}^{-1}\\
\boldsymbol{\epsilon}_k^{m} \leftarrow \textbf{MRP}(&\delta \boldsymbol{q}_k^m)\\
\end{aligned}
\end{equation}
Store $\boldsymbol{\epsilon}_k^m$ in $\boldsymbol{z}_k$.
\item Compute the updated state covariance and mean as in Eq. \ref{eq:UpdateStateCovariance} and \ref{eq:UpdateStateMean}.
\item Retrieve $\hat{\boldsymbol{\epsilon}}_k^+$ from $\hat{\boldsymbol{\xi}}_k^+$ and compute the updated error attitude quaternion $\delta \hat{\boldsymbol{q}}_k^+$ from $\hat{\boldsymbol{\epsilon}}_k^+$ using \acp{MRP}. Rotate the predicted attitude quaternion $\hat{\boldsymbol{q}}_k^-$ of $\delta \hat{\boldsymbol{q}}_k^+$ to obtain the current quaternion estimate of the attitude as in Eq. \ref{ukf:upd_step:current_attitude_estimate}.
\begin{equation}
\begin{aligned} \label{ukf:upd_step:current_attitude_estimate}
&\delta \hat{\boldsymbol{q}}_k^+ \leftarrow \textbf{MRP}(\hat{\boldsymbol{\epsilon}}_k^+)\\
\hat{\boldsymbol{q}}_k^+ = \hat{\boldsymbol{q}}_k^- \otimes &\delta \hat{\boldsymbol{q}}_k^+\\
\end{aligned}
\end{equation}
\end{enumerate}

\section{System Model}\label{sec:model}
\label{sec:SystemModel}
In this section we derive a model for $N$ agents connected to a payload via a spherical joint. First we introduce a dynamic model for a rigid payload subject to the actuation force of an arbitrary number of \ac{MAV} connected to it. Then, we derive a model for the control algorithms of a slave agent and for the master agent. These results will be used in Section \ref{sec:robust_tuning} to quantify the model uncertainties and to study the stability and performance of the collaborative transportation strategy subject to the model uncertainties.
\subsection{Mechanical model of $N$ agents connected to a payload via a spherical joint}
\paragraph{Model assumptions}
We consider the payload to be a rigid body with mass $m_p$ and inertia tensor $\prescript{}{}{\boldsymbol{J}}_p$. We remark that the position of the payload is defined with respect to an inertial reference frame $W$, such that $z$-axis points upwards (gravity is negative), and a non-inertial reference frame $P$ attached to the \ac{CoG} of the object. The pose of the payload in space is defined via a vector $\prescript{}{W}{\boldsymbol{p}}_{WP}$ and via a rotation matrix $\prescript{}{W}{\boldsymbol{R}}_{WP}$ which defines the rotation from $P$ to $W$ expressed in the frame $W$.
\par We model each of the $N$ robots connected to the object as a rigid body of mass $m_i$ with a diagonal inertia tensor $\prescript{}{}{\boldsymbol{J}}_i$, where $i = 1\dots N$ refers to the \textit{$i$-th} agent. Only for modeling purpose, we introduce the following assumptions:
\begin{itemize}
	\item the translation of every \textit{$i$-th} vehicle's reference frame $B_i$ with respect to the body frame of the payload $P$ is known and expressed as $\prescript{}{P}{\boldsymbol{r}}_{PB_{i}}$
	\item the inertial reference frames $I_i$ of every agent coincides with the inertial reference frame $W$ of the payload.
	\item the rotation from the body frame $B_i$ of the \textit{$i$-th} agent and the payload body frame $P$ is expressed by the rotation matrix $\prescript{}{P}{\boldsymbol{R}}_{PB_{i}}$
	\end{itemize}
\par
\begin{figure}
	\centering
	\includegraphics[width=0.5\textwidth]{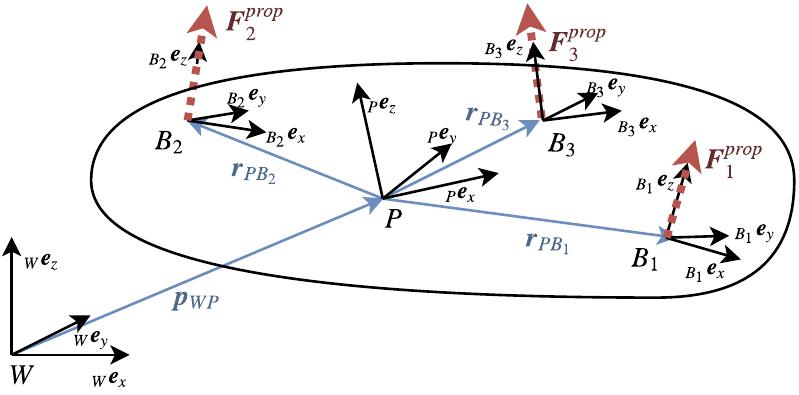}
	\caption{Representation of the coordinate frames used for modeling. The three main frames are the inertial frame $W$, the body frame attached in the \acf{CoG} of the payload $P$ and the frame connected in the \ac{CoG} of every agent $B_i$. The vector $\prescript{}{}{\boldsymbol{p}_{WP}}$ represents the position of the payload. The vectors $\prescript{}{}{\boldsymbol{r}_{PB_{i}}}$ represent the position of each agent in the payload frame and are assumed to be constant. The thrust force produced by every robot is indicated as $\prescript{}{}{\boldsymbol{F}_i^\textit{prop}}$. The inertial reference frame of each agent $I_i$ is not indicated since we have assumed that coincides with $W$.}
	\label{fig:CoordinateSysModeling}
\end{figure}

 Each \textit{$i$-th} vehicle connected to the payload can produce a force input $\prescript{}{B_{i}}{\boldsymbol{F}_{i}^{\textit{prop}}}$, which is expressed in the vehicle reference frame $B_i$. In addition, we assume that every robot is attached to the payload via an ideal spherical joint (no frictional torques and no kinematic constraints on the rotation of the connected bodies) whose revolution point corresponds to the \ac{CoG} of the \ac{MAV}, so that the torque produced by the vehicle cannot be transmitted to the structure. An example of the system described can be found in Figure \ref{fig:CoordinateSysModeling}.
Following these assumptions we can derive the kinematic equations of the system, as well as the rotational and translational dynamics.
\paragraph{System kinematics}
The translational kinematic model for the transported structure is defined by $\prescript{}{W}{\boldsymbol{p}}_{WP}$, $\prescript{}{W}{\boldsymbol{v}}_{WP}$ and $\prescript{}{W}{\dot{\boldsymbol{v}}_{WP}}$. Its rotational kinematic is given by $\prescript{}{W}{\boldsymbol{R}}_{WP}$ and $\prescript{}{P}{\boldsymbol{\omega}}$, $\prescript{}{P}{\dot{\boldsymbol{\omega}}}$.
\par
The translational kinematic $\prescript{}{W}{{\boldsymbol{p}}_{WB_{i}}}$, $\prescript{}{W}{\boldsymbol{v}}_{WB_{i}}$ and $\prescript{}{W}{\dot{\boldsymbol{v}}_{WB_{i}}}$ for the \textit{$i$-th} slave agent  expressed in $W$ can be obtained from the kinematic of the payload as:
\begin{equation}
\prescript{}{W}{\boldsymbol{p}}_{WB_{i}} = \prescript{}{W}{\boldsymbol{p}}_{WP}+\prescript{}{W}{\boldsymbol{R}}_{WP}\prescript{}{P}{\boldsymbol{r}}_{PB_{i}}
\end{equation}
\begin{equation}
\prescript{}{W}{\boldsymbol{v}}_{WB_{i}} = \prescript{}{W}{\boldsymbol{v}}_{WP}+ \prescript{}{P}{\boldsymbol{\omega}} \times  \prescript{}{P}{\boldsymbol{r}}_{PB_i}
\end{equation}
\begin{equation}
\prescript{}{W}{\dot{\boldsymbol{v}}_{WB_{i}}} =\prescript{}{W}{\dot{\boldsymbol{v}}}_{WP}+ \prescript{}{P}{\dot{\boldsymbol{\omega}}} \times  \prescript{}{P}{\boldsymbol{r}}_{PB_{i}} +  \prescript{}{P}{\boldsymbol{\omega}} \times ( \prescript{}{P}{\boldsymbol{\omega}} \times  \prescript{}{P}{\boldsymbol{r}}_{PB_{i}})
\end{equation}
The rotational kinematic of each agent is independent from the kinematic of the payload thanks to the connection via the spherical joint.
\paragraph{Translational dynamics} The translational dynamic of the payload can be expressed in the inertial reference $W$ as:
\begin{equation}
\prescript{}{W}{\dot{\boldsymbol{v}}}_{WP} = \frac{1}{m_{\text{sys}}}{\prescript{}{W}{\bm{R}}_{WP}}\Big(
\prescript{}{P}{\boldsymbol{F}^{\textit{agents}}}
- \prescript{}{P}{\boldsymbol{F}^{\textit{drag}}}
\Big)
- \prescript{}{W}{\boldsymbol{g}}.
\end{equation}
Where we have defined $\prescript{}{P}{\boldsymbol{F}^{\textit{agents}}}$ as the total force applied by the agents on the payload, expressed in $P$, obtained from
\begin{equation}
\prescript{}{P}{\boldsymbol{F}^{\textit{agents}}} = \sum_{i = 1}^{N}{\prescript{}{P}{\bm{R}}_{PB_{i}}}\prescript{}{B_{i}}{\boldsymbol{F}}_{i}^{\textit{prop}}
\end{equation}
and  $\prescript{}{P}{\boldsymbol{F}^{\textit{drag}}}$ is a drag force proportional to the velocity of the object, which depends on the aerodynamic properties of the payload.
We additionally define $m_{\text{sys}}$ as the total mass of the system, given by
\begin{equation}
m_{\text{sys}} = m_p +  \sum_{i=1}^{N}{m_i}.
\end{equation}
\paragraph{Rotational dynamics}
The rotational dynamic of the transported structure can be expressed in the body reference frame of the payload $P$ as
\begin{equation}
\prescript{}{P}{\dot{\boldsymbol{\omega}}} = \prescript{}{P}{\boldsymbol{J}_{\text{sys}}}^{-1}
\Big(
\prescript{}{P}{\boldsymbol{M}}^{\textit{agents}}
- \prescript{}{P}{\boldsymbol{\omega }} \times \prescript{}{P}{\boldsymbol{J}_{\text{sys}}} \prescript{}{P}{\boldsymbol{\omega}}
- \prescript{}{P}{\boldsymbol{M}}^{\textit{drag}}
\Big)
\end{equation}
where $\boldsymbol{M}^{\textit{agents}}$ represents the total torque produced by the agents on the payload. It is obtained from
\begin{equation}
\prescript{}{P}{\boldsymbol{M}}^{\textit{agents}} = \sum_{i = 1}^{N}{
\prescript{}{P}{\boldsymbol{r}}_{PB_{i}} \times (\prescript{}{P}{\bm{R}}_{PB_{i}}\prescript{}{B_{i}}{\boldsymbol{F}_{i}^{\textit{prop}}})}
\end{equation}
and $\boldsymbol{M}^{\textit{drag}}$ is a drag torque which takes into account aerodynamic effects.
 $\boldsymbol{J}_{\text{sys}}$ represents the total inertia of the system and depend on the inertia of the payload and the mass and position of the connected \ac{MAV} agents. As a remark, we observe that the inertia of the single robots $\boldsymbol{J}_i$ is not taken into account, since the spherical joint decouples the rotational dynamic of the object from the rotational dynamic of the robots.
\subsection{Agent's Low-Level Control Architecture}
\emph{Notation declaration} In this Subsection we drop the notation $i$ to indicate the \textit{$i$-th} agent since we only consider a single, generic \ac{MAV}.
\par 
Every agent acts on the payload via the thrust force $\boldsymbol{F}^{\textit{prop}}$ produced by the propellers. This force, by assuming that the axis of rotation of all the propellers is parallel to $\prescript{}{B}{\boldsymbol{e}}_{z}$, corresponds to:
\begin{equation}
\prescript{}{B}{\boldsymbol{F}}^{\textit{prop}} = \prescript{}{B}{\boldsymbol{e}}_{z} F_{\text{prop}}
\end{equation}
where $F_{\text{prop}}$ can be directly obtained from the rotation speed of the propellers.
If expressed in the payload frame $P$, the thrust force can be written as
\begin{equation}
	\prescript{}{P}{\boldsymbol{F}}^{\textit{prop}} = \prescript{}{P}{\bm{R}_{PB}}\prescript{}{B}{\boldsymbol{e}}_{z} F_{\text{prop}}.
	\label{eq:model:thrust_force_in_world_frame}
\end{equation}
From Eq. \ref{eq:model:thrust_force_in_world_frame} it si possible to observe that the direction of the force acting on the payload can only be controller by changing the attitude $\prescript{}{P}{\bm{R}_{PB}}$ of the vehicle. We make the assumption that the \ac{MAV} are mechanically free to assume arbitrary attitude thanks to the spherical joint.
As a consequence, the dynamics of the actuation force produced by each agent on the payload depend on the control algorithms of each robot.
 \par In order to describe the dynamics of $\prescript{}{P}{\boldsymbol{F}}^{\textit{prop}}$, we consider a simplified low-level control architecture common to master and slave agents, constituted by the following submodules:
\begin{inparaenum}[(a)]
	\item position controller
	\item attitude controller and \ac{MAV}'s attitude dynamics.
\end{inparaenum}
We assume to have perfect knowledge of the state (pose, velocity and angular velocity) of each agent.
\paragraph{Position controller}
We assume that the position of each agent is controlled by a thrust vector controller. Such a controller provides a thrust command the vehicle expressed in the inertial reference frame $W$, given a position and a velocity reference expressed in the same frame. The thrust command is then additionally mapped in an attitude reference for the cascaded attitude controller. The controller is designed by assuming that the dynamic of the \ac{MAV} can be expressed in an inertial reference as
\begin{equation}
\prescript{}{W}{\dot{\boldsymbol{v}}}_{WB} = \frac{1}{m}
\prescript{}{W}{\boldsymbol{F}^{\textit{cmd}}}
- \prescript{}{W}{\boldsymbol{g}}
\end{equation}
where $\prescript{}{W}{\boldsymbol{F}^{\textit{cmd}}}$ is the output of the controller.
\par The control output $\prescript{}{W}{\boldsymbol{F}^{\textit{cmd}}}$ is computed using three separate PD controllers, one for each axis of $W$:
\begin{multline} \label{eq:NominalTfPositionController}
\prescript{}{W}{\boldsymbol{F}^{\textit{cmd}}} = \boldsymbol{K}_{P}
(\prescript{}{W}{\boldsymbol{\Lambda}_{r}} - \prescript{}{W}{\boldsymbol{p}}_{WB})
+ \\
\boldsymbol{K}_{D}
(\prescript{}{W}{\dot{\boldsymbol{\Lambda}}_{r}}-\prescript{}{W}{\boldsymbol{v}_{WB}}) + m\prescript{}{W}{\boldsymbol{g}}
\end{multline}
where the positive, diagonal matrices $\boldsymbol{K}_{P} = \diag(K_{P_{x}}, K_{P_{y}}, K_{P_{z}})$ and   $\boldsymbol{K}_{D} = \diag(K_{D_{x}}, K_{D_{y}}, K_{D_{z}})$ are tuning parameters. The vector $m\prescript{}{W}{\boldsymbol{g}}$ represent the feed-forward compensation of the gravity force.

\par Given $\prescript{}{W}{\boldsymbol{F}^{\textit{cmd}}}$ and a desired heading reference $\psi_{r}$, we can retrieve  the roll $\phi_{\text{cmd}}$, pitch $\theta_{\text{cmd}}$ and thrust $\prescript{}{B}{\boldsymbol{e}}_{z}F_{\text{cmd}}$ commands for the attitude controller from:
\begin{equation}
F_{\text{cmd}} = \norm{\prescript{}{W}{\boldsymbol{F}^{\textit{cmd}}}}
\end{equation}
\begin{equation}
\bm{R}_{W}^T(\prescript{}{B}{\boldsymbol{e}_z},\psi)\frac{\prescript{}{W}{\boldsymbol{F}^{\textit{cmd}}}}{F_{\text{cmd}}}=
\begin{bmatrix}
\sin(\theta_{\text{cmd}})\cos(\phi_{\text{cmd}})\\
-\sin(\phi_{\text{cmd}}) \\
\cos(\theta_{\text{cmd}})\cos(\phi_{\text{cmd}})
\label{eq:RPCommandAllocation}
\end{bmatrix}
\end{equation}
where $\boldsymbol{R}_{W}^T(\prescript{}{B}{\boldsymbol{e}_z},\psi)$ represent the rotation matrix of the current attitude $\psi$ around the \textit{$z$-axis} expressed in $W$ reference frame.

\paragraph{Attitude controller and \ac{MAV}  attitude dynamics}
The attitude of the vehicle is controlled by an inner feedback loop, shaped so that it responds to a roll, pitch or yaw command as a second order system. For example, given the tuning constants $K_{P_{\phi}}$ and $K_{D_{\phi}}$, the roll dynamic can be expressed as
\begin{equation}
\ddot{\phi} = \frac{1}{J_{xx}}(K_{P_{\phi}}(\phi_{\text{cmd}}-\phi)+K_{D_{\phi}}(-\dot{\phi}))
\label{eq:AttitudeDynamic}
\end{equation}
where $\boldsymbol{J}_{xx}$ corresponds to the entry $(1,1)$ of the diagonal inertia tensor $\boldsymbol{J}$.
We additionally assume that $\abs{\phi_{\text{cmd}}}$ and $\abs{\theta_{\text{cmd}}}$ are upper bounded, respectively, by the tuning parameters $\phi_{\text{cmd}}^{\textit{max}}$ and $\theta_{\text{cmd}}^{max}$. Last, we assume that the commanded yaw is always constant.

\paragraph{Performance limitations}
The kinematic and dynamic of the attitude limit the maximum performance achievable in terms of the maximum produced thrust force $\prescript{}{W}{\boldsymbol{F}}^{\textit{prop}}$ and its rate of change $\prescript{}{W}{\dot{\boldsymbol{F}}}^{\textit{prop}}$.
\par
The limit on $\prescript{}{W}{\boldsymbol{F}}^{\textit{prop}}$ can be derive from Equation \ref{eq:RPCommandAllocation}, by assuming, without loss of generality, a constant yaw angle $\psi$ equal to zero:
\begin{equation}
\abs{\prescript{}{W}{F^{\textit{prop}}_{\text{max}, x}}} \leq \sin(\phi_{\text{cmd}}^{\textit{max}}) F_{\text{prop}}^{\textit{max}}
\end{equation}
\begin{equation}
\abs{\prescript{}{W}{F^{\textit{prop}}_{\text{max}, y}}} \leq \sin(\theta_{\text{cmd}}^{max}) F_{\text{prop}}^{\textit{max}}
\end{equation}
\begin{equation}
\norm{\prescript{}{W}{\boldsymbol{F}^{\textit{prop}}_{max}}} = F_{\text{prop}}^{\textit{max}}
\end{equation}
where $F_{\text{prop}}^{\textit{max}}$ is the maximum thrust produced by the robot on $\prescript{}{B}{\boldsymbol{e}}_{z}$.
\par
Last, by assuming Eq. \ref{eq:AttitudeDynamic} to be shaped so that the system is critically damped with time constant $\tau_{\text{att}}$, we can approximate the dynamic of $\prescript{}{W}{F^{\textit{prop}}_{x}}$, $\prescript{}{W}{F^{\textit{prop}}_{y}}$ and $\prescript{}{W}{F^{\textit{prop}}_{z}}$ to be given by:
\begin{equation}
\prescript{}{W}{\dot{F}}^{\textit{prop}}_{j} = \frac{1}{\tau_{\text{att}}}(\prescript{}{W}{F}_{j}^{\textit{cmd}}-\prescript{}{W}{F}^{\textit{prop}}_{j})
\end{equation}
with $j = x,y,z$.


\section{Robust Tuning}\label{sec:robust_tuning}
The choice of the apparent physical parameters of a team of \acp{MAV} when they are mechanically coupled to the payload is not obvious and require laborious tuning procedure, especially if we aim to guarantee stability and acceptable performance. In this section we discuss the robust stability analysis for the system of a team of aerial robots collaboratively transporting a payload. Moreover, we define a target performance on the sensitivity transfer function and search for tuning parameters that guarantee the achievement of the desired performance despite the presence of uncertainties. First, we identify sources of uncertainty and quantify it. Afterwards, performance requirements are defined on the forces generated by each \ac{MAV}. Finally, the robust stability and robust performance margins are calculated for a grid of controller parameters. The optimal parameters that guarantee performance and stability in worst case scenario are chosen as tuning parameters.  

\subsection{Uncertainty Modeling}
Robust control theory framework allows us to explicitly take into account mismatching and uncertainties between a plant and its dynamic model. The sources of uncertainty that we take into consideration in this work can be grouped in two categories:
\begin{itemize}
	\item \textbf{Parametric (real) uncertainty}: due to their nature, some model parameters can assume different values, bounded within a region. 
	\item \textbf{Dynamic (frequency-dependent) uncertainty}: uncertainty caused  by missing dynamics from the considered nominal model. This uncertainty usually increases with the frequency and will always exceed 100\% at some frequency.
\end{itemize}
Taking into account uncertainty in the plant model has several advantages, for example it allows us to:
\begin{itemize}
	\item derive a more generic model which includes parameters that cannot be known a priori, such as the mass of the payload
	\item work with a simpler (lower-order) nominal model, where the neglected dynamics are represented as uncertainties
	\item simplify the description of highly non-linear dynamics, difficult to capture due to the change in the operating conditions.
\end{itemize}
Uncertainties can be quantified by means of physical considerations or system identification techniques. We use the first approach to establish the bounds of parametric uncertainty. The second method is instead used to define frequency-dependent uncertainties. 
\par The frequency-dependent uncertainty is modeled using the so called \textit{multiplicative uncertainty model}. This method allows us to define a relative uncertainty with respect to the nominal plant model. The magnitude of the relative uncertainty is estimated by comparing the nominal and actual plant frequency responses. The actual plant frequency response can be obtained using system identification techniques or using a sophisticated high fidelity model. 
\paragraph{Uncertain payload modeling}
We assume that the mass $m_p$ and the inertia of the payload $\boldsymbol{J}_p$ can be subject to parametric uncertainty.
\par For the robust stability and performance analysis detailed we will assume that $m_p \in [m_{p, min}, m_{p, max}]$, where $ m_{p, min} $ and $ m_{p, max} $ are an estimate of the minimum and maximum expected payload mass.
\par The uncertainty about the diagonal inertial tensor $\boldsymbol{J}_p$  can be modeled as a relative uncertainty with respect to a nominal value, which depends on the geometric and physical properties of the payload. For the robust stability and performance analysis we will assume  $10\%$ uncertainty on each of its diagonal entries with respect to a specified nominal inertia.
\paragraph{Uncertainty quantification of a single agent}
We consider each slave \ac{MAV} agent to be subject to two main sources of frequency-dependent uncertainty, due to the neglected dynamics of:
\begin{itemize}
	\item force estimator, whose dynamics can vary according to the estimation technique employed and the tuning parameters
	\item position controller. Our experimental platform employs a non-linear \ac{MPC}.
\end{itemize}
These uncertainties are quantified by means of system identification using experimental data or high-fidelity simulation tools which implement the control algorithms that we intend to evaluate.
\par
\begin{figure}
	\centering
	\includegraphics[width=0.45\textwidth]{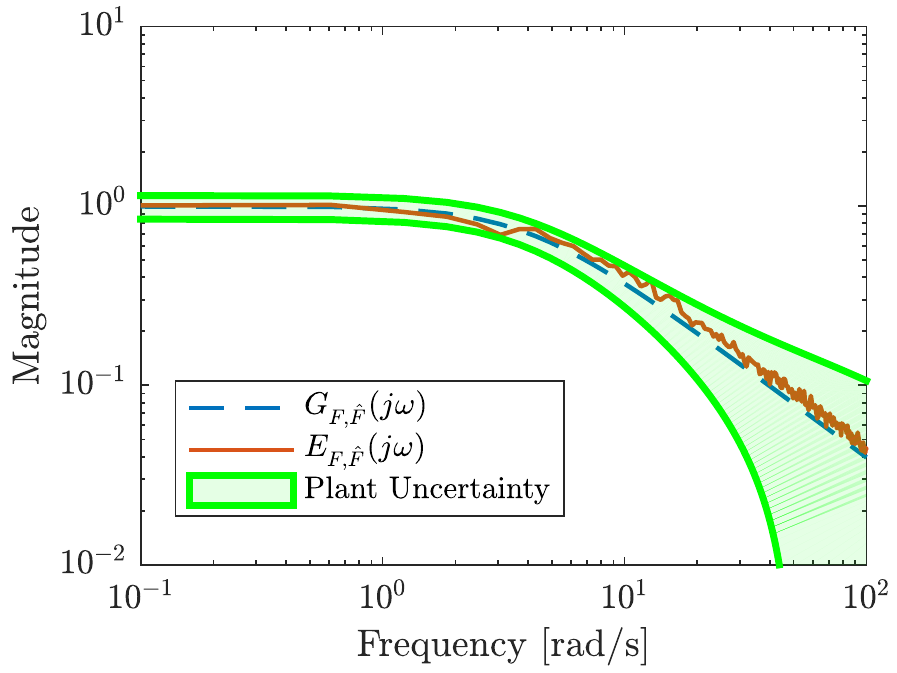}
	\caption{Modelling of the force estimator based on the Disturbance Observer.}
	\label{fig:DoUncertaintyId}
\end{figure}

 We then model the uncertainty as follows.
First we denote the actual transfer function of the force estimator as $G_{\text{est}}(j\omega)$, and by exploiting the multiplicative uncertainty model, we have:
\begin{equation} \label{eq:MultiplicativeUncertaintyForceEstimator}
G_{\text{est}}(j\omega) = (1+w_{\text{est}}(j\omega)\Delta_{\text{est}}(j\omega))G_{\text{est}}^{\text{nom}}(j\omega)
\end{equation}
where $G_{\text{est}}^{\text{nom}}(j\omega)$ corresponds to the transfer function derived from equation \ref{eq:IdealTfForceEstimator}. The weighting function $w_{\text{est}}(j\omega)$ is chosen so that 
\begin{equation}
w_{\text{est}}(j\omega) \geq \Bigg| \frac{G_{\text{est}}(j\omega) - G_{\text{est}}^{\text{nom}}(j\omega)}{G_{\text{est}}^{\text{nom}}(j\omega)}\Bigg|
\end{equation}
In this way any function $\Delta_{\text{est}}(j\omega)$ such that $|\Delta_{\text{est}}(j\omega)|<1$ will satisfy \ref{eq:MultiplicativeUncertaintyForceEstimator}, for any frequency $\omega$. 
$G_{\text{est}}(j\omega)$ is replaced with the transfer function obtained via system identification from the high-fidelity simulator \textbf{RotorS} developed by \cite{Furrer2016}, where we implemented model of the used \ac{MAV} experimental platform with the considered force estimators. An example of uncertainty modeling for the force estimator is represented in Figure \ref{fig:DoUncertaintyId}.
\par
\begin{figure}
	\centering
	\includegraphics[width=0.3\textwidth]{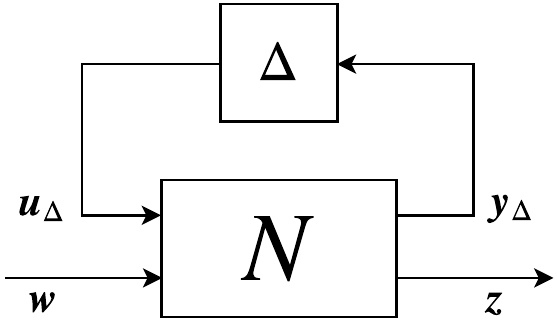}
	\caption{\textit{N-$\Delta$} block structure for analysis of \acf{MIMO} systems with uncertainty.}
	\label{fig:ModelInCanonicalForm}
\end{figure}

 We proceed similarly for the uncertainty identification of the position controller, where we wish to establish the relationship:
\begin{equation}
G_{\text{MPC}}(j\omega) = (1+w_{\text{MPC}}(j\omega)\Delta_{\text{MPC}}(j\omega))G_{\text{MPC}}^{\text{nom}}(j\omega)
\end{equation}
The transfer function $G_{\text{MPC}}(j\omega)$ can be obtained experimentally by sending position references to the \ac{MAV} agent by computing the produced thrust force $\prescript{}{B}{\boldsymbol{T}}_i$ from: 
\begin{equation}
\prescript{}{B}{\boldsymbol{T}}_i = 
\begin{bmatrix}
\sin(\hat{\theta})\cos(\hat{\phi})\\
-\sin(\hat{\phi}) \\
\cos(\hat{\theta})\cos(\hat{\phi})
\end{bmatrix}
U_{4,i}
\end{equation}
where $\hat{\phi}$, $\hat{\theta}$ are given by the state estimator.
The nominal transfer function of the position controller ca $G_{\text{MPC}}^{\text{nom}}(j\omega)$ is obtained from \ref{eq:NominalTfPositionController}, where $kp$ and $kp$ are again identified by means of system. identification.
\paragraph{Model in canonical form}
As a last modeling step, we rewrite out model according to the generalized \textit{N-$\Delta$} plant structure, as represented in Figure \ref{fig:ModelInCanonicalForm}. Such a structure constitutes the canonical representation for analysis of a MIMO system and allows to take into account of uncertainty..  
\par The block $\Delta$ contains a \textit{block-diagonal} matrix that includes all the possible perturbations to the system and $\| \Delta \|_{\infty} \leq 1$. The vector $\boldsymbol{w}$ represents the input to the system and correspond to a velocity command for the master agent. The quantity $\boldsymbol{z}$ corresponds to the monitored system output and contains the performance metrics. The signals $\boldsymbol{u}_{\Delta}$ and $\boldsymbol{y}_{\Delta}$ include the perturbation due to uncertainty, coming from the following sources: 
\begin{inparaenum}[(a)]
	\item payload mass,
	\item payload inertia,
	\item position and attitude controller of every agent, and 
	\item force estimator for every slave agent.
\end{inparaenum}

\subsection{Performance Requirement}
The performance requirements are specified in frequency domain. We mainly focus on limiting the lateral forces generated by each agent to avoid infeasible desired tilting angles that might cause instability. Robust performance is achieved if the maximum singular value of the weighted closed-loop uncertain transfer function between the system input signal $ \boldsymbol{w} $ and performance metric $ \boldsymbol{z} $ is below unity. i.e. the following inequality is satisfied:

\begin{equation}
\| T \|_{\infty} = \max_{\omega\in\mathbf{R}}{\bar{\sigma}(T(j\omega))}\leq 1
\end{equation}
\noindent where $ T $ is the transfer function between $ \boldsymbol{w} $ and $ \boldsymbol{z} $. The frequency domain weighting on the force along the $ j-th $ axis generated by each agent is given by $ \bm{w}_{p, F}(j \omega ) $ as shown in Figure~\ref{fig:weight_force_performance}. Note that we shape the weight such that larges forces are allowed during transient in the range between $ 0.01  $ \si{\radian/\second} and $  0.067 $ \si{\radian/\second}. 

\begin{figure}
	\centering
	\includegraphics[width=0.48\textwidth]{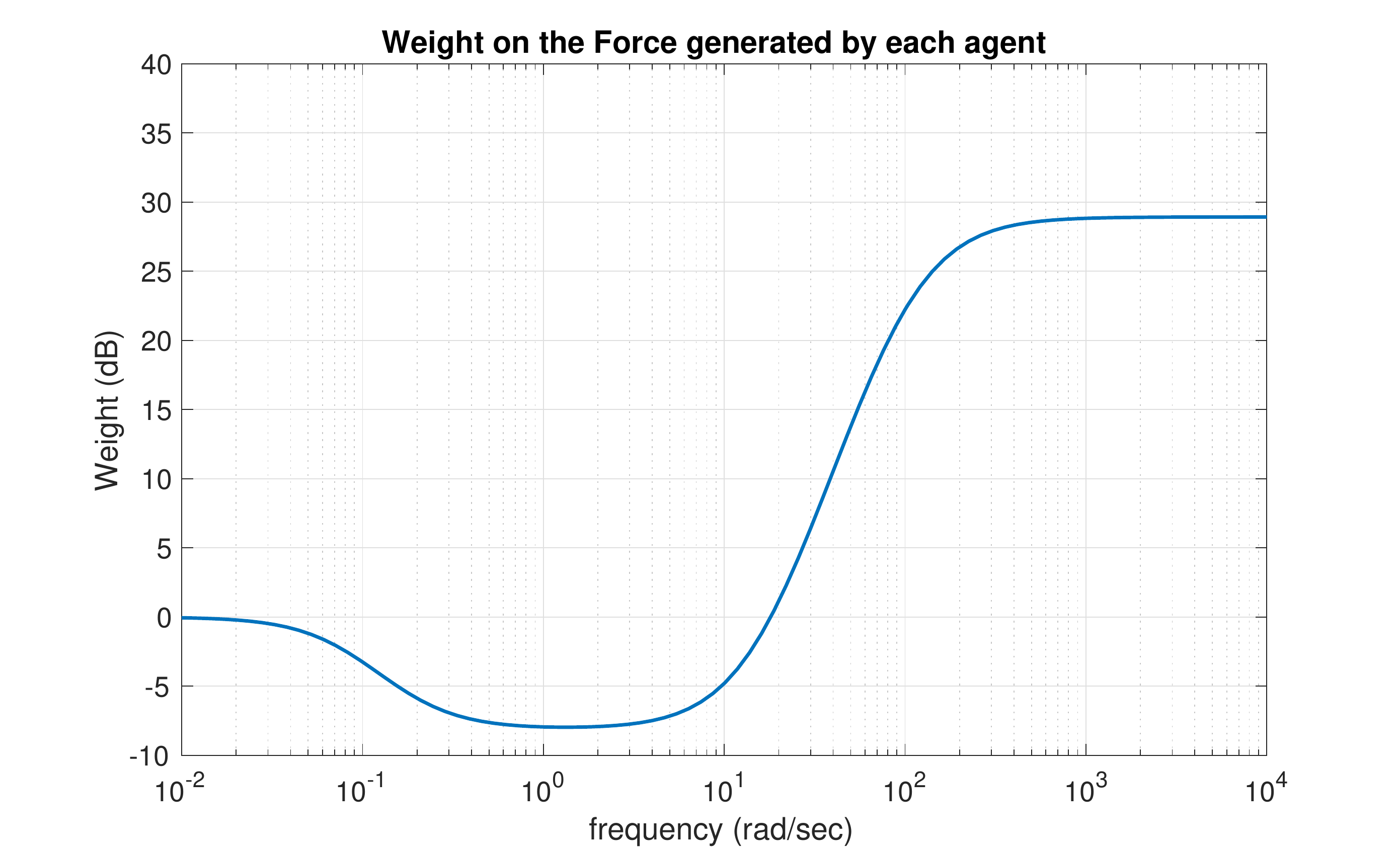}
	\caption{Weighting on the force generated by each agent.}
	\label{fig:weight_force_performance}
\end{figure}
\subsection{Robust Stability and Performance Margins}
Using $ \mu $ analysis, the robust stability and robust performance margins of the closed loop system can be characterized for a given admittance controller parameters. A grid of parameters are explored and the margins are calculated for each point on the parameters grid. Given that the system has mixed real and complex uncertainty, the robust stability and performance margins are calculated from the structured singular value of the closed loop system. The amount of perturbation that the system can be robust against is determined by the peak in frequency domain of the structured singular value \cite{balas1993mu}. The robustness margins are defined as the inverse of the structured singular value of the system. A robustness margin greater than unity indicate that the system is stable for all possible perturbations.

\section{Evaluation}\label{sec:evaluation}
\label{sec:Evaluation}
\subsection{System description}
\label{subsec:SystemDescription}
\paragraph{Hardware}
The \acp{MAV} used for our experiments are the hexa-copters \textbf{AscTec Neo}, from \cite{AsctecWebsite}, equipped with an on-board computer based on a \textbf{quad-core 2.1 GHz Intel i7} processor and \textbf{8 GB of RAM}. The state estimation system is either based on an external \ac{MCS}, or by the on-board \ac{VI} Navigation System developed by the Autonomous Systems Lab at ETHZ and Skybotix AG., as described in \cite{J.Nikolic2014}.
\par Each \ac{MAV} is equipped with a magnetic gripper. The gripper is designed so that it can retract to ease the landing of the hexa-copter and extend to ease the transportation of the payload, without interfering with the landing gear. The magnetic plate is mounted on the stem through a spherical joint. More details about the magnetic gripper can be found in the work of \cite{bahnemann2017decentralized}.

\paragraph{Software}
The admittance controller and the force estimators described are implemented using \textbf{C++} and \textbf{ROS}. All the control algorithms run on-board to avoid issues related to wireless communication. The multi-sensor fusion and state estimation algorithm used for the \ac{VI} Navigation System is based on the \ac{ROVIO} framework developed by \cite{Bloesch2015a}.
\subsection{Model validation}
\begin{figure*}
	\begin{minipage}{0.70\textwidth}
		\centering
		\raisebox{-0.5\height}{
		\includegraphics[width=\textwidth]{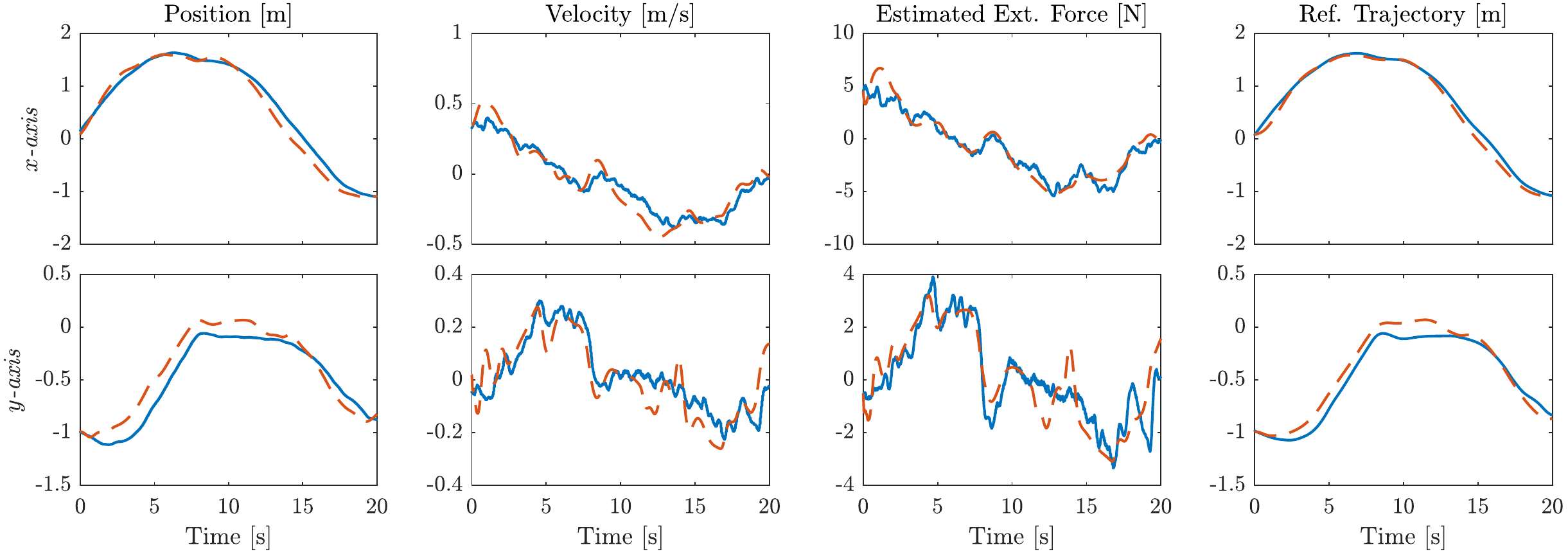}
		}
		\refstepcounter{figure}
		\label{fig:ModelValidation2Mavs}
	\end{minipage}
	\hfill
	\begin{minipage}{0.30\textwidth}
		\centering
		\raisebox{-0.5\height}{
			\scalebox{0.75}{
			\begin{tabular}{|r|c|c|}
				\hline
				Model Output & RMS Error & Unit \\
				\hline \hline
				Position \textit{$x$-axis}& $0.077$ & \si{\meter}  \\ \hline
				Position \textit{$y$-axis}& $0.072$ & \si{\meter}  \\ \hline
				Velocity \textit{$x$-axis}& $0.147$ & \si{\meter \per \second}  \\ \hline
				Velocity \textit{$y$-axis}& $0.139$ & \si{\meter \per \second}  \\ \hline
				Est. Force \textit{$x$-axis}& $0.888$ & \si{\newton} \\ \hline
				Est. Force \textit{$y$-axis}& $0.790$ & \si{\newton} \\ \hline
				Ref. Traj. \textit{$x$-axis} &$0.099$  & \si{\meter} \\ \hline
				Ref. Traj. \textit{$y$-axis} & $0.097$  & \si{\meter} \\ \hline
			\end{tabular}
		}}
		\refstepcounter{table}
		\label{tab:ModelValidation2Mavs}	
	\end{minipage}
	\setcounter{figure}{\thefigure - 1} 
	\setcounter{table}{\thetable - 1} 
	\captionlistentry[table]{\label{tab:ModelValidation2Mavs}}
	\captionsetup{labelformat=andtable}
	\caption{Validation of the collaborative transportation model. The test is performed by comparing the output of the model with the output of an experiment where two \acp{MAV} transport a beam of wood, with mass equal to $1.8$ kg and $1.5$ m long. The continuous line represents the measurements from the experiment, while the dashed line represents the output from the proposed model. All the plots refers to the slave agent. The attached table contains the \acf{RMS} Error between the measured output and the model prediction.}
\end{figure*}


We now compare the analytical model presented in Section \ref{sec:SystemModel} with data experimentally collected. These results allow us to verify if the main dynamics of our system are properly captured and help us to validate the tuning assumptions for the nominal model parameters.
\par The validation experiment is conducted by transporting a $1.8$ kg beam of wood, $1.5$ m long, at a constant altitude of $1.2$ m. The transportation is performed by employing two agents, the first acting as a master and the second as a slave. The tuning parameters of the slave agent are set via the analysis detailed in Section \ref{sec:robust_tuning} and are chosen among the robustly stable parameters which guarantee maximum robust performance margin. The state estimator used  is based on a \ac{MCS} because allows us to simplify the validation since it provides the same inertial reference frame for both the vehicles.
\par Our validation test consists of feeding the same input reference trajectory to the modeled and to the real master agent and by comparing the output of the real system and the model. By output we define the following quantities related to the slave \ac{MAV}: 
\begin{inparaenum}[(a)]
	\item position,
	\item velocity,
	\item estimated external force and
	\item reference trajectory from the admittance controller.
	\end{inparaenum}
The initial state of the model is matched with the initial state of the experimental setup, and the nominal mass and inertia of the modeled payload are matched with the one used for the experiment. 
\par A comparison between the output of the real system and our proposed model is represented in Figure \ref{fig:ModelValidation2Mavs}. The \ac{RMS} Error of our validation is contained in Table \ref{tab:ModelValidation2Mavs}.

\subsection{Robust performance and robust stability analysis results}
\begin{figure*}
	\centering
	\includegraphics[width=\textwidth]{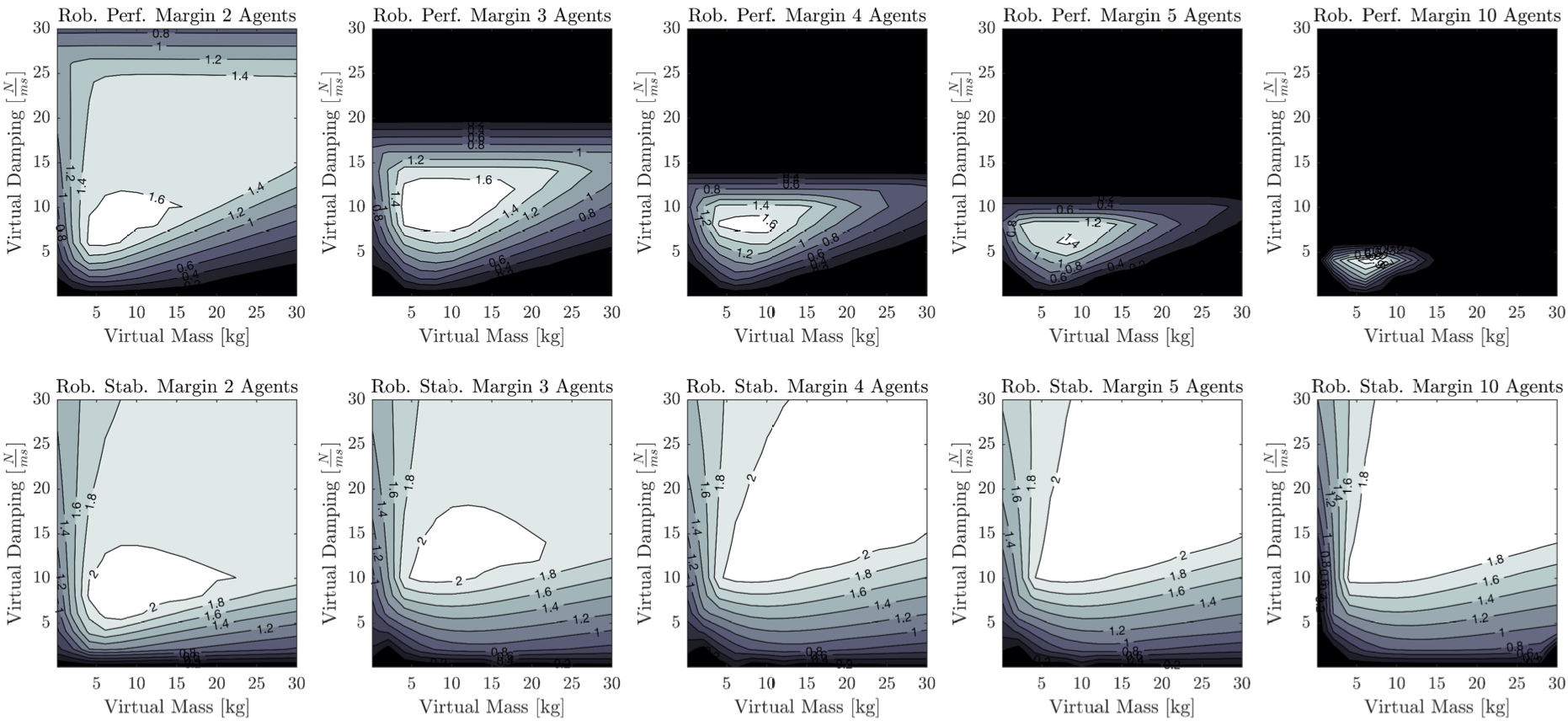}
	\caption{Result of the robust performance and robust stability analysis for a different number of agents and different tuning of the admittance controller. The first row represent the robust performance margin as a function of the virtual mass and the virtual damping of each slave agent. The second row represent the robust stability margin as a function of the virtual mass and virtual damping of the admittance controller of each slave agent.}
	\label{fig:RsRpAnalysisResult}
\end{figure*}
In this section we present the result for the robust stability and robust performance analysis of the collaborative transportation system, based on the model developed and validated in the previous sections. We remark some critical modeling assumptions and the step used to perform the analysis.
\par The modeling parameters used are collected in Table \ref{tab:ParametersRsRpAnalysis}, and are related to the hardware described in Subsection \ref{subsec:SystemDescription}. In addition, here we remark the main modeling assumptions used for the analysis:
\begin{itemize}
	\item All the slave agents have the same nominal model and use the same values of the tuning parameters for the admittance controller.
	\item The tuning parameters of the admittance controller are defined by the virtual mass $\mathcal{M}_x = \mathcal{M}_y = \mathcal{M}$ and the virtual damping $\mathcal{C}_x$ =  $\mathcal{C}_y = \mathcal{C}$ for the \textit{$\prescript{}{W}{\boldsymbol{e}}_x{\boldsymbol{e}}_y$-plane} in the inertial reference frame. The value of the virtual damping $\mathcal{C}$ is chosen from the set $S_{\mathcal{C}} := [0, 30] \subset \Re$ expressed in $\si{\newton \per \meter \per \second}$, while the virtual mass $\mathcal{M}$ belongs to the set $S_{\mathcal{M}} := [0, 30] \subset \Re$ expressed in $\si{\kilogram}$. 
	\item The transported object is assumed to be shaped as a regular polygon with $n$ sides, where $n$ corresponds to the total number of agents. We assume the length of each side to be fixed to $1.2$ m, which represents the minimum distance achievable between two agents given our hardware platform. The nominal mass of the payload $m_p$ is set equal to $m_p = 1.5\bar{m}$, where $\bar{m}$ is the maximum payload capacity of one agent. The total inertia of the system is computed by only taking into account the inertial effect of the mass of the agents rigidly connected to the payload.
\end{itemize}
\par Given a desired number of agents $n$, the robust stability and performance margins are computed via the following steps. First we grid-sample the tuning space of the admittance controller, given by $S = S_\mathcal{M} \times S_\mathcal{C}$. For every point $(\mathcal{M}, \mathcal{C}) \in S$, we linearize the system at rest, which means that all the agents and the payload have zero velocity and zero interaction force. Then, for every stable operating point found, we proceed by linearizing the system around a transportation scenario, corresponding to the state of the system after $5.0$ \si{\second} given a reference velocity to the master of $\dot{\Lambda}_d = (0.5, 0.5, 0)^T$ \si{\meter \per \second}. The robust performance margin is computed around this second operating point. The robust stability margin is computed as the minimum between the  robust stability margin obtained with the system at rest and with the system in transportation scenario.
\par
The main results of our analysis are collected in Figure \ref{fig:RsRpAnalysisResult}. The first row of the plot contains the level curves which define the robust performance margin for $2$, $3$, $4$, $5$ and $10$ agents, for different values of $\mathcal{C}$ (\textit{Virtual Damping}) and $\mathcal{M}$ (\textit{Virtual Mass}) of the admittance controller. Similarly, the lower row represents the level curves which define the robust stability margin. 
\par From such analysis we can observe that:
\begin{itemize}
	\item robust stability can be achieved for every of the number of agents taken into account. The cardinality of the subset of robustly stable points $S_{rob\_stab} \subset S$ increases as the number of agents increases. The maximum value of robust stability margin achievable decreases as the number of agents increases.
	\item robust performance can be achieved for every of the number of agents taken into account. The cardinality of the set of robustly performant tuning points $S_{rob\_perf} \subset S$ decreases as the number of agents increases. Similarly, the maximum value of robust performance margin achievable decreases as the number of agents increases. 
	\item if the number of agents increases, in order to guarantee peak performance, the \textit{Virtual Damping} has to be decreased. This has the negative effect of pushing the system in a less robustly-stable area.
\end{itemize}

\paragraph{Stability and performance analysis validation for two and five agents}
\begin{figure*}
	\captionsetup[subfigure]{justification=centering}
	\centering
	\begin{subfigure}{.5\textwidth}
		\centering
		\includegraphics[width=1.05\linewidth]{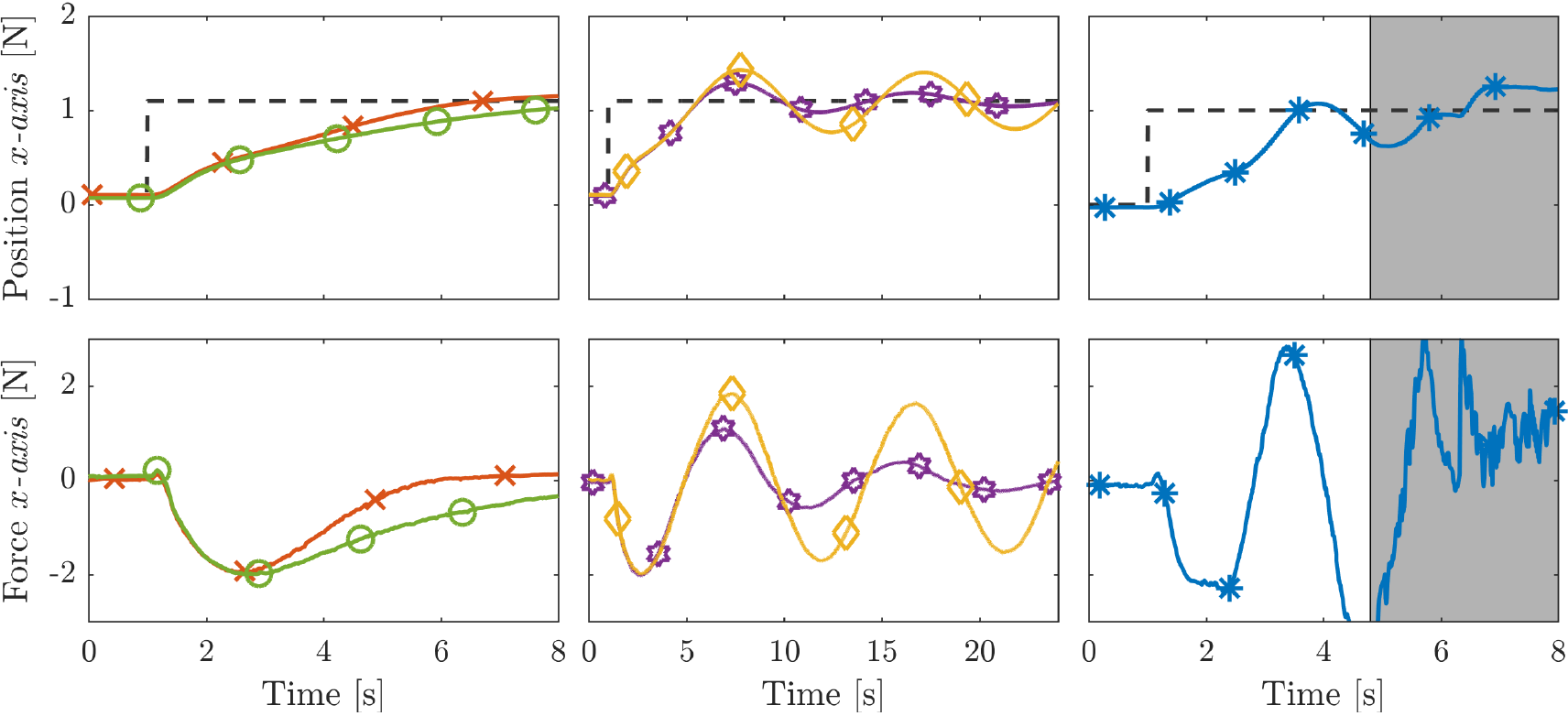}
		\caption{}
		\label{fig:Sim2MavsPosForce}
	\end{subfigure}%
	\begin{subfigure}{.5\textwidth}
		\centering
		\includegraphics[width=.8\linewidth]{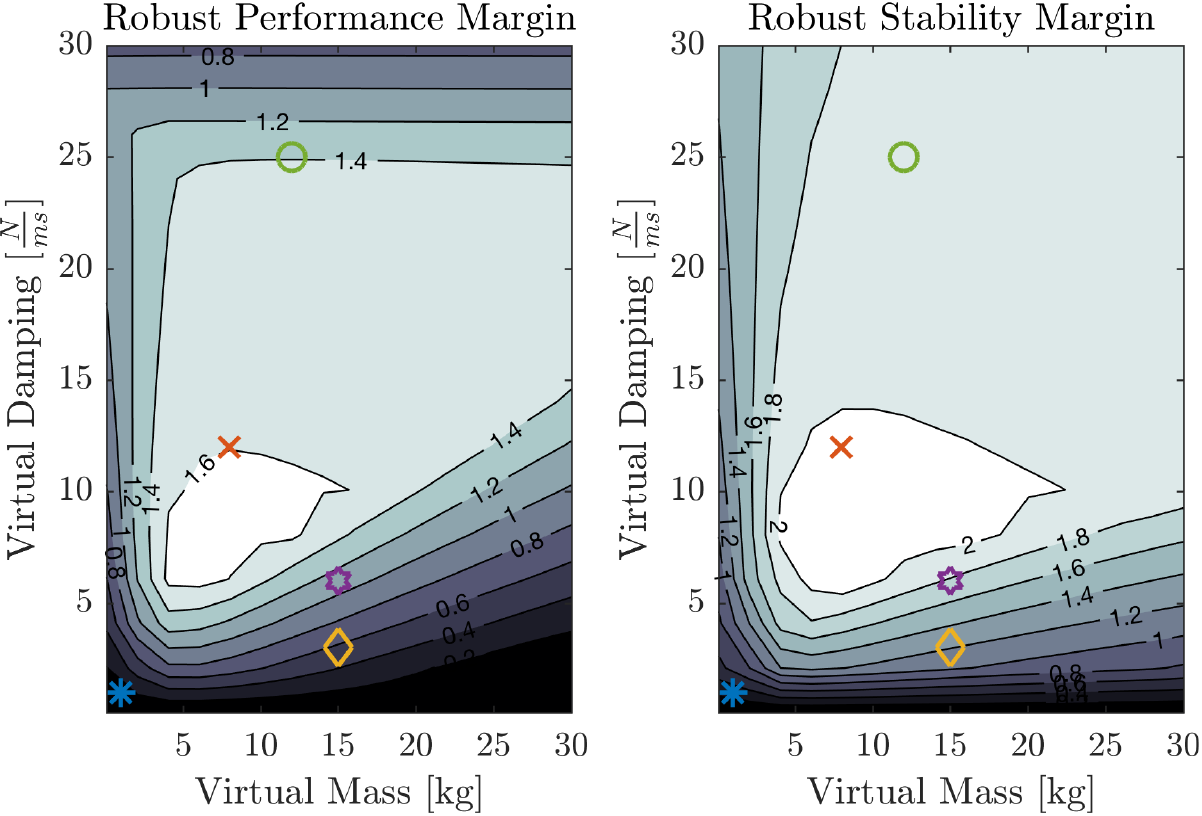}
		\caption{}
		\label{fig:RsRp2Mavs}
	\end{subfigure}
	\caption{Stability and performance results for different tuning points of the system constituted by one master and one slave agent. The results are obtained via the high-fidelity simulator. Regarding Figure \ref{fig:Sim2MavsPosForce}, the dashed lines represent the reference position for the master agent, while the continuous line the position of the master agent and the estimated external force on it. 
	Figure \ref{fig:RsRp2Mavs} represents the results for the robust stability and performance using two agents (as in Figure \ref{fig:RsRpAnalysisResult}), where we have additionally highlighted the tuning points selected for the validation of our analysis. We can observe that tuning points corresponding to higher values for the robust performance margin perform better in therm of reference tracking and convergence of the estimated interaction force to zero (first two columns). The non-stable tuning point causes the instability of the system and the disconnection (gray area) of the two agents from the payload (last column) }
	\label{fig:2MAVsBadParams}
\end{figure*}
\begin{figure*}
	\captionsetup[subfigure]{justification=centering}
	\centering
	\begin{subfigure}{.5\textwidth}
		\centering
		\includegraphics[width=1.02\linewidth]{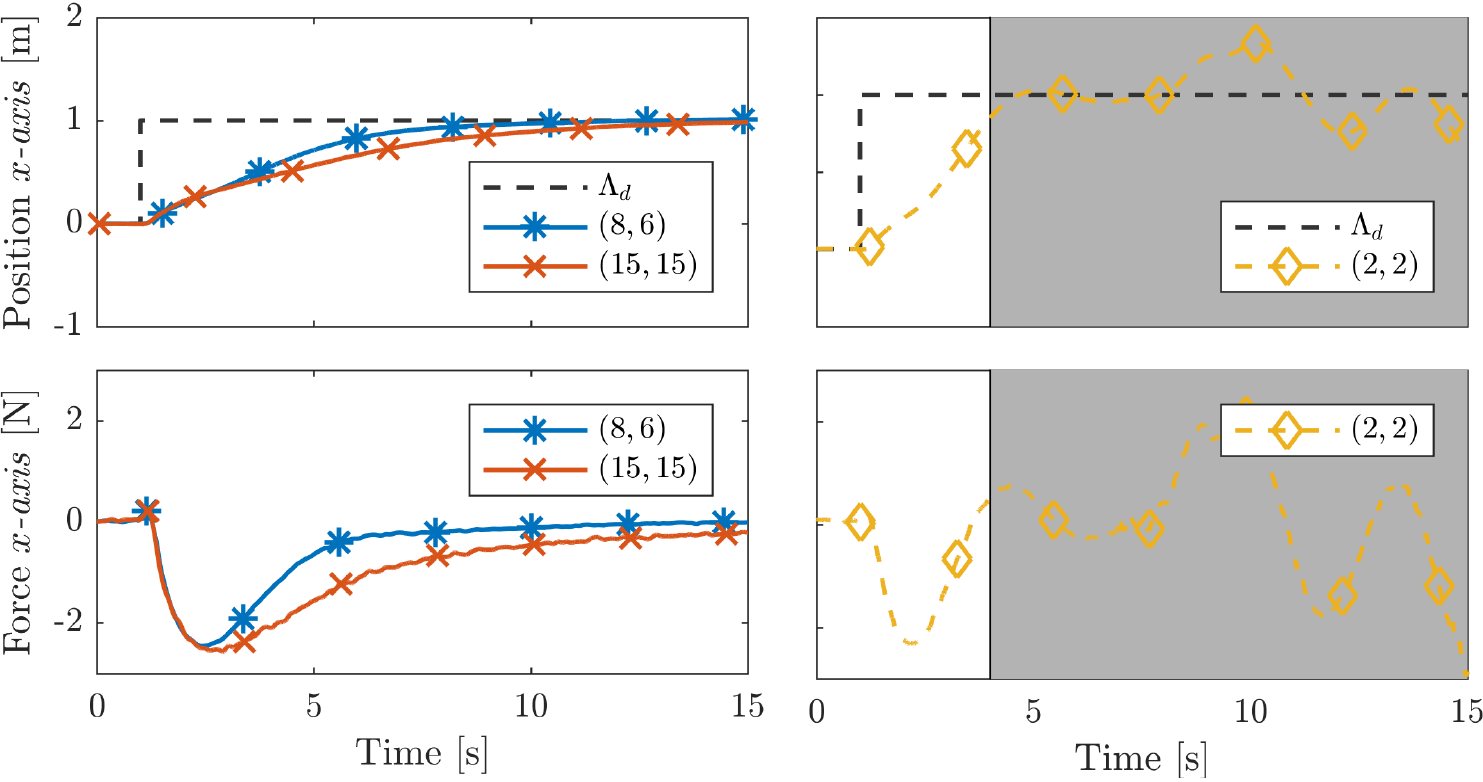}
		\caption{}
		\label{fig:Sim5MavsPosForce}
	\end{subfigure}%
	\begin{subfigure}{.5\textwidth}
		\centering
		\includegraphics[width=.8\linewidth]{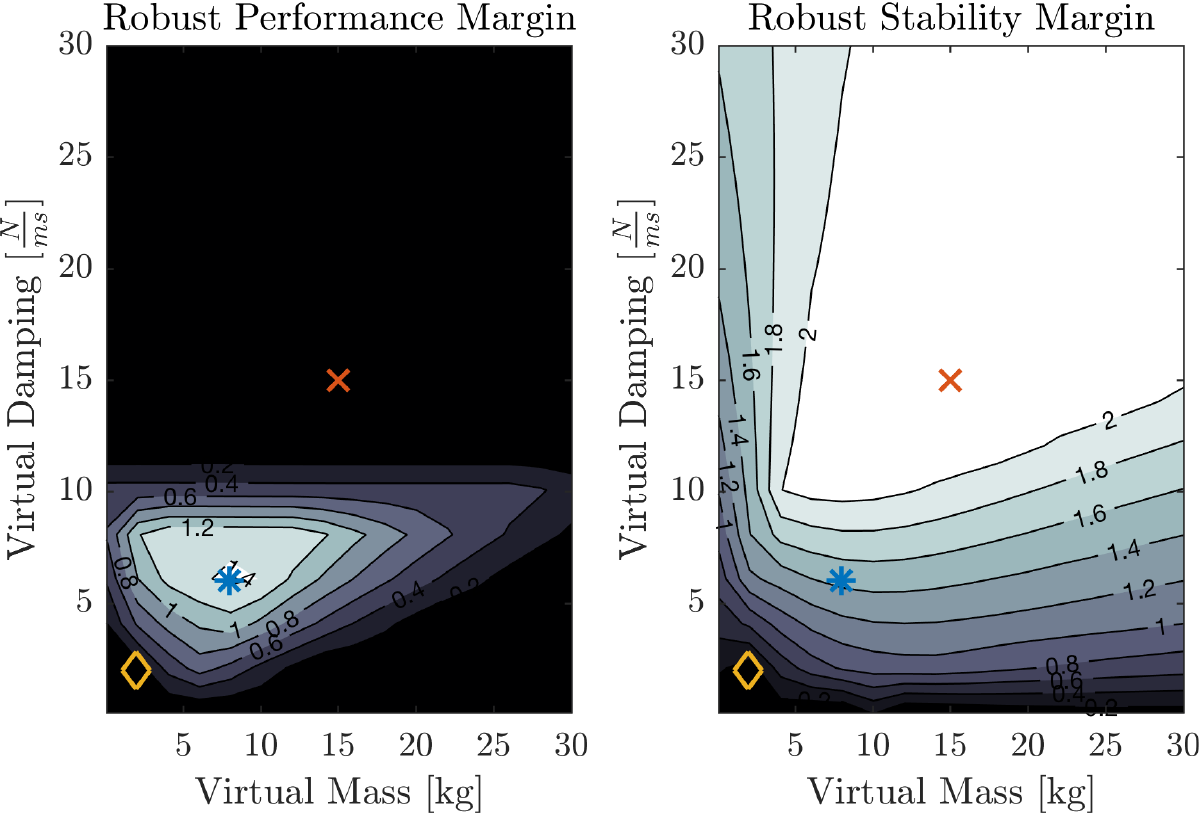}
		\caption{}
		\label{fig:RsRp5Mavs}
	\end{subfigure}
	\caption{Similarly to Figure \ref{fig:2MAVsBadParams}, here we present the stability and performance results for different tuning points of the system constituted by five \acp{MAV}. The results are obtained via the high-fidelity simulator. In the simulated scenario, the robots are rigidly connected via a magnetic gripper with a spherical joint to a $2.0$ kg payload. Figure \ref{fig:Sim5MavsPosForce} shows the position and force response of the master agent to a $1.0$ m reference position step, for different tuning points of the admittance controller. Figure \ref{fig:RsRp5Mavs} shows the result of the robust performance and stability analysis with 5 \acp{MAV}. In red we have highlighted the tuning points used in the validation scenario of Figure \ref{fig:Sim5MavsPosForce}.}
	\label{fig:5MAVsBadParams}
\end{figure*}
\begin{figure*}
	\captionsetup[subfigure]{justification=centering}
	\begin{subfigure}{.5\textwidth}
			\centering
		\includegraphics[width=.85\linewidth]{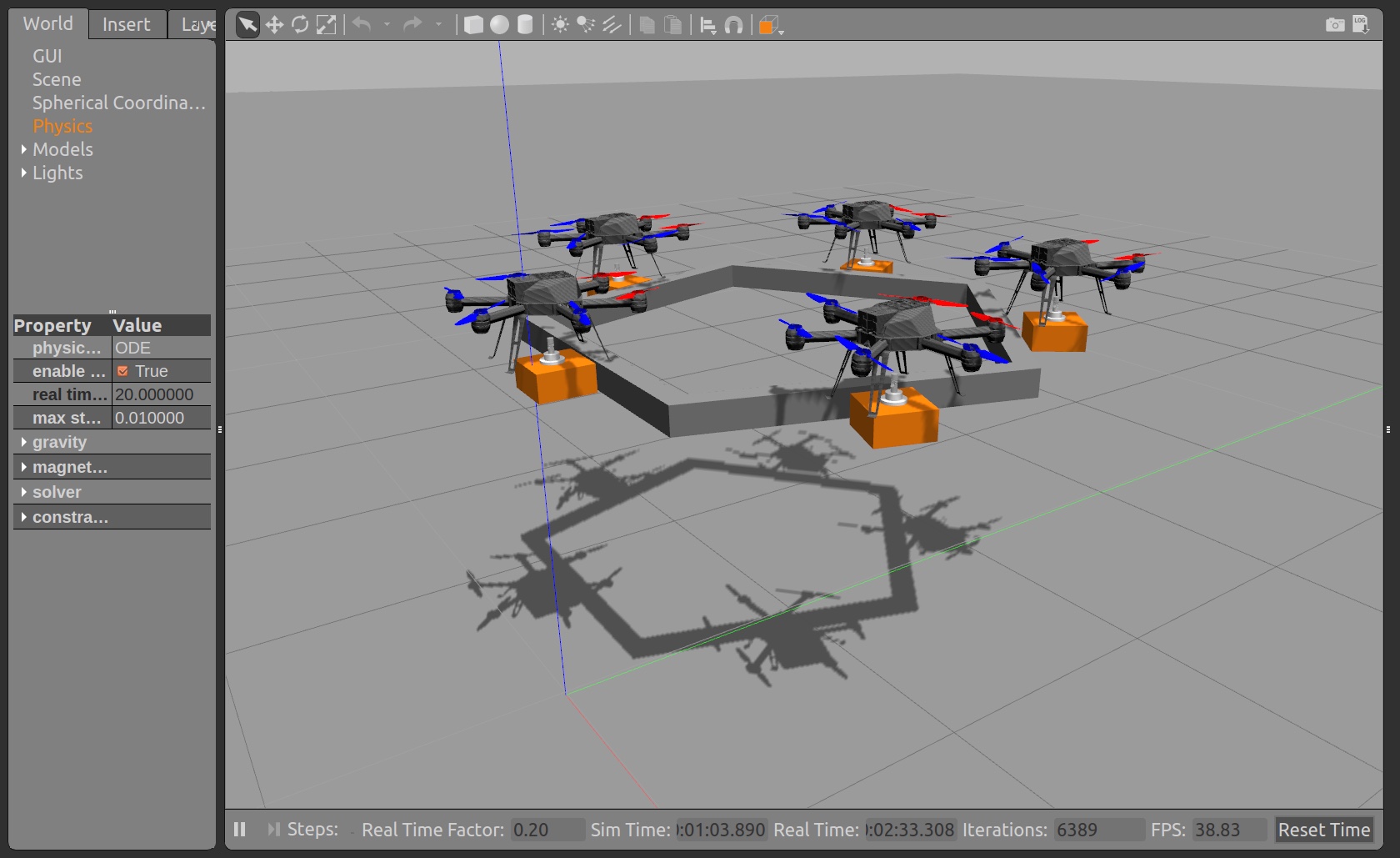}
		\caption{Stable tuning}
		\label{fig:5MAVsSimulationScenarioTrajGazeboStable}
	\end{subfigure}%
	\begin{subfigure}{.5\textwidth}
			\centering
		\includegraphics[width=.85\linewidth]{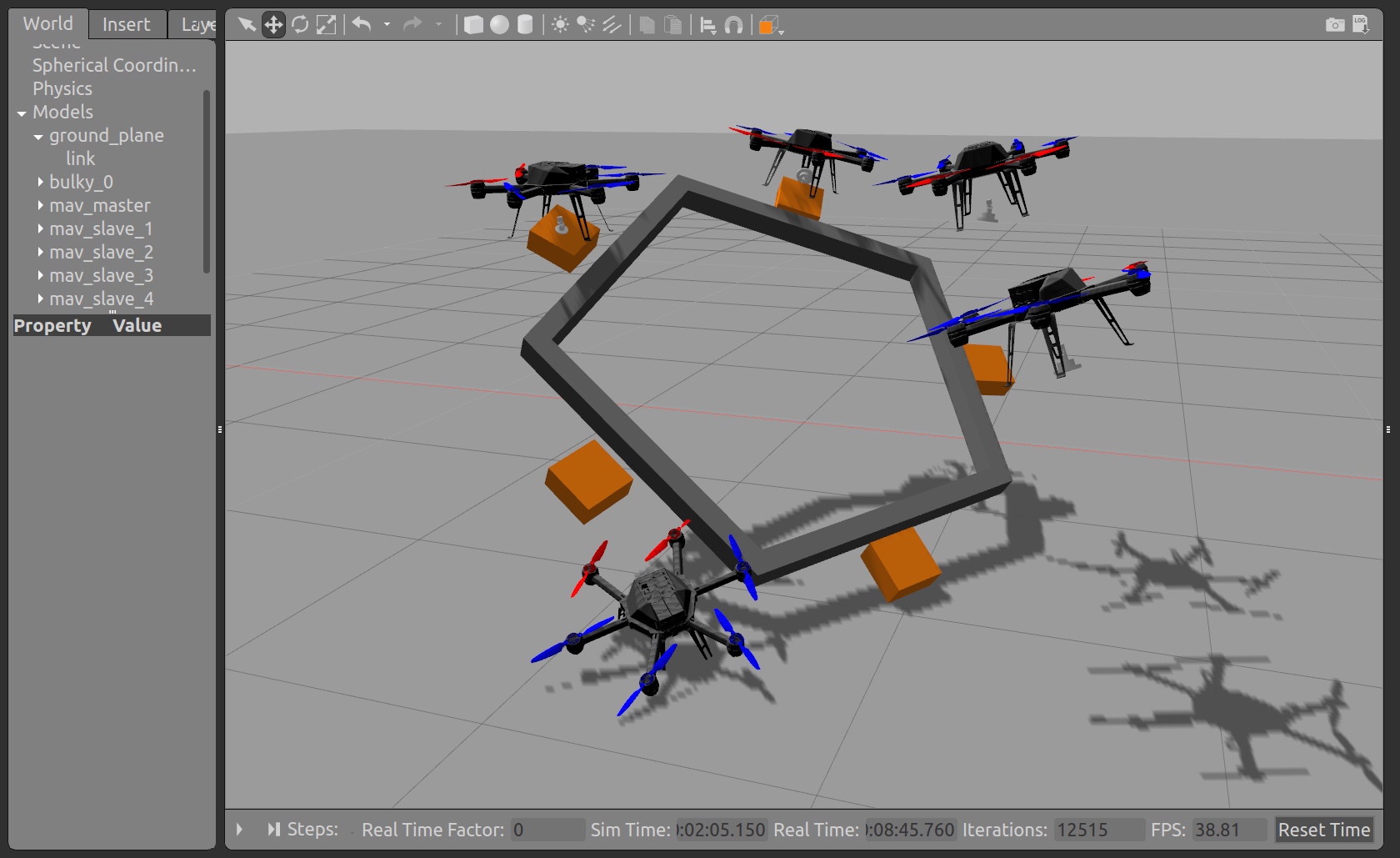}
		\caption{Unstable tuning}
		\label{fig:5MAVsSimulationScenarioTrajGazeboUnstable}
	\end{subfigure}
	\label{fig:5MAVsSimulationScenario}
	\caption{Simulation scenario for the collaborative transportation using five agents. The simulation environment used is the open-source software \textbf{RotorS} developed by \cite{Furrer2016} at the Autonomous Systems Lab, ETH Zurich. The simulator includes a model of the magnetic gripper equipped with a spherical joint and a model of the \textbf{Asctec Neo} hexa-copter, used for the experimental results.}
\end{figure*}
\begin{figure}
	\centering
	\includegraphics[width=\linewidth]{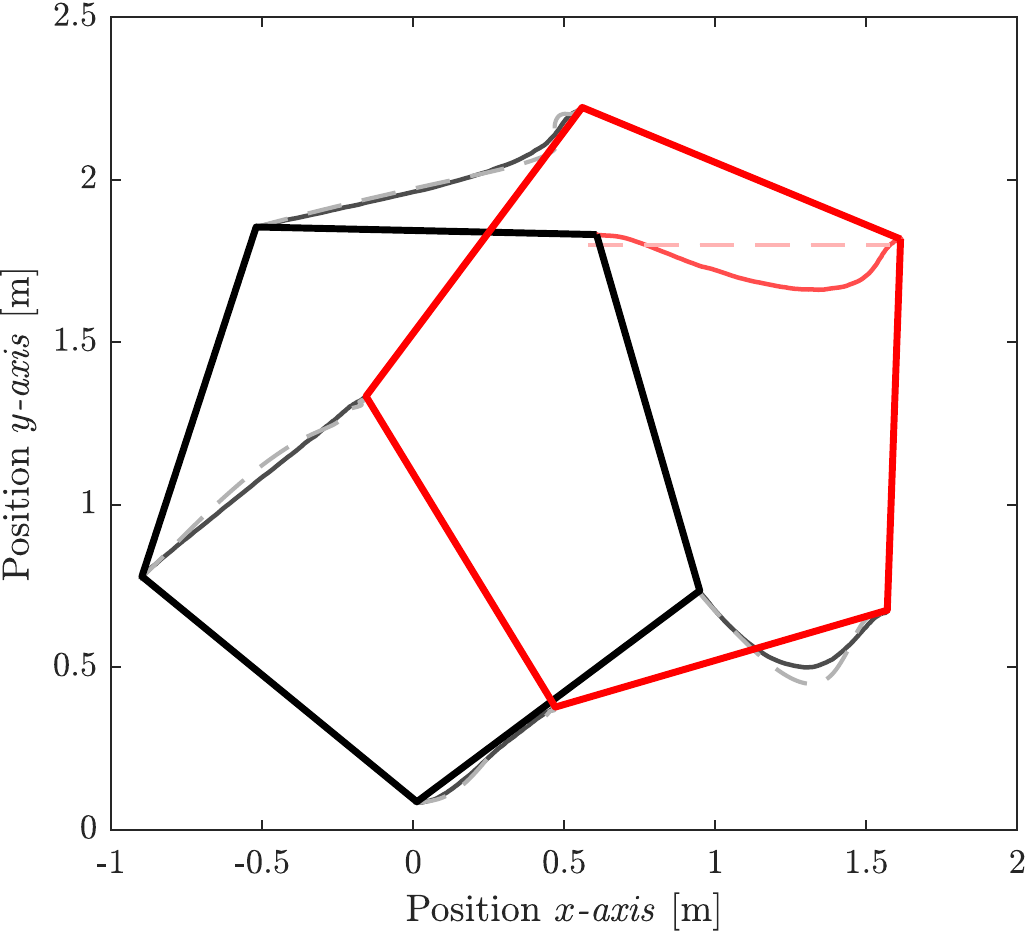}
	\caption{Reference trajectories (dashed lines) and the actual position (continuous lines) of the five agents, given a step reference of $1.0$ \si{\meter} along the \textit{$x$-axis} to the master agent. The system is tuned using the robustly stable and robustly performant parameters $(\mathcal{M}, \mathcal{C}) = (8, 6)$. Reference and position of the master agent are colored in red. The red object represent the final position reached by the system.}
	\label{fig:5MAVsStableTrajectories}
\end{figure}
In order to validate our analysis, we compare the stability and performance of some carefully chosen tuning points with some values arbitrary selected from the considered tuning space $S$ of the admittance controller. The comparison is performed using the hi-fidelity simulation environment \textbf{RotorS} developed by \cite{Furrer2016}, where we model our hardware platform, including the magnetic gripper equipped with a spherical joint, and the payload.
\par We validate out analysis by applying a step of $1.0$ \si{\meter} on the position reference of the simulated master agent and by monitoring the convergence to zero of the estimated external force on it, meaning that the system correctly follows the movements of the master, generating zero internal forces. as well as the reference trajectory generated by the admittance controller. 
\par The results of the validation for two \acp{MAV} are represented in Figure \ref{fig:2MAVsBadParams}, while the results for five \acp{MAV} are represented in Figure \ref{fig:5MAVsBadParams}. In both the cases we apply a step of $1.0$ \si{\meter} to the position reference of the master agent and we monitor its position and the interaction force acting on it. For both the cases of two and five agents, all the points selected from the region with robust stability margin bigger than one show a stable behavior. The points selected from an area with robust stability margin much smaller than one show instead an unstable behavior. Performance can be compared by observing response of the interaction force estimated on the master. As expected, points corresponding to higher values of the robust performance margin show a quicker convergence of the force to zero. 

\begin{table}
\begin{center}
\begin{tabular}{|r|c|c|}
	\hline
	Parameter & Value & Unit \\
	\hline \hline
    Agent's mass $m$ & $3.5$ & \si{\kilogram}  \\ \hline
	Agent's max. payload $\bar{m}$ & 1.0 & \si{\kilogram}  \\ \hline
	Agent's $\phi_{\text{cmd}}^{max}$& 0.26 & \si{\radian} \\ \hline
	Agent's $\theta_{\text{cmd}}^{max}$& 0.26 & \si{\radian} \\ \hline
	Sys. inertia uncertainty & $(10\%, 10\%, 10\%)$ & - \\ \hline
	Pos. controller $\boldsymbol{K}_{P}$ & $\diag(17, 17, 30)$ & - \\ \hline
	Pos. controller $\boldsymbol{K}_{D}$ & $\diag(15, 15, 10)$ & - \\ \hline
	Attitude time const. $\tau_{\text{att}}$ & $0.25$ & \si{\second} \\ \hline
	Force est. time const. $\tau_{\text{est}}$ & $0.2$ & \si{\second} \\ 
	\hline	
\end{tabular}
\end{center}
\caption{Model parameter selected for the robust performance and stability analysis shown in Figure \ref{fig:RsRpAnalysisResult}.}
\label{tab:ParametersRsRpAnalysis}
\end{table}
\subsection{Outdoor transportation using 2 MAVs}
In this section we evaluate the robustness of our approach in outdoor conditions. The transported payload is a beam of wood, $1.5$ \si{\meter} long and $1.8$ \si{\kilogram} heavy, equipped with two metallic plates at its extremities to allow the connection of the magnetic gripper. The agents employed in the transportation are two, one acting as a master and the other as a slave. We remark that the weight of the transported object far exceeds the maximum payload capacity of a single of ours \ac{MAV} (approximately $1.0$ \si{\kilogram}). The state estimation is provided for both the agents by the \ac{VI} Navigation System and the \ac{ROVIO} framework. The experimental setup is represented in Figure \ref{fig:2MavsOutdoor}.
\par The tuning parameters for the admittance controller of the slave agent are obtained from the robust stability and robust performance analysis and are set to $(\mathcal{M}, \mathcal{C}) = (8, 12)$.
\par The experiment is performed as follows. Both the agents starts on the ground, already connected to the payload via the magnetic gripper. A centralized state machine, which runs on an external processor, coordinates the takeoff, making sure that the agents reach simultaneously the same increment of altitude from their starting position. Once the desired increment of altitude is reached, the centralized state machines engages the slave's admittance controller and gives control of the master agent to a human operator. From now on, and during the entire transportation phase, no communication between the robots or the external processor is required. The landing maneuver is again performed by the centralizes processor, which disengages the admittance controller and commands altitude decrements to the agents. 
\par
\begin{figure}
	\centering
	\includegraphics[width=0.4\textwidth]{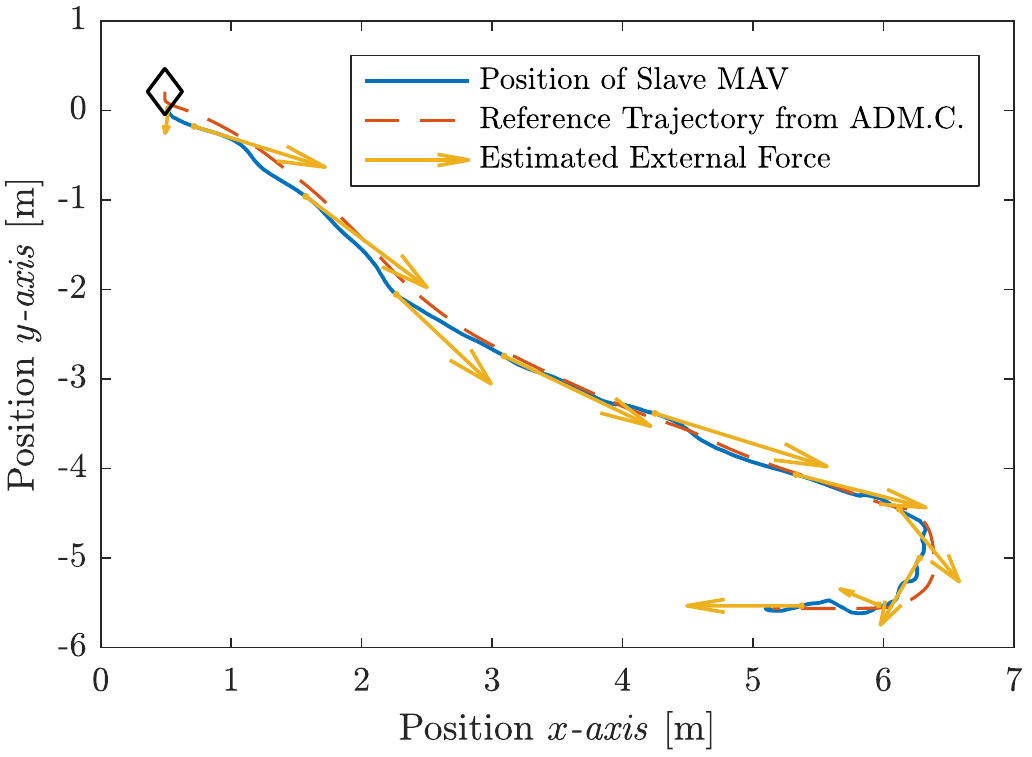}
	\caption{Slave's \ac{MAV} position, reference trajectory and estimated external force during the outdoor transportation of a $1.8$ \si{\kilogram} wooden beam, $1.5$ \si{\meter} long.}
	\label{fig:OutdoorTrajectoryXy}
\end{figure}

\begin{figure}
	\centering
	\includegraphics[width=0.45\textwidth]{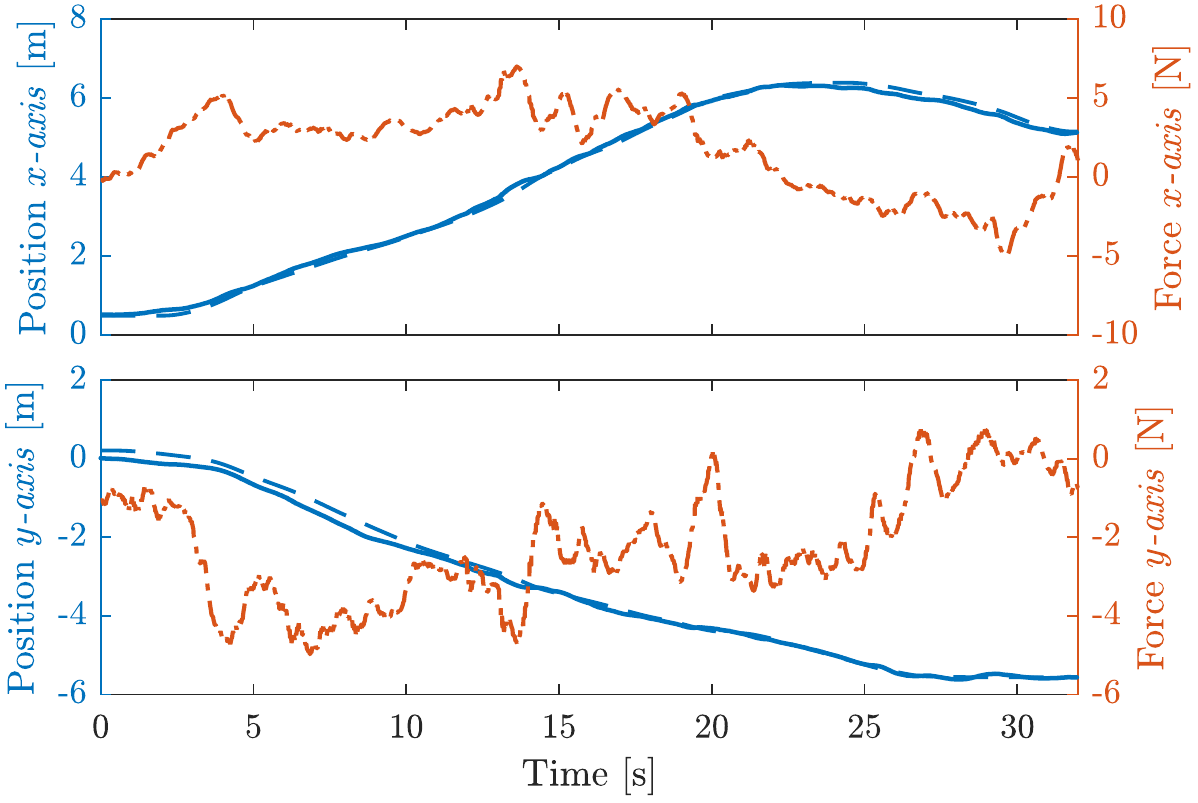}
	\caption{Slave's \ac{MAV} position, reference trajectory and estimated external force during the outdoor transportation of a $1.8$ \si{\kilogram} wooden beam, $1.5$ \si{\meter} long. The initial position is marked with a diamond.}
	\label{fig:OutdoorTrajectoryXyPerAxis}
\end{figure}

 The trajectory of the slave agent during the outdoor collaborative transportation is shown in Figure \ref{fig:OutdoorTrajectoryXy} and \ref{fig:OutdoorTrajectoryXyPerAxis}, where we also highlight the estimated external force used for the generation of the trajectory.
\subsection{Cooperative transportation using 3 MAVs} 
\begin{figure}
	\centering 
	\includegraphics[width=0.45\textwidth]{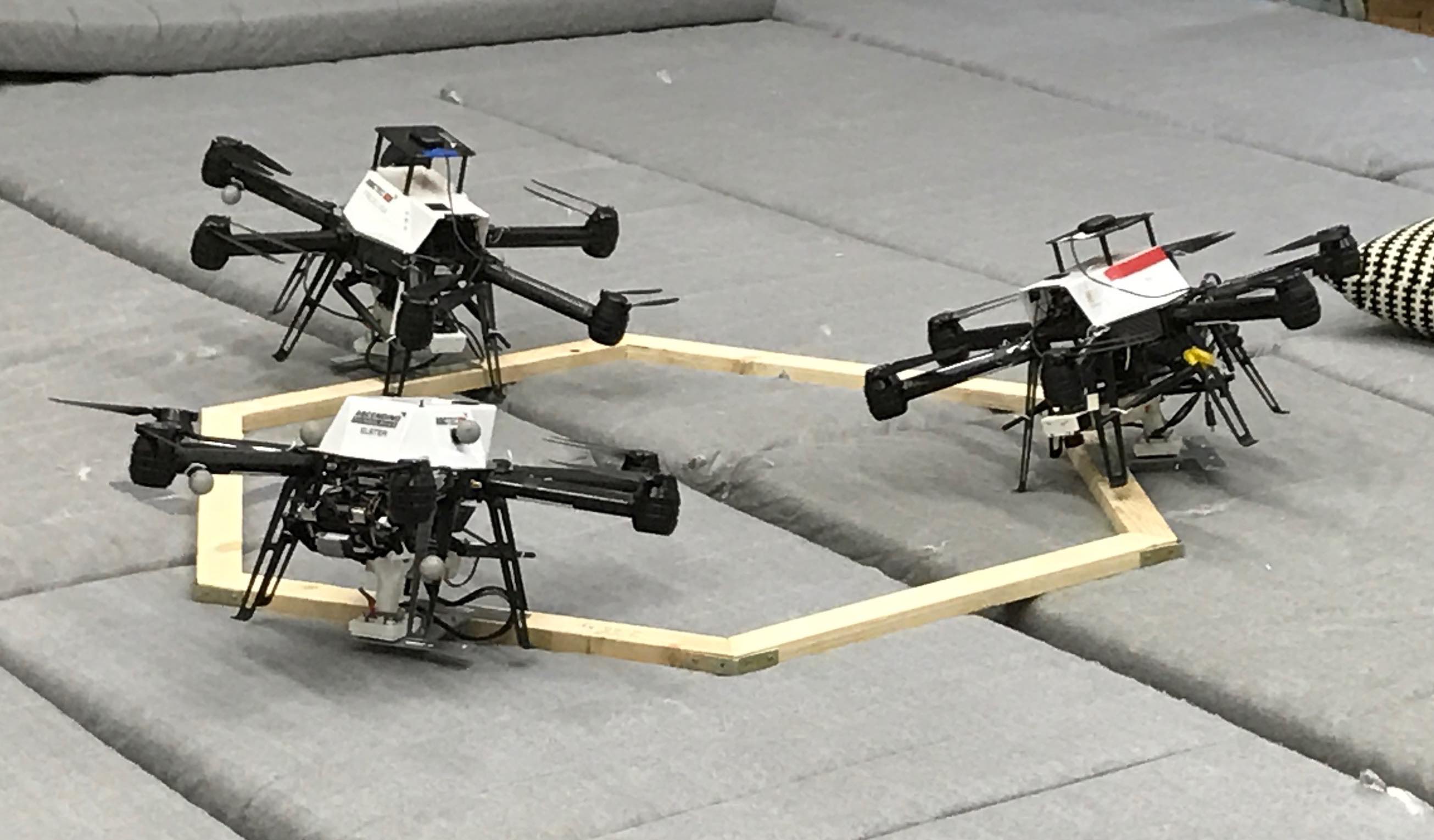}
	\caption{Experimental setup for the collaborative transportation of a bulky payload using three \acp{MAV}. The payload is an hexagon made of wood of $0.7$ m of side length, weighting $2.46$ kg. The distance between each \ac{MAV} is approximately $1.2$ m.}
	\label{fig:3MAVsExperimentalSetup}
\end{figure}
Finally we present the experimental results for the cooperative transportation of a bulky object using 3 \acp{MAV}. The setup of the experiment is represented in Figure \ref{fig:3MAVsExperimentalSetup}. The payload is a hexagonal structure made of wood, of $0.7$ \si{\meter} of side length and mass $2.46$ \si{\kilogram}. It presents three metallic supports used to connect the hexa-copters to it via the magnetic gripper. The distance between each \ac{MAV} is approximately $1.2$ \si{\meter}. The state estimator used for this experiment is based on the on-board \ac{VI} Navigation System and \ac{ROVIO} framework. We remark that the weight of the transported structure exceeds the maximum payload capacity of two of our \acp{MAV}.
\par
The master agent is remotely controlled by a human operator, while the trajectory for each slave agent is generated on board by the admittance controller. We remark that in this experiment there is no global reference frame shared between the robots, and every \ac{MAV} has its own arbitrary inertial reference system. During the transportation phase all the algorithms are executed on-board and no communication between the agents or with a ground station is required.
\par The coordination of the agents during the lifting and landing maneuver is instead guaranteed by a centralized \ac{FSM}. The \ac{FSM} makes sure that the robots simultaneously reach a predefined increment of altitude with respect to their starting position. Once every vehicle has reached the agreed increment of altitude, the \ac{FSM} engages the admittance controller on every predefined slave agent and allows an operator to control the master.
\par The tuning parameters for the admittance controller are obtained from the robust performance and stability analysis for three \acp{MAV} and are set, for each slave agent, to $(\mathcal{M}, \mathcal{C}) = (5, 10)$.
\par
\begin{figure*}
	\centering
	\includegraphics[width=0.7\textwidth]{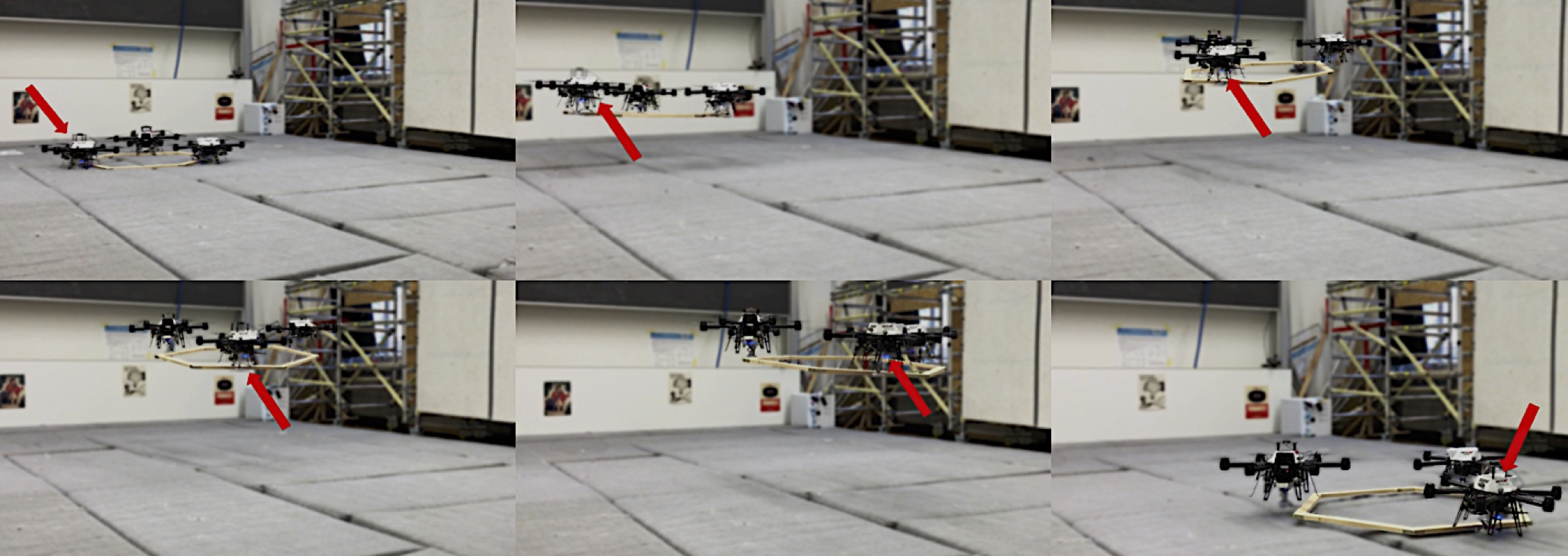}
	\caption{Experimental transportation of a $2.46$ kg wood structure using 3 \acp{MAV}. The master agent is marked with a red arrow.}
	\label{fig:3MavsMovieStrip}
\end{figure*}

\begin{figure}
	\centering
	\includegraphics[width=0.45\textwidth]{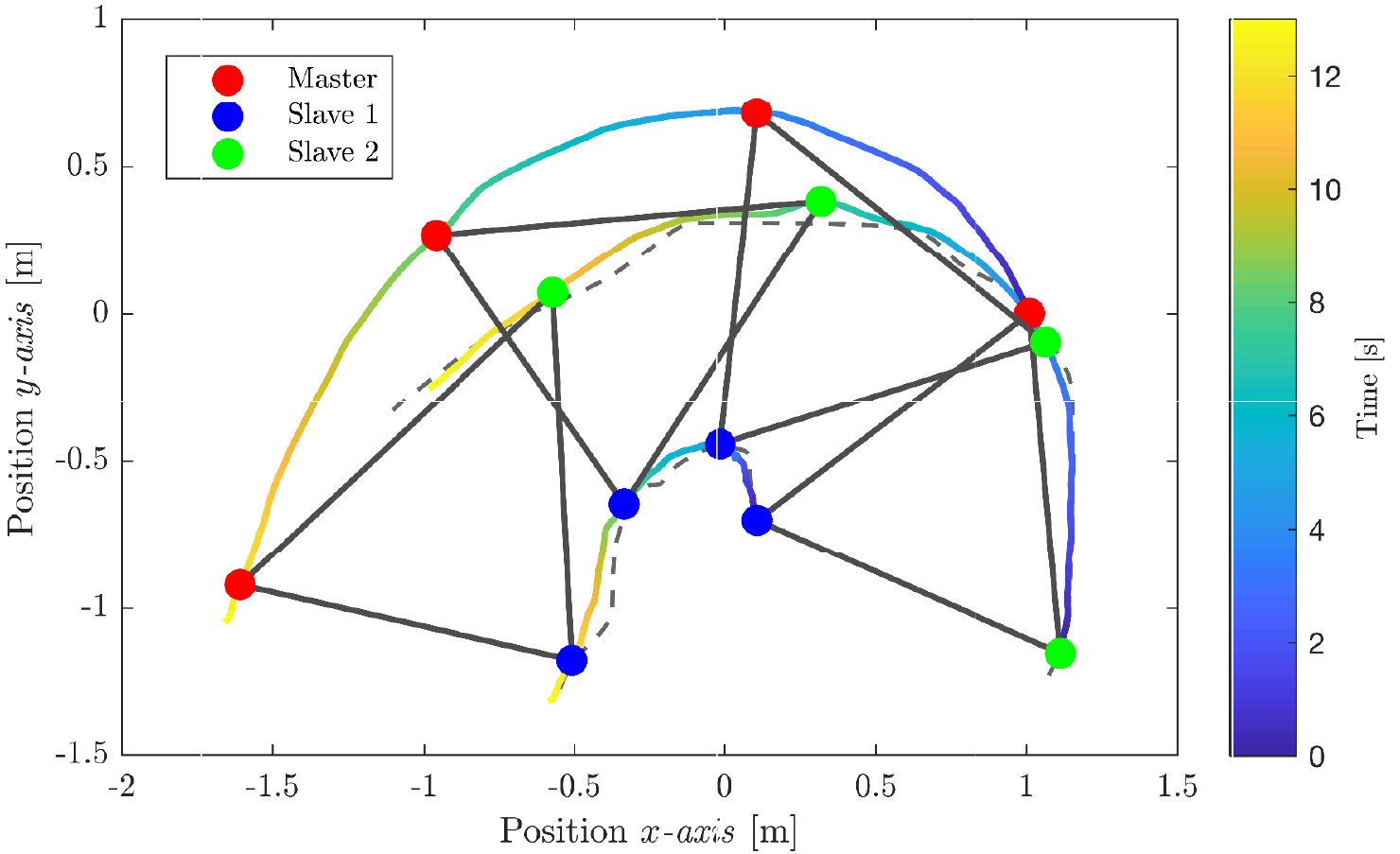}
	\caption{Position of the three agents in a U-turn maneuver during collaborative transportation of a $2.46$ kg wood structure. The dashed line represents the reference trajectory generated by each slave \ac{MAV}. The rotation and translation of the payload can be observed via the black lines which connects the agent at specific time stamps. The estimated external forces on each agent during such maneuver are displayed in figure \ref{fig:3MavsTrajectoryForce}. }
	\label{fig:3MavsTrajectory}
\end{figure}

\begin{figure}
	\centering
	\includegraphics[width=0.45\textwidth]{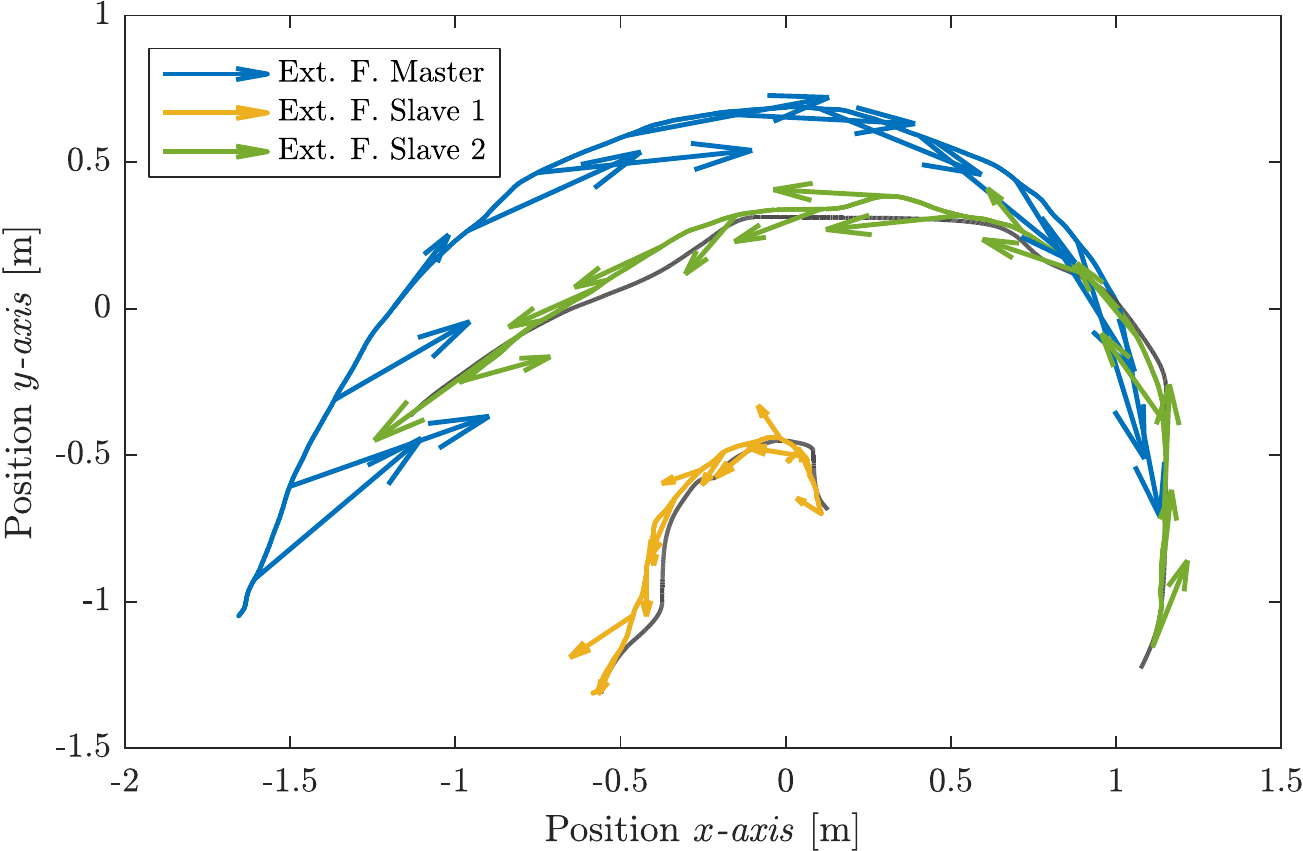}
	\caption{Estimated external force on each agent during the U-turn transportation maneuver described in \ref{fig:3MavsTrajectory}. The grey lines represents the trajectory generated by the admittance controller of each slave agent.}
	\label{fig:3MavsTrajectoryForce}
\end{figure}

 Some key frames of the experiment are shown in Figure \ref{fig:3MavsMovieStrip}. The collaborative transportation has been repeated using a \ac{MCS}-based state estimator in order to record force and trajectories of the different agents with respect to the same inertial frame and simplify the analysis and the display of the recorded data. The results of this second transportation are shown in Figure \ref{fig:3MavsTrajectory}, where we show the trajectory of each agent while the payload makes a U-turn, and in Figure \ref{fig:3MavsTrajectoryForce}, where we show the estimated external force on each agent during the U-turn maneuver.



\appendix
\section{Appendix A}
\subsection{\acf{MRP}}
Let $\boldsymbol{q} = [{\mathbf{\boldsymbol{q}_v}^T}, {q}_s]$ be a quaternion, where $\mathbf{\boldsymbol{q}_v}$ is the vector part and $q_s$ is the scalar part. A corresponding (but singular) attitude representation $\mathbf{p}$, with $\mathbf{p} \in \mathbb{R}^3$, can be obtained as
\begin{equation} \label{eq:FromQuatToMrp}
\mathbf{p} = f \frac{\mathbf{\boldsymbol{q}_v}}{a+q_s} 
\end{equation}
where $a$ is a parameter from 0 to 1 and $f$ is a scale factor. We will choose $f = 2(a+1)$ and $a$ to be a tunable parameter of the filter. We define $\mathbf{p}$ as the \acf{MRP} corresponding to the normalized attitude quaternion $\boldsymbol{q}$. The inverse representation is obtained as 
\begin{equation} 
\label{eq:FromMrpToQuat}
\begin{aligned}
q_s =& \frac{-a\ \norm{\mathbf{p}}^2 + f \sqrt{f^2+(1-a^2) \norm{\mathbf{p}}^2}}{f^2+\norm{\mathbf{p}}^2} \\
\mathbf{\boldsymbol{q}_v} =& f^{-1}(a+q_s) \mathbf{p}
\end{aligned}
\end{equation}
\par The abbreviation $\boldsymbol{p} \leftarrow \textbf{MRP}(\boldsymbol{q})$ corresponds to Eq. \ref{eq:FromQuatToMrp}, while the abbreviation $\boldsymbol{q} \leftarrow \textbf{MRP}(\boldsymbol{p})$ corresponds to Eq. \ref{eq:FromMrpToQuat}.

\begin{acks}
This work was supported by the European Union's Horizon 2020 research and innovation programme under grant agreement n.644128 (Aeroworks) and the Mohamed Bin Zayed International Robotics Challenge 2017.
\end{acks}

\newpage
\bibliographystyle{SageH}
\bibliography{bibliography/bibliography.bib}

\begin{thebibliography}{36}
\providecommand{\natexlab}[1]{#1}
\providecommand{\url}[1]{\texttt{#1}}
\providecommand{\urlprefix}{URL }
\expandafter\ifx\csname urlstyle\endcsname\relax
  \providecommand{\doi}[1]{DOI:\discretionary{}{}{}#1}\else
  \providecommand{\doi}{DOI:\discretionary{}{}{}\begingroup
  \urlstyle{rm}\Url}\fi

\bibitem[{Achtelik et~al.(2013)Achtelik, Lynen, Chli and
  Siegwart}]{Achtelik2013}
Achtelik MW, Lynen S, Chli M and Siegwart R (2013) {Inversion based direct
  position control and trajectory following for micro aerial vehicles}.
\newblock \emph{IEEE International Conference on Intelligent Robots and
  Systems} : 2933--2939\doi{10.1109/IROS.2013.6696772}.

\bibitem[{Alonso-Mora et~al.(2017)Alonso-Mora, Baker and
  Rus}]{multirobotJavier2017}
Alonso-Mora J, Baker S and Rus D (2017) Multi-robot formation control and
  object transport in dynamic environments via constrained optimization.
\newblock \emph{The International Journal of Robotics Research} 36(9):
  1000--1021.
\newblock \doi{10.1177/0278364917719333}.
\newblock \urlprefix\url{http://dx.doi.org/10.1177/0278364917719333}.

\bibitem[{Ascending-Technologies(2017)}]{AsctecWebsite}
Ascending-Technologies (2017) Leading uas drone technology, autopilot \&
  multirotor uav solutions : Ascending technologies.
\newblock \url{http://www.asctec.de/en/}.
\newblock (Accessed on 08/17/2017).

\bibitem[{Augugliaro and D'Andrea(2013)}]{augugliaro2013admittance}
Augugliaro F and D'Andrea R (2013) Admittance control for physical
  human-quadrocopter interaction.
\newblock In: \emph{Control Conference (ECC), 2013 European}. IEEE, pp.
  1805--1810.

\bibitem[{B{\"a}hnemann et~al.(2017)B{\"a}hnemann, Schindler, Kamel, Siegwart
  and Nieto}]{bahnemann2017decentralized}
B{\"a}hnemann R, Schindler D, Kamel M, Siegwart R and Nieto J (2017) A
  decentralized multi-agent unmanned aerial system to search, pick up, and
  relocate objects.
\newblock \emph{arXiv preprint arXiv:1707.03734} .

\bibitem[{Balas et~al.(1993)Balas, Doyle, Glover, Packard and
  Smith}]{balas1993mu}
Balas GJ, Doyle JC, Glover K, Packard A and Smith R (1993) $\mu$-analysis and
  synthesis toolbox.
\newblock \emph{MUSYN Inc. and The MathWorks, Natick MA} .

\bibitem[{Bernard et~al.(2011)Bernard, Kondak, Maza and
  Ollero}]{bernard2011autonomous}
Bernard M, Kondak K, Maza I and Ollero A (2011) Autonomous transportation and
  deployment with aerial robots for search and rescue missions.
\newblock \emph{Journal of Field Robotics} 28(6): 914--931.

\bibitem[{Bircher et~al.(2016{\natexlab{a}})Bircher, Kamel, Alexis, Burri,
  Oettershagen, Omari, Mantel and Siegwart}]{Bircher2016}
Bircher A, Kamel M, Alexis K, Burri M, Oettershagen P, Omari S, Mantel T and
  Siegwart R (2016{\natexlab{a}}) Three-dimensional coverage path planning via
  viewpoint resampling and tour optimization for aerial robots.
\newblock \emph{Autonomous Robots} 40(6): 1059--1078.
\newblock \doi{10.1007/s10514-015-9517-1}.
\newblock \urlprefix\url{https://doi.org/10.1007/s10514-015-9517-1}.

\bibitem[{Bircher et~al.(2016{\natexlab{b}})Bircher, Kamel, Alexis, Oleynikova
  and Siegwart}]{nbvp2016}
Bircher A, Kamel M, Alexis K, Oleynikova H and Siegwart R (2016{\natexlab{b}})
  Receding horizon "next-best-view" planner for 3d exploration.
\newblock In: \emph{2016 IEEE International Conference on Robotics and
  Automation (ICRA)}. pp. 1462--1468.
\newblock \doi{10.1109/ICRA.2016.7487281}.

\bibitem[{Bloesch et~al.(2015)Bloesch, Omari, Hutter and
  Siegwart}]{Bloesch2015a}
Bloesch M, Omari S, Hutter M and Siegwart R (2015) {Robust Visual Inertial
  Odometry Using a Direct EKF-Based Approach} \doi{10.3929/ethz-a-010566547}.
\newblock \urlprefix\url{http://e-collection.library.ethz.ch/view/eth:48374}.

\bibitem[{Bouabdallah(2007)}]{bouabdallah2007design}
Bouabdallah S (2007) Design and control of quadrotors with application to
  autonomous flying .

\bibitem[{Burri et~al.(2016)Burri, Nikolic, Oleynikova, Achtelik and
  Siegwart}]{burri2016maximum}
Burri M, Nikolic J, Oleynikova H, Achtelik MW and Siegwart R (2016) Maximum
  likelihood parameter identification for mavs.
\newblock In: \emph{Robotics and Automation (ICRA), 2016 IEEE International
  Conference on}. IEEE, pp. 4297--4303.

\bibitem[{Crassidis and Markley(2003)}]{crassidis2003unscented}
Crassidis JL and Markley FL (2003) Unscented filtering for spacecraft attitude
  estimation.
\newblock \emph{Journal of Guidance Control and Dynamics} 26(4): 536--542.

\bibitem[{Fumagalli et~al.(2012)Fumagalli, Naldi, Macchelli, Carloni,
  Stramigioli and Marconi}]{fumagalli2012modeling}
Fumagalli M, Naldi R, Macchelli A, Carloni R, Stramigioli S and Marconi L
  (2012) Modeling and control of a flying robot for contact inspection.
\newblock In: \emph{Intelligent Robots and Systems (IROS), 2012 IEEE/RSJ
  International Conference on}. IEEE, pp. 3532--3537.

\bibitem[{Furrer et~al.(2016)Furrer, Burri, Achtelik and Siegwart}]{Furrer2016}
Furrer F, Burri M, Achtelik M and Siegwart R (2016) \emph{Robot Operating
  System (ROS): The Complete Reference (Volume 1)}, chapter RotorS---A Modular
  Gazebo MAV Simulator Framework.
\newblock Cham: Springer International Publishing.
\newblock ISBN 978-3-319-26054-9, pp. 595--625.

\bibitem[{Gassner et~al.(2017)Gassner, Cieslewski and
  Scaramuzza}]{gassner2017dynamic}
Gassner M, Cieslewski T and Scaramuzza D (2017) Dynamic collaboration without
  communication: Vision-based cable-suspended load transport with two
  quadrotors.
\newblock In: \emph{ICRA}, EPFL-CONF-228458.

\bibitem[{Gawel et~al.(2017)Gawel, Kamel, Novkovic, Widauer, Schindler, von
  Altishofen, Siegwart and Nieto}]{gawel2017aerial}
Gawel A, Kamel M, Novkovic T, Widauer J, Schindler D, von Altishofen BP,
  Siegwart R and Nieto J (2017) Aerial picking and delivery of magnetic objects
  with mavs.
\newblock In: \emph{Robotics and Automation (ICRA), 2017 IEEE International
  Conference on}. IEEE, pp. 5746--5752.

\bibitem[{Gelblum et~al.(2015)Gelblum, Pinkoviezky, Fonio, Ghosh, Gov and
  Feinerman}]{gelblum2015ant}
Gelblum A, Pinkoviezky I, Fonio E, Ghosh A, Gov N and Feinerman O (2015) Ant
  groups optimally amplify the effect of transiently informed individuals.
\newblock \emph{Nature communications} 6.

\bibitem[{Girard et~al.(2004)Girard, Howell and Hedrick}]{BorderPatroUAV1}
Girard A, Howell A and Hedrick J (2004) Border patrol and surveillance missions
  using multiple unmanned air vehicles.
\newblock In: \emph{Decision and Control, 2004. CDC. 43rd IEEE Conference on}.

\bibitem[{Hogan and Buerger(2005)}]{hogan2005impedance}
Hogan N and Buerger SP (2005) Impedance and interaction control.

\bibitem[{Julier and Uhlmann(1996)}]{Julier1996}
Julier SJ and Uhlmann JK (1996) {A General Method for Approximating Nonlinear
  Transformations of Probability Distributions} :
  1--27\doi{10.1371/journal.pone.0006243}.

\bibitem[{Julier and Uhlmann(2004)}]{julier2004unscented}
Julier SJ and Uhlmann JK (2004) Unscented filtering and nonlinear estimation.
\newblock \emph{Proceedings of the IEEE} 92(3): 401--422.

\bibitem[{Kamel et~al.(2016)Kamel, Burri and Siegwart}]{2016arXiv161109240K}
Kamel M, Burri M and Siegwart R (2016) Linear vs nonlinear mpc for trajectory
  tracking applied to rotary wing micro aerial vehicles.
\newblock \emph{arXiv preprint arXiv:1611.09240} .

\bibitem[{Kamel et~al.(2017)Kamel, Stastny, Alexis and Siegwart}]{kamelmpc2016}
Kamel M, Stastny T, Alexis K and Siegwart R (2017) Model predictive control for
  trajectory tracking of unmanned aerial vehicles using robot operating system.
\newblock In: \emph{Robot Operating System (ROS)}. Springer, pp. 3--39.

\bibitem[{Khatib et~al.(1996)Khatib, Yokoi, Chang, Ruspini, Holmberg and
  Casal}]{570849}
Khatib O, Yokoi K, Chang K, Ruspini D, Holmberg R and Casal A (1996)
  Vehicle/arm coordination and multiple mobile manipulator decentralized
  cooperation.
\newblock In: \emph{Intelligent Robots and Systems '96, IROS 96, Proceedings of
  the 1996 IEEE/RSJ International Conference on}, volume~2. pp. 546--553 vol.2.
\newblock \doi{10.1109/IROS.1996.570849}.

\bibitem[{Markley et~al.(2007)Markley, Cheng, Crassidis and
  Oshman}]{Markley2007}
Markley FL, Cheng Y, Crassidis JL and Oshman Y (2007) {Quaternion Averaging}.
\newblock \emph{NASA Goddard Space Flight Center} : 1--10.

\bibitem[{Maza et~al.(2009)Maza, Kondak, Bernard and Ollero}]{maza2009multi}
Maza I, Kondak K, Bernard M and Ollero A (2009) Multi-uav cooperation and
  control for load transportation and deployment.
\newblock In: \emph{Selected papers from the 2nd International Symposium on
  UAVs, Reno, Nevada, USA June 8--10, 2009}. Springer, pp. 417--449.

\bibitem[{McKinnon and Schoellig(2016)}]{mckinnon2016unscented}
McKinnon CD and Schoellig AP (2016) Unscented external force and torque
  estimation for quadrotors.
\newblock In: \emph{IROS}. pp. 5651--5657.

\bibitem[{Michael et~al.(2011)Michael, Fink and Kumar}]{michael2011cooperative}
Michael N, Fink J and Kumar V (2011) Cooperative manipulation and
  transportation with aerial robots.
\newblock \emph{Autonomous Robots} 30(1): 73--86.

\bibitem[{Mueller et~al.(2016)Mueller, Hehn and
  D’Andrea}]{mueller2016covariance}
Mueller MW, Hehn M and D’Andrea R (2016) Covariance correction step for
  kalman filtering with an attitude.
\newblock \emph{Journal of Guidance, Control, and Dynamics} .

\bibitem[{Nikolic et~al.(2014)Nikolic, Rehder, Burri, Gohl, Leutenegger,
  Furgale and Siegwart}]{J.Nikolic2014}
Nikolic J, Rehder J, Burri M, Gohl P, Leutenegger S, Furgale PT and Siegwart R
  (2014) {A Synchronized Visual-Inertial Sensor System with FPGA Pre-Processing
  for Accurate Real-Time SLAM}.
\newblock In: \emph{Robotics and Automation (ICRA), 2014 IEEE Int. Conf. on}.
\newblock ISBN 9781479936847, pp. 431--437.

\bibitem[{Shuster(1993)}]{shuster1993survey}
Shuster MD (1993) A survey of attitude representations.
\newblock \emph{Navigation} 8(9): 439--517.

\bibitem[{Siciliano and Khatib(2016)}]{siciliano2016springer}
Siciliano B and Khatib O (2016) \emph{Springer handbook of robotics}.
\newblock Springer.

\bibitem[{Simon(2006)}]{DSimon2006}
Simon D (2006) \emph{Optimal State Estimation: Kalman, H Infinity, and
  Nonlinear Approaches.}

\bibitem[{Sugar and Kumar(2002)}]{988979}
Sugar TG and Kumar V (2002) Control of cooperating mobile manipulators.
\newblock \emph{IEEE Transactions on Robotics and Automation} 18(1): 94--103.
\newblock \doi{10.1109/70.988979}.

\bibitem[{Tsiamis et~al.(2015)Tsiamis, Verginis, Bechlioulis and
  Kyriakopoulos}]{7353473}
Tsiamis A, Verginis CK, Bechlioulis CP and Kyriakopoulos KJ (2015) Cooperative
  manipulation exploiting only implicit communication.
\newblock In: \emph{2015 IEEE/RSJ International Conference on Intelligent
  Robots and Systems (IROS)}. pp. 864--869.
\newblock \doi{10.1109/IROS.2015.7353473}.

\end{thebibliography}

\begin{acronym}
	\acro{TRADR}{``Long-Term Human-Robot Teaming for Robots Assisted Disaster Response''}
	\acro{MAV}{Micro Aerial Vehicle}
	\acro{MBZIRC}{Mohamed Bin Zayed International Robotics Challenge}
	\acro{UAV}{Unmanned Aerial Vehicle}
	\acro{UKF}{Unscented Kalman Filter}
	\acro{CoG}{Center of Gravity}
	\acro{VI}{Vistual-Inertial}
	\acro{MPC}{Model Predictive Controller}
	\acro{FSM}{Finite State Machine}
	\acro{EKF}{Extended Kalman Filter}
	\acro{MRP}{Modified Rodrigues Parameters}
	\acro{UT}{Unscented Transformation}
	\acro{MIMO}{Multiple Input-Multiple Output}
	\acro{USQUE}{Unscented Quaternion Estimator}
	\acro{ROVIO}{Robust Visual Inertial Odometry}
	\acro{MCS}{Motion Capture System}
	\acro{RMS}{Root Mean Square}
	\acro{IMU}{Inertial Measurement Unit}
\end{acronym}

\end{document}